\documentclass[10pt,journal,compsoc]{IEEEtran}

\usepackage{booktabs} %

\usepackage{indentfirst}
\usepackage{adjustbox}
\usepackage{tabularx}
\usepackage{color}
\usepackage{xcolor,colortbl}
\usepackage[english]{babel}
\usepackage{multirow}
\usepackage{makecell}
\usepackage{tikz}
\usetikzlibrary{spy,backgrounds}
\usepackage{soul}
\usepackage{graphicx}
\usepackage{url}
\usepackage{subcaption}
\usepackage{dsfont}

\usepackage{amssymb}
\usepackage{hyperref}
\usepackage{amsmath}
\usepackage{arydshln}
\usepackage{float}
\usepackage{amsmath}
\usepackage{amssymb}
\usepackage{amsmath,tikz}
\usetikzlibrary{arrows,chains,matrix,positioning,scopes}
\usepackage{booktabs}
\usepackage{lipsum}
\usepackage[ruled]{algorithm2e} %

\SetAlFnt{\small}
\SetAlCapFnt{\small}
\SetAlCapNameFnt{\small}
\SetAlCapHSkip{0pt}

\newcommand{\tabincell}[2]{\begin{tabular}
		{@{}#1@{}}#2\end{tabular}}

\newcommand{\zoomin}[9]{ %
    \begin{tikzpicture}[spy using outlines={rectangle,red,magnification=#8,size=#6}]   
        \node[anchor={#9},inner sep=0]  {\includegraphics[width=#7]{#1}};
        \spy on (#2, #3) in node at (#4,#5);
    \end{tikzpicture}
}

\newcommand{\redbox}[4]{%
    \begin{tikzpicture}
        \node[anchor=south west, inner sep=0] (image) at (0,0) {\includegraphics[width=#4]{#1}};
        \draw[red, thick] #2 rectangle #3;
    \end{tikzpicture}%
}
\newcommand{\bc}[1]{{\cellcolor[HTML]{ffb2b2}#1}}
\newcommand{\sbc}[1]{{\cellcolor[HTML]{ffcc99}#1}}
\newcommand{\tbc}[1]{{\cellcolor[HTML]{ffffb2}#1}}

\newcommand{\wt}[1]{{\color{black}#1}}
\newcommand{\yl}[1]{{\color{black}#1}}

\newcommand{\wttp}[1]{{\color{black}#1}}

\newcommand{\name}{DeferredGS\xspace}
\newcommand{\neusbranch}{\textit{Instant-RefNeuS}\xspace}

\begin{document}
\title{\name: Decoupled and Editable Gaussian Splatting with Deferred Shading}
\author{Tong Wu, Jia-Mu Sun, Yu-Kun Lai, Yuewen Ma, Leif Kobbelt and Lin~Gao\IEEEauthorrefmark{1}
    \thanks{\IEEEauthorrefmark{1} Corresponding Author is Lin Gao (gaolin@ict.ac.cn).}}

\markboth{IEEE TRANSACTIONS ON PATTERN ANALYSIS AND MACHINE INTELLIGENCE, VOL. XX, NO. XX, March 2024}%
{Shell \MakeLowercase{\textit{et al.}}: Bare Demo of IEEEtran.cls for Computer Society Journals}

\IEEEtitleabstractindextext{%
   
    \begin{abstract}
Reconstructing and editing \yl{3D objects and scenes} both play crucial roles in computer graphics and computer vision.
Neural radiance fields \yl{(NeRFs) can achieve} %
realistic reconstruction and editing results but suffer from inefficiency in rendering. 
Gaussian splatting significantly accelerates rendering by rasterizing  Gaussian \yl{ellipsoids}.
However, \yl{Gaussian} splatting utilizes a single Spherical Harmonic (SH) function to model both texture and lighting,  limiting independent editing \yl{capabilities} of these components.  
Recently, attempts have been made to decouple texture and lighting with the \yl{Gaussian} splatting representation but may fail to produce plausible geometry and decomposition results on reflective scenes. 
Additionally, the \textit{forward shading} technique they employ introduces noticeable blending artifacts during relighting, as the geometry attributes of Gaussians are optimized under the original illumination and may not be suitable for novel lighting conditions. 
To address these issues, we introduce \name, a method for decoupling and editing the \yl{Gaussian} splatting representation using \textit{deferred shading}. 
To achieve successful decoupling, we model the illumination with a learnable environment map and define additional attributes such as texture parameters and normal direction on \yl{Gaussians}, where the normal is distilled from a jointly trained signed distance function. 
More importantly, we apply \textit{deferred shading}, resulting in more realistic relighting effects compared to previous methods. 
Both qualitative and quantitative experiments demonstrate the superior performance of \name in novel view synthesis and editing tasks. 
\end{abstract}

}

\maketitle
\IEEEdisplaynontitleabstractindextext
\IEEEpeerreviewmaketitle

\section{Introduction}
\label{sec:intro}

Reconstructing and editing 3D scenes from multi-view images is an important task in computer graphics and computer vision. 
Recently, Neural Radiance Fields \yl{(NeRFs)}~\cite{NeRF} have shown promising results in novel view synthesis, based on which a number of editing methods~\cite{NeuMesh, NeuTex, EditNeRF, DE-NeRF} are proposed. 
However, NeRF-based methods require dense sampling of camera rays, making rendering less efficient. 
Gaussian splatting~\cite{GS} instead approximates a scene with a set of \yl{3D Gaussians} and visualizes it with a rasterization-based renderer to achieve real-time performance. 
However, despite high-quality and efficient rendering, \yl{Gaussian} splatting still lacks the ability of editing.

In the original \yl{Gaussian} splatting representation, the appearance is approximated by a view-related spherical harmonic function which entangles both texture and lighting information, \yl{making} separate texture and lighting editing impossible. 
\wt{Independent editing requires texture and lighting decomposition so that one part can be manipulated without influencing the other. 
However, different from 3D representations like textured \yl{meshes} or \yl{NeRFs} that have an explicit surface or a continuous implicit field, which can provide a good geometry basis like normal direction for the texture and lighting decomposition, \yl{Gaussian} splatting models a scene's geometry with a discrete set of \yl{Gaussians}, and \yl{this} discrete representation makes the geometry less tractable, posing challenges for later texture and lighting decoupling.}

\begin{figure}[t!]
  	\includegraphics[width=0.99\linewidth]{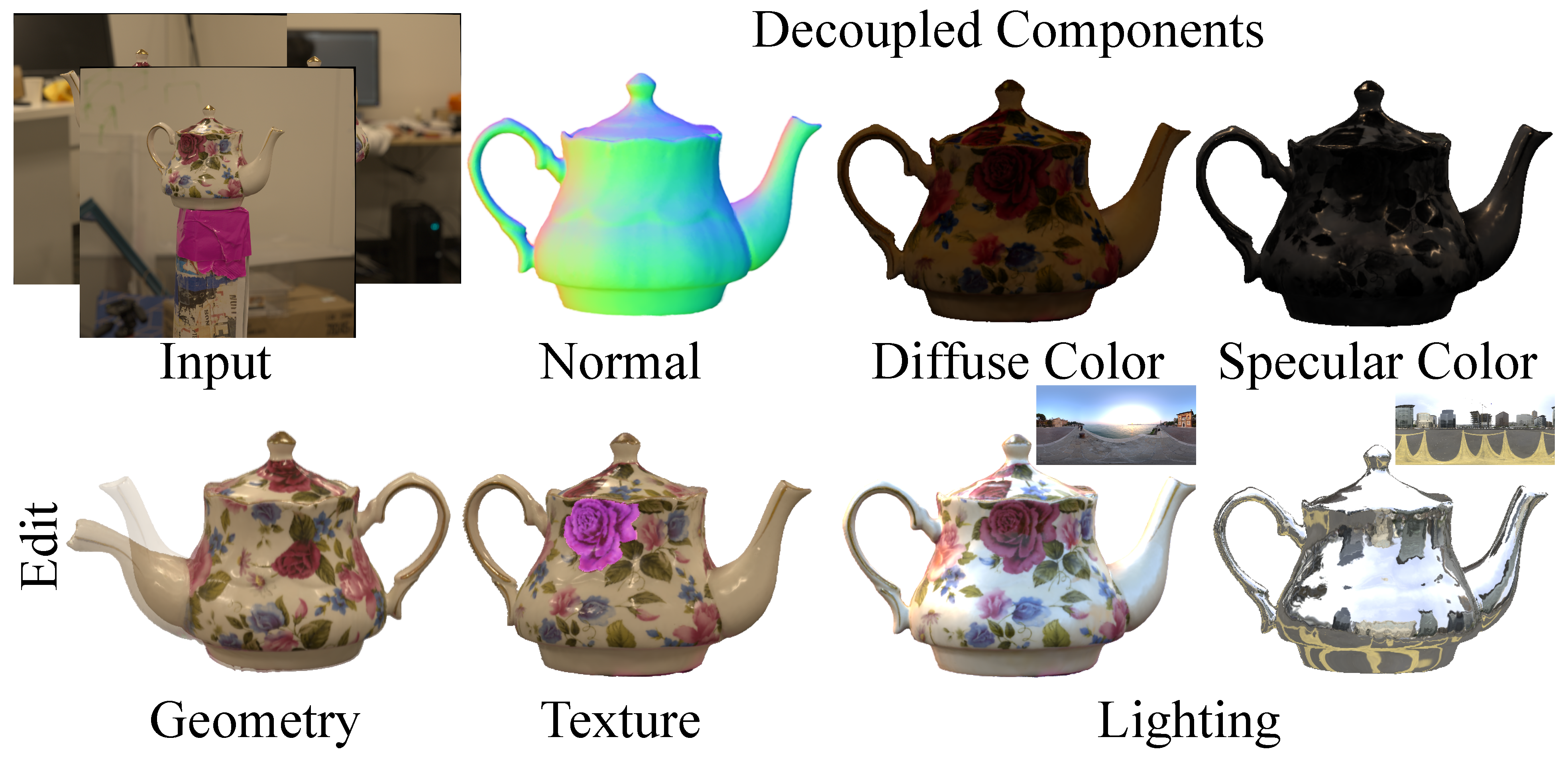}
    \vspace{-1mm}
    \caption{
        Given multi-view images, \name optimizes a \yl{Gaussian} splatting representation with decoupled geometry, texture, and lighting. 
        With this decoupled representation, \name not only allows geometry and texture editing, but \yl{can also} render the input scene (the 3rd column, bottom row) or the edited scene (the last column, bottom row) under a novel illumination. 
    }
    \vspace{-5mm}
    \label{fig:teaser}
\end{figure}

Recently, in the context of \yl{Gaussian} splatting, a few methods have made attempts to edit it. 
\yl{Two independent works both named}
GaussianEditor~\cite{GaussianEditor,GaussianEditor_text} utilize the 2D segmentation model~\cite{SAM} and text-to-image diffusion models~\cite{Diffusion} to enable scene editing like object removal/insertion or text-guided modification similar to Instruct-NeRF2NeRF~\cite{Instruct-NeRF2NeRF}. 
To enable better-controlled editing on texture and lighting, \yl{the works}~\cite{RelightableGaussian, GS-IR, GaussianShader, GIR} try to disentangle texture and lighting to enable individual editing for both diffuse scenes and reflective scenes. 
Although obtaining plausible decomposition results, they use \textit{forward shading} that first calculates shading for each \yl{Gaussian} and then blends shaded \yl{Gaussians}, which can cause blending artifacts when %
the scene is relighted.
This is because the geometry attributes of \yl{Gaussians} like opacity are optimized under the input lighting condition while under novel illuminations shaded \yl{Gaussians} might be blended in a wrong way (\yl{see} 3rd to 6th 
 rows in Fig.~\ref{fig:relight} \yl{for some examples}). 
In addition, they \yl{learn} geometry from the \yl{Gaussian} splatting representation with the normal direction derived from \yl{the} rendered depth map so their geometry may contain \yl{noise} and degenerate on scenes with reflective surfaces (\yl{see} the ``Normal" rows in Fig.~\ref{fig:decom}). 

To resolve the aforementioned issues, we propose \name that decouples a scene's texture and its surrounding illumination of the \yl{Gaussian} splatting representation with \yl{the} \textit{deferred shading} technique. 
Given multi-view captures of an input scene, we define a set of \yl{Gaussians} with extra learnable texture attributes and normal directions attached to model texture and geometry respectively. 
To enable faithful geometry reconstruction, we distill the normal field from a signed distance function \yl{(SDF)} onto the \yl{Gaussians}. 
We also optimize a learnable environment map to approximate the surrounding lighting conditions. 
In the rendering process, geometry and texture buffers are first rasterized and shading is calculated at the pixel level with the \textit{deferred shading} technique to enable faithful decoupling and editing. 
\wt{Despite involving more complex shading computation, \name still allows real-time rendering ($\sim$30FPS at $800 \times 800$ resolution on a single 3090 GPU).}

Our contributions are as follows:
\begin{itemize}
    \item We introduce \name, a decoupled \yl{Gaussian} splatting representation with editable geometry, texture and lighting, where its geometry is enhanced by a normal distillation module. 
    \item \yl{To the best of our knowledge,} \name is the first to apply the \textit{deferred shading} technique to \yl{Gaussian} splatting, which alleviates blending artifacts of previous methods.
    \item Experiments show that \yl{our} \name produces more faithful decomposition and editing results compared to previous methods. 
\end{itemize}

\section{Related Work}

\subsection{Neural Rendering}
With the development of recent implicit \yl{representations}~\cite{OccNet, IMNet, DeepSDF} and volume rendering~\cite{DVR}, neural rendering has become popular in 3D scene reconstruction. 
The \yl{pioneering} work \yl{Neural Radiance Field (NeRF)}~\cite{NeRF} models a sample point's density and view-dependent color with two neural networks and renders the scene by integrating sample points' colors \yl{along the rays}. 
NeRF can synthesize realistic novel \yl{view} results but dense sampling and network query make both training and rendering slow. 
To improve the efficiency, a few \textit{hybrid} rendering methods bake view-independent features onto regular or irregular geometry proxies.  
NSVF~\cite{NSVF} models 3D scenes with \yl{an} octree structure and prunes voxels with low-density values so that sample points outside the octree can be ignored in the rendering process. 
DVGO~\cite{DVGO} stores density values and appearance features on two voxel grids and directly optimizes these two voxel grids to reconstruct 3D scenes. 
TensoRF~\cite{TensoRF} models a scene with a set of matrices and vector pairs that depict the scene's geometry and appearance on corresponding planes, which not only accelerates the training speed but also reduces the storage size. 
Instant-NGP~\cite{InstantNGP} instead utilizes multi-resolution hash encoding as its scene representation and enables training convergence in \yl{as little as} 5 seconds.  
MobileNeRF~\cite{MobileNeRF} and BakedSDF~\cite{BakedSDF} share a similar idea to first reconstruct the mesh of a scene and attach learnable appearance information onto mesh vertices, which enables real-time rendering on consumer devices. 
Moving a step forward, \yl{Gaussian} splatting~\cite{GS} abandons the neural network and models a 3D scene as a set of 3D \yl{Gaussians} on which attributes like position, rotation, scaling, opacity, and spherical harmonic coefficients can be tuned. 
Gaussian splatting uses a rasterization-based renderer which can produce rendered results from a viewpoint in real time.
\wttp{
To better reconstruct reflective scenes, GaussianShader~\cite{GaussianShader} proposes to model view-dependent effects separately similar to RefNeRF~\cite{RefNeRF} and Ye~\textit{et al.}~\cite{ZJUDeferred} further introduce deferred rendering for reflection modeling.
}
Despite the high-quality novel view results, these methods still lack the capability of editing, making appearance change or relighting impossible.

\subsection{Neural Surface Reconstruction}
\yl{Although} NeRF and \yl{Gaussian} splatting can synthesize realistic novel view results, they do not have an explicit surface representation, limiting their application in the traditional graphics pipeline. 
Therefore, a few methods propose to build the connection between the distance field and the density field in volume rendering. 
UniSurf~\cite{UNISURF} models the geometry as an occupancy function and replaces alpha values in volume rendering with occupancy values. 
NeuS~\cite{NeuS} instead treats the geometry as a signed distance field (SDF) and proposes an unbiased and occlusion-aware transformation from SDF to density field. 
VOLSDF~\cite{VolSDF} also uses SDF as its geometry representation but is based on a Laplace distribution's Cumulative Distribution Function. 
To reconstruct open surfaces, several concurrent works~\cite{NeUDF, NeuralUDF, NeAT} introduce the unsigned distance field (UDF) into the training of NeRF. 
However, for shiny scenes, these methods may produce unsatisfactory results like concave surfaces. 
\yl{To address this, the works~}\cite{Factored-NeuS, DE-NeRF, Ref-NeuS} decompose the appearance of a scene into view-independent appearance and view-dependent appearance based on RefNeRF~\cite{RefNeRF} and better reconstruct shiny scenes. 
In the field of \yl{Gaussian} splatting, NeuSG~\cite{NeuSG} introduces a NeuS~\cite{NeuS} branch to align \yl{Gaussians'} normals with corresponding normal directions in NeuS but requires a long training time (16 hours) compared to the original \yl{Gaussian} splatting. 
SuGaR~\cite{SuGaR} instead introduces a depth self-regularization loss term to \yl{constrain Gaussians to} lie on the surface. 

\subsection{Inverse Rendering}
The goal of inverse rendering is to recover geometry, material and lighting information from a set of images of a single scene under the same or different illuminations. 
NeRV~\cite{NeRV} is the pioneering work that decomposes geometry and BRDF (Bidirectional Reflectance Distribution) material under a given lighting condition. 
To reduce dependence on light input, later methods work under unknown light conditions. 
NeRD~\cite{NeRD} and PhySG~\cite{PhySG} use Spherical Gaussian (SG) as the light representation and utilize \yl{a} density field and \yl{a} signed distance field as their geometry representations. 
Based on PhySG, InvRender~\cite{invrender} further takes visibility into consideration to better recover material information in occluded regions. 
NeROIC~\cite{NeROIC} approximates the illumination with Spherical Harmonic \yl{functions} and enables training on Internet images under different lighting conditions. 
NeRFactor~\cite{NeRFactor} represents the lighting with a low-resolution image. 
To better estimate material from images, it also trains a BRDF autoencoder. 
NDR~\cite{NvDiffRec} and NDRMC~\cite{nvdiffrecmc} use a hybrid geometry representation DMTet~\cite{shen2021deep} to enable efficient inverse rendering. 
\wttp{NMF~\cite{NMF} and TensoIR~\cite{TensoIR} introduce the importance sampling strategy for effective shading calculation.}
To model reflective objects, NeRO~\cite{NeRO} and DE-NeRF~\cite{DE-NeRF} express high-frequency lighting as MLP networks. 
Apart from works concentrating on object-level scenes, there are methods~\cite{NeRF-OSR, FEGR, SOL-NeRF} targeting large-scale outdoor scenes.  
There are also concurrent inverse rendering works based on the \yl{Gaussian} splatting representation like \cite{RelightableGaussian, GS-IR, GIR, GaussianShader}. 
Different from them, our method introduces \wt{a normal distillation module to \yl{enable} more accurate normal estimation and \yl{alleviate} \wt{blending} artifacts with the \textit{deferred shading} technique when relighting \yl{Gaussians}.}

\section{Method}
\begin{figure*}[t!]
	\includegraphics[width=0.99\linewidth]{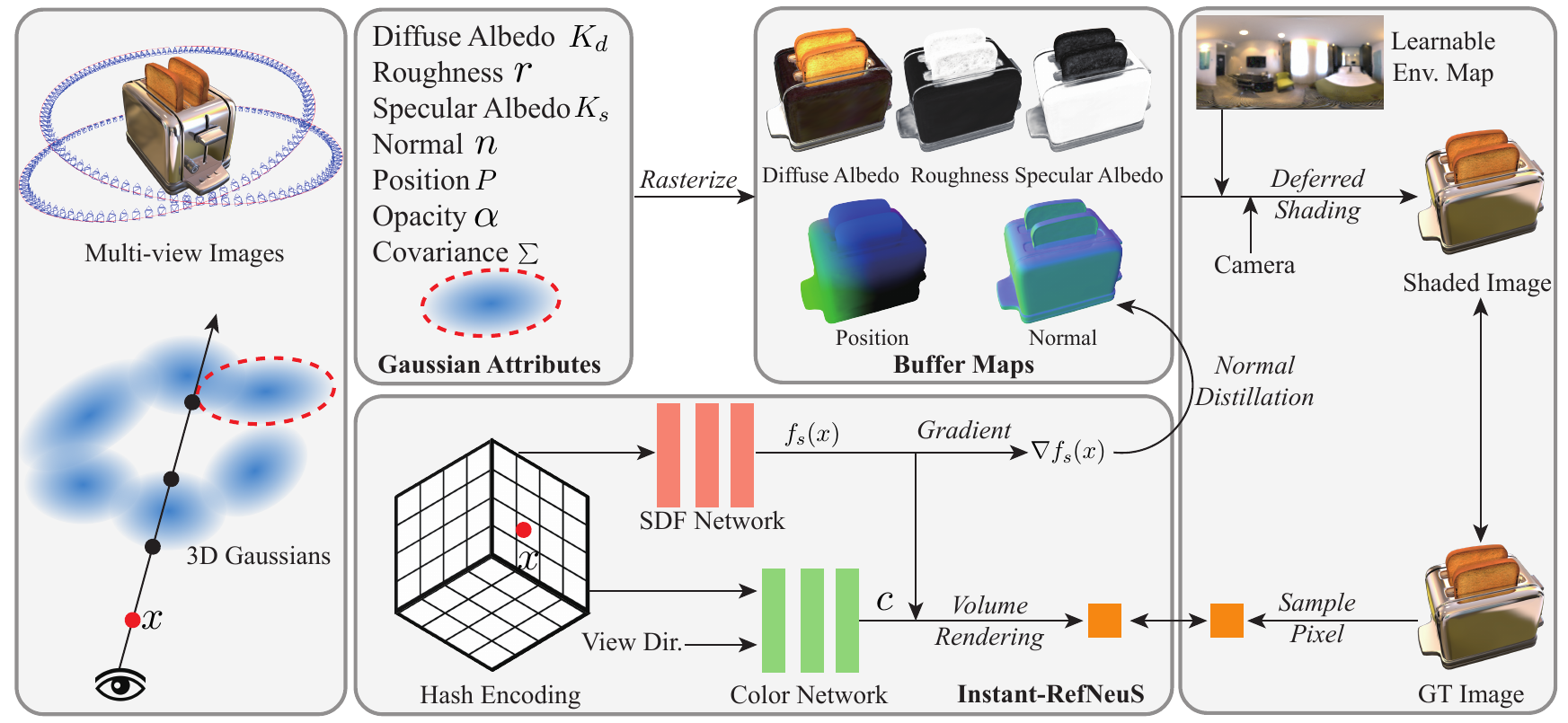}
	\caption{
		Overview of \name. 
        Given multi-view images, \name builds a decoupled \yl{Gaussian} splatting representation where auxiliary texture attributes are defined on each \yl{Gaussian}. 
        To enable successful geometry and appearance separation, we attach an extra normal direction to each \yl{Gaussian} and optimize it by distilling the normal field from a jointly trained signed distance function. 
        For texture and lighting decomposition, \name rasterizes geometry and texture attributes into buffer maps and computes shading at \yl{the} pixel level under the illumination of a learnable environment map with \textit{deferred shading}. 
	}
	\label{fig:pipeline}
\end{figure*}

We propose \name, a decoupled \yl{Gaussian} splatting representation that decomposes geometry, texture, and lighting from multi-view images, which enables downstream applications like scene relighting and texture editing. 
The overview of our method is shown in Fig.~\ref{fig:pipeline}. 
First, we briefly introduce the rendering formulation in \yl{Gaussian} splatting and define the texture parameters we use in rendering (Sec.~\ref{subsec:GSR}).  
Since directly optimizing everything from scratch can lead to suboptimal results, especially on the geometry side, we introduce a normal field distillation module (Sec.~\ref{subsec:NFD}) that utilizes an SDF network to guide \yl{Gaussians} to a better surface representation. 
With geometry set up properly, we bind texture parameters onto the \yl{Gaussian} splatting representation and perform \textit{deferred shading} to obtain plausible decomposition (Sec.~\ref{subsec:DS_GS}). 
Finally, we demonstrate how this decomposition enables scene editing applications like relighting and texture editing (Sec.~\ref{subsec:Edit}). 

\subsection{Gaussian Splatting and Texture Definition}
\label{subsec:GSR}
Gaussian splatting models a 3D scene by a set of \yl{Gaussians} with which attributes like position $P$, 
covariance $\Sigma$, 
scale $S$, opacity $\alpha$ and Spherical Harmonic coefficients (SH) representing view-dependent appearance are associated. 
It uses a point-based $\alpha$-blending rendering formulation similar to NeRF~\cite{NeRF}. 
\begin{equation}
\label{eqn:GSR}
C_* = \sum_{i=1}^N *_i \alpha_i T_i 
\end{equation}
where $N$ is the number of sample points on a ray. 
$T_i = \prod_{j=1}^{i-1} (1-\alpha_j)$ is accumulated transmittance and $*$ can be any attribute of the sample point to be blended such \yl{as} depth, color and normal direction. 
Gaussian splatting avoids densely sampling on the ray and efficiently projects 3D \yl{Gaussians} onto a 2D image plane and approximates 3D scenes by blending view-dependent color represented by SH. 
Instead of modeling the appearance as \yl{a} whole with SH, \yl{our} \name replaces SH with optimizable texture parameters including diffuse albedo $k_d$, roughness $r$, and specular albedo $k_s$ as shown in Fig.~\ref{fig:pipeline} to enable texture and lighting decoupling. 

\subsection{Normal Field Distillation}
\label{subsec:NFD}

Normal estimation is important for extracting correct texture and lighting information. 
However, it is challenging to obtain plausible surface \yl{normals} in \yl{Gaussian} splatting as mentioned in Sec.~\ref{sec:intro} since it is a discrete representation and only promotes rendering quality without constraining the geometry. 
To resolve this issue, we propose a normal field distillation module that integrates current multi-view geometry reconstruction techniques with \yl{Gaussian} splatting. 
Specifically, we jointly train a NeRF-like network equipped with multi-level hash encoding~\cite{InstantNGP} namely \neusbranch 
and a \yl{Gaussian} splatting representation. 
For the \neusbranch, we utilize the signed distance field (SDF) as the geometry representation which can be parameterized as a neural network $f_s(x)$ where $x$ is a sample point in 3D space. 
For the color branch, instead of modeling all the appearances with a single view-dependent network like NeRF~\cite{NeRF}, we separate the appearance into a view-independent branch and a view-dependent branch to avoid geometric artifacts like concave surfaces on reflective scenes~\cite{Ref-NeuS, Factored-NeuS}. 
The view-independent branch predicts the diffuse color $c_d$ and the specular tint $p$ for a sample point $x$. 
The view-dependent branch takes the view direction as input and outputs the specular lighting $c_l$. 
The final color can be composed by $c' = c_d + p c_l$. 
For better understanding, all terms with superscript $'$ refer to the NeRF network while terms without refer to the Gaussian splatting representation. 
To render a pixel $C_c'(v)$, we integrate colors of sample points on a ray $v$ with Eqn.~(\ref{eqn:GSR}).
Adopting the occlusion-aware and unbiased volume rendering technique from NeuS~\cite{NeuS}, the SDF value can be transformed to volume density by $\sigma(t) = \max\left(-\frac{\frac{d\Phi_s}{dt}(f(x(t))}{\Phi_s(f(x(t))}, 0\right)$, where $\mathit{\Phi_s(x)=(1+e^{-sx})^{-1}}$ and $\mathit{s}$ is a trainable deviation parameter.
The opaque value for point $x_i$ is derived from the density function $\sigma(t)$ by $\alpha_i = 1 - exp\left(-\int_{t_i}^{t_{i+1}} \sigma(t) dt\right)$. 
The loss function for \neusbranch is as follows:
\begin{equation}
    \label{eqn:loss_refneus}
    L_{nerf} = \sum_{v {\in V}} \left\|C_c'(v) - C^{t}(v) \right\| + \lambda \sum_{v {\in V}} \sum_{i{=1}}^{M} \left\| \|\nabla_{x_{v, i}} \| - 1 \right\|_2^2,
\end{equation}
where $V$ is the number of rays in a batch. 
$M$ is the number of sample points on a single ray and $C^{t}(v)$ is the corresponding ground truth color. 
The second term is the Eikonal loss, where $\|\nabla f_s({x_{v, i})} \|$ is the spatial norm of the SDF network $f_s(x)$'s gradient at \yl{the} $i$th sample point $x_{v, i}$ on the ray $v$.
In order to distill the normal information from \neusbranch to \yl{Gaussian} splatting, we define an extra normal direction $n$ on each \yl{Gaussian} as shown in Fig.~\ref{fig:pipeline} and render the normal $C_n$ of Gaussian splatting for a pixel using Eqn.~(\ref{eqn:GSR}). 
Then we distill the normal field \yl{Gaussian} splatting by minimizing the difference between $C_n$ and the corresponding rendered normal in \neusbranch:
\begin{equation}
\label{eqn:L_nd}
\begin{aligned}
     L_{nd} = \sum_{v \in V} \left\|1 - C_n(v) \cdot C_n'(v)\right\| = \sum_{v \in V} \left\|1 - C_n(v) \cdot C_{\nabla x_{v,i}} \right\|
\end{aligned}
\end{equation}
where $\cdot$ is the dot product operator. 

\subsection{Deferred Shading in Gaussian Splatting}
\label{subsec:DS_GS}
For shading calculation, \yl{we} follow the rendering equation~\cite{RenderingEquation}:
\begin{equation}
\label{eqn:rendering_eqn}
    L_o(\omega_{o}) = \int_{\Omega} L_i(\omega_{i}) f(\omega_{i}, \omega_{o}) (\omega_{i} \cdot n) d \omega_i
\end{equation}
where $L_o(\omega_{o})$ and $L_i(\omega_{i})$ represent outgoing lighting in direction $\omega_{o}$ and incident lighting from direction $\omega_{i}$ respectively. 
$f(\omega_{i}, \omega_{o})$ is the BRDF 
at a point which can be parameterized by the Disney shading model~\cite{DisneyBRDF}:
\begin{equation}
    f(\omega_{i}, \omega_{o}) = \frac{k_d}{\pi} + \frac{DFG}{4(\omega_{i} \cdot n)(\omega_{o} \cdot n)}
\end{equation}
where $D, F, G$ correspond to normal distribution function, \yl{Fresnel} term, and geometry term. 
Their detailed calculations can be found in~\cite{microfacet}. 

We model the scene's environment lighting with a $6 \times 512\times 512$ High Dynamic Range (HDR) cube map and utilize the SplitSum~\cite{SplitSum} approximation which separates the integrals of lighting and BRDF to enable efficient shading calculation. 
T   he diffuse color $c_{diff}$:
\begin{equation}
    c_{diff} = \frac{k_d}{\pi} \int_{\Omega} L_i(\omega_{i}) (\omega_{i} \cdot n) d \omega_i
\end{equation}
which is only dependent on the normal direction $n$ and can be pre-computed and stored in a 2D texture. 

For the specular color $c_{spec}$:
\begin{equation}
\begin{aligned}
    &c_{spec} \approx Int_{light} \cdot Int_{BRDF} \\
            &= \int_{\Omega}  L_i(\omega_{i}) D(\omega_i, \omega_o) (\omega_{i} \cdot n) d \omega_i \cdot
    \int_{\Omega} f(\omega_{i}, \omega_{o}) (\omega_{i} \cdot n) d \omega_i
\end{aligned}
\end{equation}
where $Int_{light}$ represents the integral of incident light with the normal distribution function $D$. 
Given a fixed environment map, $Int_{light}$ is only related to the roughness value $r$ and therefore can be pre-computed and stored in a mipmap where each mip level corresponds to a fixed roughness value. 
$Int_{BRDF}$ is the integral of specular BRDF under a uniform white environment lighting. 
It is determined by the roughness value in the BRDF and the dot product between incident light direction and normal direction $\omega_{i} \cdot n$. 
Again it can be pre-computed and stored in a 2D texture and efficiently looked up in rendering. 
The final shaded color is $c_g = c_{diff} + c_{spec}$. 

A straightforward way to obtain a decoupled \yl{Gaussian} splatting representation is performing \textit{forward shading} for each \yl{Gaussian} to get its color $c_i$ and rasterizing \yl{Gaussians} with Eqn.~(\ref{eqn:GSR}) similar to previous works~\cite{invrender, NeRFactor, GIR, GaussianShader}. 
Though obtaining plausible decoupled results, we observe notable artifacts when \yl{rendering} the decoupled \yl{Gaussians} under a novel lighting condition (see ``F.S." columns in Fig.~\ref{fig:ablate_DS} and ``RGS" and ``GShader" columns in Fig.~\ref{fig:relight}). 
This is because even \yl{though} Eqn.~(\ref{eqn:L_nd}) constrains the \yl{rasterized} normal to be close to that of \neusbranch at pixel level, the \yl{rasterized} normal is still blended from multiple \yl{Gaussians}' normals. 
The geometry attributes like position, covariance and opacity may overfit the given lighting condition to produce proper blended normal and shaded color and it does not guarantee that each \yl{Gaussian}'s individual normal or shaded color is correct. 
Thus when changing the lighting condition, the trained geometry attributes may not work for the shaded colors under a new lighting condition and blend them in \yl{a} wrong way.  

To resolve this issue, we introduce the \textit{deferred shading} technique into the rendering of \yl{Gaussians}. 
That is to say, we first rasterize components like position $P$, normal $n$, diffuse albedo $k_d$, roughness $r$, specular albedo $k_s$, and opacity $\alpha$ into 2D pixels $C_{P}, C_{n}, C_{k_d}, C_{r}, C_{k_s}$, and $C_{\alpha}$ with Eqn.~(\ref{eqn:GSR}). 
By aggregating all pixels, we get corresponding 2D maps $I_{P}, I_{n}, I_{k_d}$, $I_{r}, I_{k_s}$, and $I_{\alpha}$. 
Then we calculate the shaded color $C$ for a pixel with Eqn.~(\ref{eqn:rendering_eqn}). 
Note that since computing shading for every single pixel in the image is time-consuming, we only perform shading on pixels with $C_{\alpha}$ bigger than 0.5. 
Overall, we optimize \yl{the} following loss:
\begin{equation}
\label{eqn:total_loss}
\begin{aligned}
    L = &L_{nerf} + \lambda_{nd} L_{nd} + \lambda_{L1}L_{L1} + \lambda_{ssim}L_{ssim} +\\
    & \lambda_{mask}L_{mask} + \lambda_{TV} L_{TV}
\end{aligned}
\end{equation}
where $L_{L1}$ and $L_{ssim}$ are the L1 loss and D-SSIM loss between the rasterized image and the ground truth image same as those in~\cite{GS}. 
$L_{mask} = \|I_{\alpha} - I_m\|_1$ is the mask loss that encourages \yl{Gaussians} to lie in the foreground and $I_m$ is the ground truth mask image. 
$L_{TV}$ is the total variation loss applied on images $I_{k_d}, I_{k_s}, I_{k_s}$ to encourage smooth texture estimation.

\subsection{Editing \yl{Decoupled} Gaussians}
\label{subsec:Edit}
After optimizing the attributes on \yl{Gaussians}, we obtain a decoupled \yl{Gaussian} splatting representation and can render the original scene with edited texture or geometry under a novel illumination. 
\subsection{Relighting}
As we extract the environment map of the original scene, we can replace it by a novel environment map \yl{with} little effort and render the scene with novel illumination using \textit{deferred shading}. 

\subsection{Geometry Editing}
For geometry editing, instead of directly operating on \yl{Gaussians}, we first extract a mesh with \neusbranch and use it as a geometry proxy to guide the deformation of \yl{Gaussians}. 
Specifically, we first deform the mesh using the \yl{As-Rigid-As-Possible (ARAP)} deformation algorithm~\cite{ARAP} which solves \yl{for} rotations and translations of mesh vertices.  
Then we find the closest neighbor 
for each \yl{Gaussian} on the original mesh \yl{surface}, whose rotation and translation can be obtained via barycentric mapping \yl{(based on the mesh triangle that contains the neighboring point)}. 
Finally, we deform each \yl{Gaussian} by applying the rotation and the translation of its corresponding closest neighbor to \yl{Gaussian}'s position, covariance matrix, and normal direction.
By rasterizing deformed \yl{Gaussians}, we can render the deformed scene. 

\subsection{Texture Editing}
\label{sec:texture_edit}
For texture editing, we follow the editing paradigm in traditional 3D modeling software~\cite{blender} to paint the scene from a given viewpoint. 
Our texture editing supports changes on all texture components. 
We first determine which Gaussians need to be adjusted by considering both whether they lie in the editing mask and the difference between the \yl{Gaussian} depth and the corresponding depth in \yl{the} depth map rendered with Eqn.~(\ref{eqn:GSR}). 
Then we can optimize these \yl{Gaussians'} texture attributes $*$ by:
\begin{equation}
\label{eqn:edit_naive}
    {\arg}\min_{*} \|I_{*}^i - I_e^i\|,\space\space\space * \in \{k_d, r, k_s\}
\end{equation}
where $I_{*}^i$ denotes a rendered texture map, \yl{e.g.} diffuse albedo map, from the input editing viewpoint $i$. 
$I_e^i$ is the edited texture map from viewpoint $i$ by putting edited \yl{content} onto $I_{*}^i$. 

However, this naive solution only considers the input editing viewpoint and may produce blending artifacts when we view the optimized \yl{Gaussians} from other viewpoints. 
Therefore, we propose to augment the optimization with random view inputs. 
That is to say, we generate edited rendered results from multiple random viewpoints by first \yl{unprojecting the} edited \yl{content} onto the mesh extracted by \neusbranch to form a textured mesh and \yl{rendering} the textured mesh from random viewpoints. 
So the final editing loss is: 
\begin{equation}
\label{eqn:edit_improved}
    {\arg}\min_{*} \sum_{j \in L}\|I_{*}^j - I_e^j\|,\space\space\space * \in \{k_d, r, k_s\}
\end{equation}
where set $L$ includes both the input viewpoint and randomly sampled viewpoints.

\begin{figure*}[!t]
    \centering
    {
        \newlength\mytmplennvs
        \setlength\mytmplennvs{.11\linewidth}
        \setlength\tabcolsep{0pt}
        \begin{tabular}{ccccccccc}
            \hspace{-4mm}
            \redbox{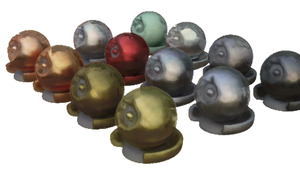}{(0.8,0.1)}{(1.2,0.5)}{\mytmplennvs}&
            \redbox{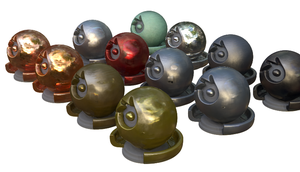}{(0.8,0.1)}{(1.2,0.5)}{\mytmplennvs}&
            \redbox{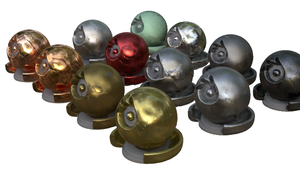}{(0.8,0.1)}{(1.2,0.5)}{\mytmplennvs}&
            \redbox{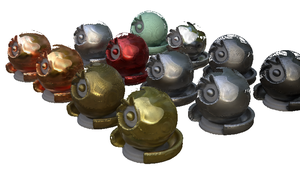}{(0.8,0.1)}{(1.2,0.5)}{\mytmplennvs}&
            \redbox{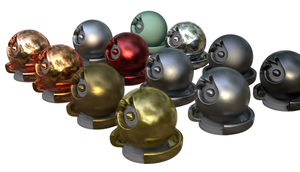}{(0.8,0.1)}{(1.2,0.5)}{\mytmplennvs}&
            \redbox{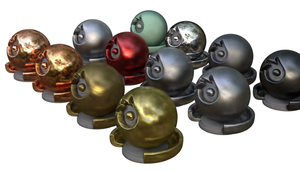}{(0.8,0.1)}{(1.2,0.5)}{\mytmplennvs}&
            \redbox{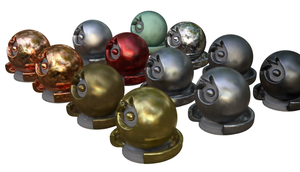}{(0.8,0.1)}{(1.2,0.5)}{\mytmplennvs}&
            \redbox{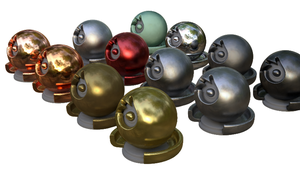}{(0.8,0.1)}{(1.2,0.5)}{\mytmplennvs}&
            \redbox{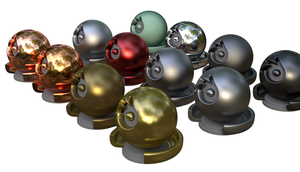}{(0.8,0.1)}{(1.2,0.5)}{\mytmplennvs}
            \\
            \hspace{-4mm}
            \redbox{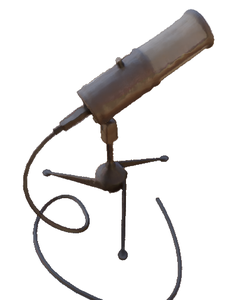}{(0.8,1.6)}{(1.2,2)}{\mytmplennvs}&
            \redbox{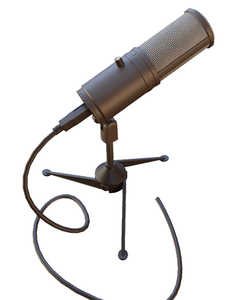}{(0.8,1.6)}{(1.2,2)}{\mytmplennvs}&
            \redbox{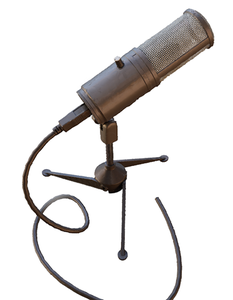}{(0.8,1.6)}{(1.2,2)}{\mytmplennvs}&
            \redbox{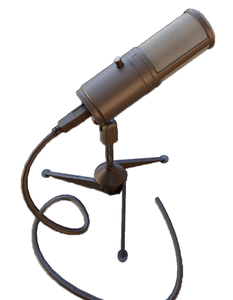}{(0.8,1.6)}{(1.2,2)}{\mytmplennvs}&
            \redbox{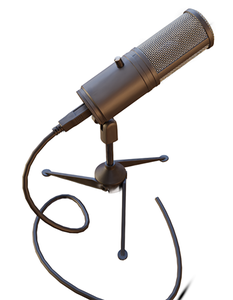}{(0.8,1.6)}{(1.2,2)}{\mytmplennvs}&
            \redbox{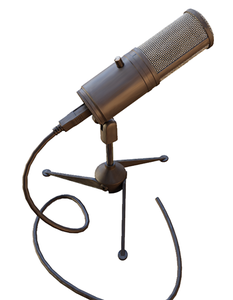}{(0.8,1.6)}{(1.2,2)}{\mytmplennvs}&
            \redbox{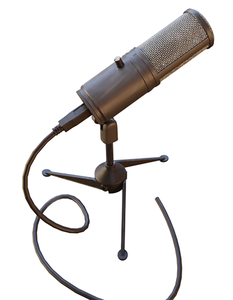}{(0.8,1.6)}{(1.2,2)}{\mytmplennvs}&
            \redbox{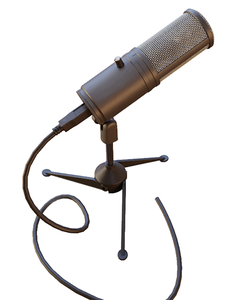}{(0.8,1.6)}{(1.2,2)}{\mytmplennvs}&
            \redbox{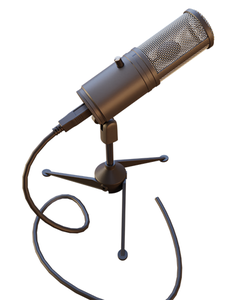}{(0.8,1.6)}{(1.2,2)}{\mytmplennvs}
            \\
            \hspace{-4mm}
            \redbox{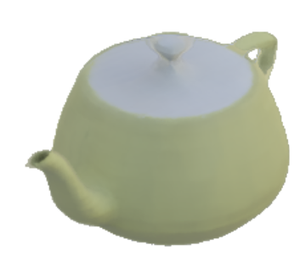}{(1,0.5)}{(1.5,1)}{\mytmplennvs}&
            \redbox{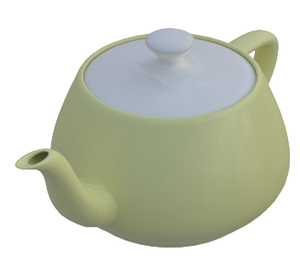}{(1,0.5)}{(1.5,1)}{\mytmplennvs}&
            \redbox{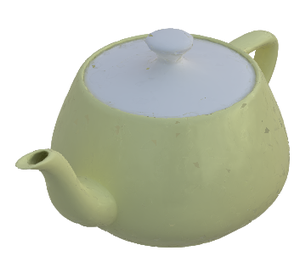}{(1,0.5)}{(1.5,1)}{\mytmplennvs}&
            \redbox{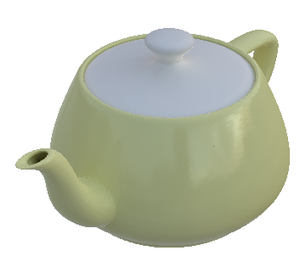}{(1,0.5)}{(1.5,1)}{\mytmplennvs}&
            \redbox{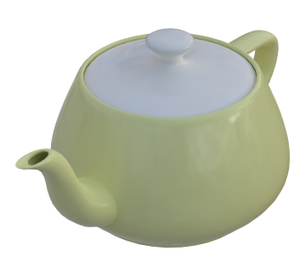}{(1,0.5)}{(1.5,1)}{\mytmplennvs}&
            \redbox{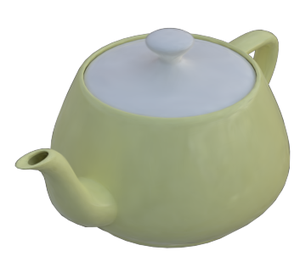}{(1,0.5)}{(1.5,1)}{\mytmplennvs}&
            \redbox{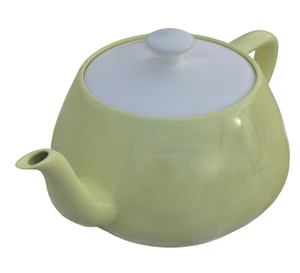}{(1,0.5)}{(1.5,1)}{\mytmplennvs}&
            \redbox{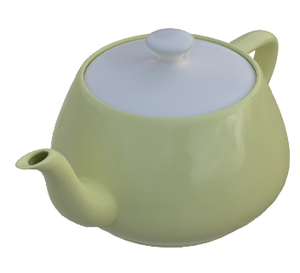}{(1,0.5)}{(1.5,1)}{\mytmplennvs}&
            \redbox{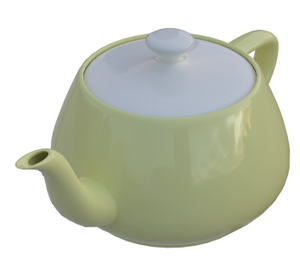}{(1,0.5)}{(1.5,1)}{\mytmplennvs}
            \\
             \hspace{-4mm}
            \redbox{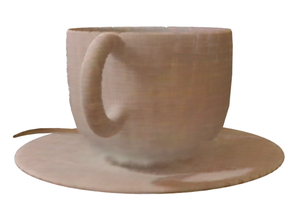}{(0.7,0.4)}{(1.5,1.2)}{\mytmplennvs}&
            \redbox{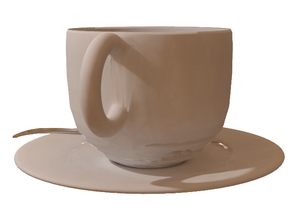}{(0.7,0.4)}{(1.5,1.2)}{\mytmplennvs}&
            \redbox{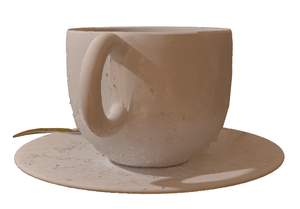}{(0.7,0.4)}{(1.5,1.2)}{\mytmplennvs}&
            \redbox{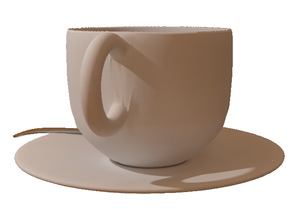}{(0.7,0.4)}{(1.5,1.2)}{\mytmplennvs}&
            \redbox{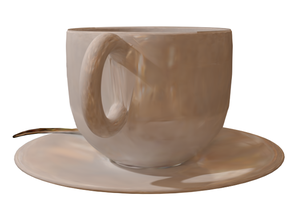}{(0.7,0.4)}{(1.5,1.2)}{\mytmplennvs}&
            \redbox{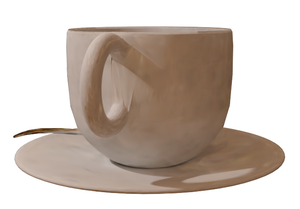}{(0.7,0.4)}{(1.5,1.2)}{\mytmplennvs}&
            \redbox{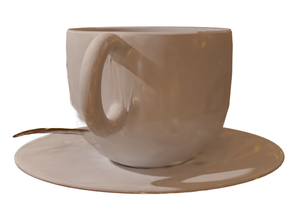}{(0.7,0.4)}{(1.5,1.2)}{\mytmplennvs}&
            \redbox{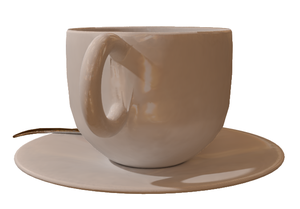}{(0.7,0.4)}{(1.5,1.2)}{\mytmplennvs}&
            \redbox{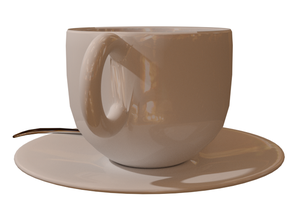}{(0.7,0.4)}{(1.5,1.2)}{\mytmplennvs}
            \\
            \hspace{-4mm}
            \redbox{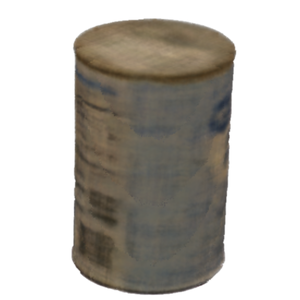}{(0.5,2)}{(1.6,1.2)}{\mytmplennvs}&
            \redbox{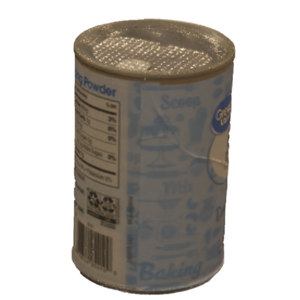}{(0.5,2)}{(1.6,1.2)}{\mytmplennvs}&
            \redbox{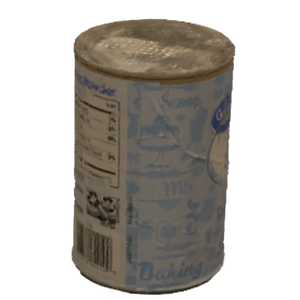}{(0.5,2)}{(1.6,1.2)}{\mytmplennvs}&
            \redbox{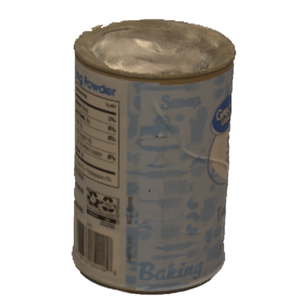}{(0.5,2)}{(1.6,1.2)}{\mytmplennvs}&
            \redbox{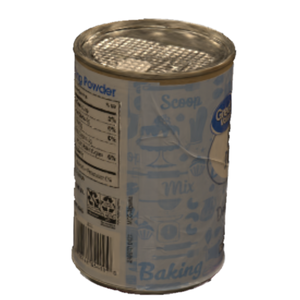}{(0.5,2)}{(1.6,1.2)}{\mytmplennvs}&
            \redbox{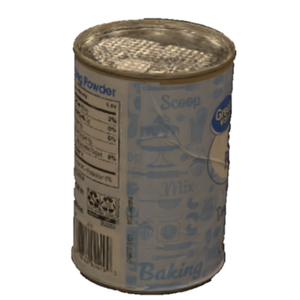}{(0.5,2)}{(1.6,1.2)}{\mytmplennvs}&
            \redbox{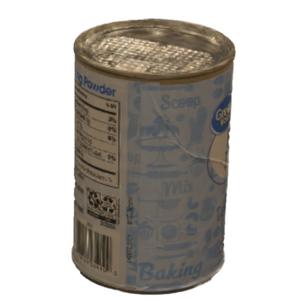}{(0.5,2)}{(1.6,1.2)}{\mytmplennvs}&
            \redbox{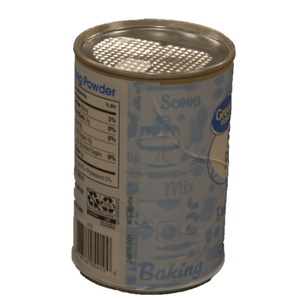}{(0.5,2)}{(1.6,1.2)}{\mytmplennvs}&
            \redbox{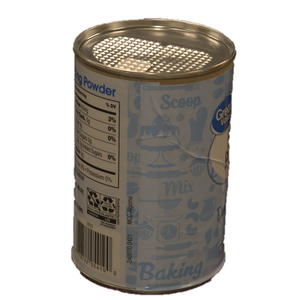}{(0.5,2)}{(1.6,1.2)}{\mytmplennvs}
            \\
            \hspace{-4mm}
            \redbox{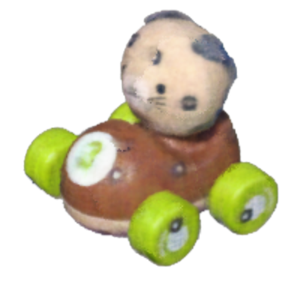}{(0.85,0.2)}{(1.9,1)}{\mytmplennvs}&
            \redbox{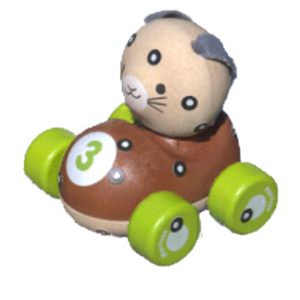}{(0.85,0.2)}{(1.9,1)}{\mytmplennvs}&
            \redbox{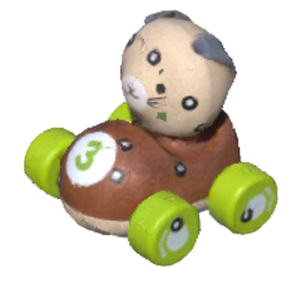}{(0.85,0.2)}{(1.9,1)}{\mytmplennvs}&
            \redbox{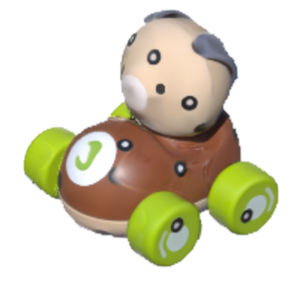}{(0.85,0.2)}{(1.9,1)}{\mytmplennvs}&
            \redbox{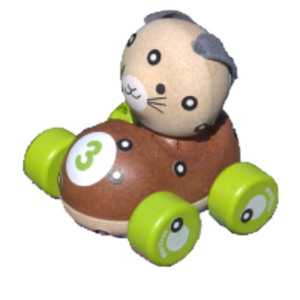}{(0.85,0.2)}{(1.9,1)}{\mytmplennvs}&
            \redbox{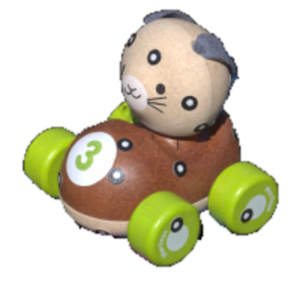}{(0.85,0.2)}{(1.9,1)}{\mytmplennvs}&
            \redbox{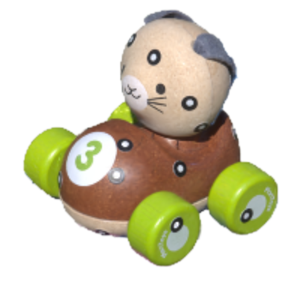}{(0.85,0.2)}{(1.9,1)}{\mytmplennvs}&
            \redbox{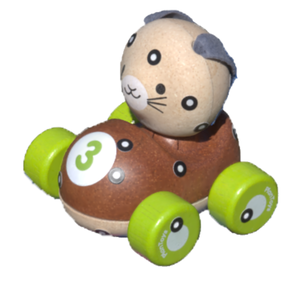}{(0.85,0.2)}{(1.9,1)}{\mytmplennvs}&
            \redbox{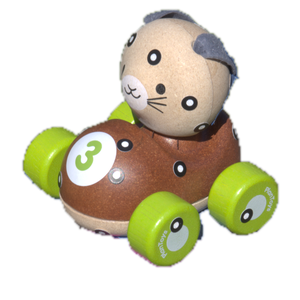}{(0.85,0.2)}{(1.9,1)}{\mytmplennvs}
            \\
            \hspace{-4mm}
            NeRFactor&
            \wttp{TensoIR}&
            NDR&
            NeRO&
            GS&
            RGS&
            GShader&
            Ours&
            GT
        \end{tabular}
    }
    \caption{Novel view synthesis comparisons with NeRFactor~\cite{NeRFactor}, \wttp{TensoIR~\cite{TensoIR}}, NDR~\cite{NvDiffRec}, NeRO~\cite{NeRO}, Gaussian Splatting (GS)~\cite{GS}, RelightableGaussian (RGS)~\cite{RelightableGaussian}, and GaussianShader (GShader)~\cite{GaussianShader}. 
    The bottom two rows are from the real Stanford ORB dataset~\cite{Stanford-ORB}.
    }
    \label{fig:nvs}
\end{figure*}

\begin{table*}[!t]
	\caption{
		\wttp{Quantitative comparison of novel view synthesis results using SSIM, PSNR and LPIPS metrics on NeRF Synthetic, Shiny Blender and Stanford ORB datasets.}
	}
		\centering
				\begin{tabular*}{\linewidth}{@{\extracolsep{\fill}} cccccccccc }

\hline
				    \multirow{2}[2]{*}{Methods} &                           \multicolumn{3}{c}{\tabincell{c}{NeRF Synthetic}}                            &                            \multicolumn{3}{c}{\tabincell{c}{Shiny Blender}}                            &  \multicolumn{3}{c}{\tabincell{c}{Stanford ORB}} \\ 
                    \cmidrule{2-10}             & \tabincell{c}{PSNR $^\uparrow$} & \tabincell{c}{SSIM $^\uparrow$} & \tabincell{c}{LPIPS $^\downarrow$} & \tabincell{c}{PSNR $^\uparrow$} & \tabincell{c}{SSIM $^\uparrow$} & \tabincell{c}{LPIPS $^\downarrow$} & \tabincell{c}{PSNR $^\uparrow$} & \tabincell{c}{SSIM $^\uparrow$} & \tabincell{c}{LPIPS $^\downarrow$}  \\ \hline

				    NeRFactor                   & 27.86                           & 0.924                           & 0.044                              & 27.04                           & 0.913                           & 0.123                              & 30.01 & 0.934 & 0.079\\
				    TensoIR                     & 29.52                           & \tbc{0.944}                     & 0.050                              & 28.01                           & 0.905                           & 0.131                              & 34.16 & 0.980 & 0.020\\
				    NDR                         & 29.05                           & 0.939                           & 0.081                              & 28.11                           & 0.935                           & 0.076                              & 32.84 & 0.985 & 0.014\\
				    NeRO                        & 29.03                           & 0.942                           & 0.051                              & \sbc{30.96}                     & \sbc{0.953}                     & \sbc{0.045}                        & 29.25 & 0.970 & 0.019\\
				    GS                          & \bc{33.30}                      & \sbc{0.969}                     & \bc{0.030}                         & 30.35                           & 0.944                           & 0.088                              & \tbc{35.77} & \tbc{0.991} & \tbc{0.007}\\
				    RGS                         & 29.25                           & 0.941                           & \tbc{0.042}                        & 28.69                           & 0.930                           & 0.081                              & \sbc{36.65} & \sbc{0.992} & \bc{0.006}\\
				    GShader                     & \tbc{30.84}                     & 0.933                           & 0.067                              & \tbc{30.72}                     & \tbc{0.951}                     & \tbc{0.069}                        & 35.43 & 0.990 & 0.008\\ \hline
				    Ours                        & \sbc{32.29}                     & \bc{0.971}                      & \sbc{0.039}                        & \bc{32.61}                      & \bc{0.968}                      & \bc{0.041}                         & \bc{37.93} & \bc{0.995} & \bc{0.006}\\ %
\hline
				\end{tabular*}
	\label{tab:nvs}
\end{table*}

\section{Results and Evaluations}
\subsection{Datasets and Metrics}
\wttp{We conduct qualitative and quantitative experiments on two synthetic datasets, NeRF Synthetic~\cite{NeRF} and Shiny Blender~\cite{RefNeRF} and one real dataset, the Stanford ORB dataset ~\cite{Stanford-ORB}.
For all datasets, we evaluate \name and baseline methods on novel view synthesis,  decomposition and relighting tasks. 
For quantitative comparisons, we use PSNR (Peak Signal-to-Noise Ratio), SSIM (Structural Similarity)~\cite{SSIM} and LPIPS (Learned Perceptual Image Patch Similarity)~\cite{LPIPS} metrics to compare the similarity between images. 
To evaluate the quality of the estimated normals, we compare the estimated normal images with the ground truth with Mean Squared Error (MSE). 
}

\subsection{Implementation Details}
\wttp{
We train the decoupled Gaussian splatting representation on a single 3090 GPU with 24GB memory. 
The training process takes 3-4 hours depending on the scenes' complexity and the input images' resolution. 
\name allows $\sim$30FPS novel view synthesis at $800 \times 800$ resolution on a single 3090 GPU. 
For hyperparameter settings, in Eqn.~(\ref{eqn:loss_refneus}), $\lambda_{1}$ is set to $0.1$.  
In Eqn.~(\ref{eqn:total_loss}), $\lambda_{nd}, \lambda_{L1}, \lambda_{ssim}, \lambda_{mask} ,\lambda_{TV}$ are set to $0.5, 0.8, 0.2, 0.5, 0.001$ in all our examples. 
}

\subsection{Novel View Synthesis}
We present novel view synthesis results on different cases' test splits in Fig.~\ref{fig:nvs} and compare \yl{our method} with NeRFactor~\cite{NeRFactor}, \wttp{TensoIR}~\cite{TensoIR}, NDR~\cite{NvDiffRec}, NeRO~\cite{NeRO}, Gaussian Splatting~\cite{GS}, RelightableGaussian~\cite{RelightableGaussian} and GuassianShader~\cite{GaussianShader}. 
NeRFactor models the geometry with a density function and utilizes a low-resolution environment map as its lighting representation, which leads to less convincing geometry (first row in Fig.~\ref{fig:nvs}) and blurry reconstruction results. 
\wttp{TensoIR is unable to synthesize appearance details with its smooth lighting representation approximated by Spherical Gaussian functions. }
NDR uses deformable \yl{tetrahedra} to approximate the geometry which may cause irregular geometry and noisy novel view synthesis results. 
\wt{NeRO produces blurry appearance (second and fourth rows) and may reconstruct wrong geometry (first row).}
\yl{Gaussian} Splatting models the texture and lighting with a single spherical harmonic function which may fail to generalize to novel views and produce wrong reflections (last row in Fig.~\ref{fig:nvs}). 
RelightableGaussian and GaussianShader separately model texture and lighting but learn geometry from scratch, which causes blurry reconstruction. 
As for quantitative comparisons, we compare novel view reconstruction results by different methods with the ground truth image using PSNR, SSIM and LPIPS metrics in Table~\ref{tab:nvs}. 
\wt{Our method achieves comparable results with baseline methods on the NeRF synthetic dataset and outperforms other baselines on the shiny blender dataset.}

\begin{figure*}[t!]
    \centering
    {
        \newlength\mytmplen
        \setlength\mytmplen{.11\linewidth}
        \setlength\tabcolsep{0pt}
            \setlength\tabcolsep{0pt}
            \begin{tabular*}{\textwidth}{@{\extracolsep{\fill}} ccccccccc }
                \hspace{-1mm}{\rotatebox{90}{\quad \hspace{4mm} {Normal}}}&
                \hspace{-2mm}\includegraphics[width=0.12\textwidth]{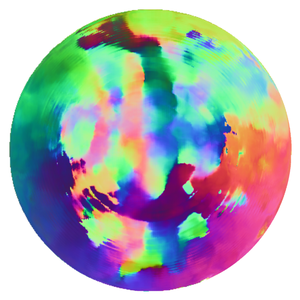}&
                \hspace{-2mm}\includegraphics[width=0.12\textwidth]{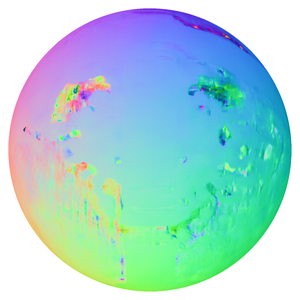}&
                \hspace{-2mm}\includegraphics[width=0.12\textwidth]{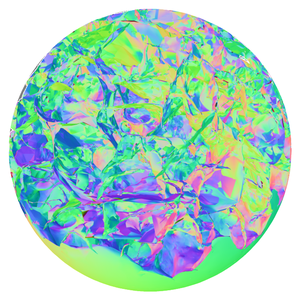}&
                \hspace{-2mm}\includegraphics[width=0.12\textwidth]{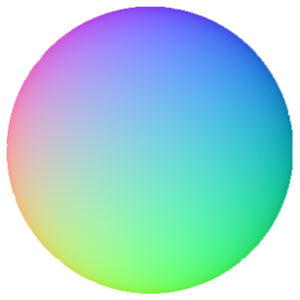}&
                \hspace{-2mm}\includegraphics[width=0.12\textwidth]{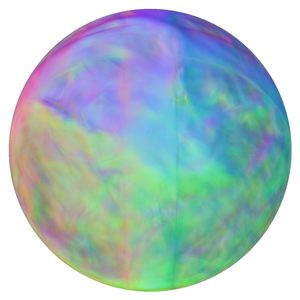}&
                \hspace{-2mm}\includegraphics[width=0.12\textwidth]{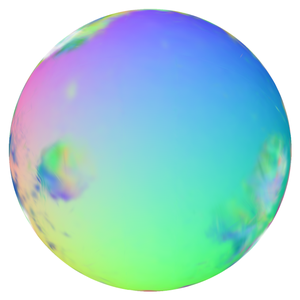}&
                \hspace{-2mm}\includegraphics[width=0.12\textwidth]{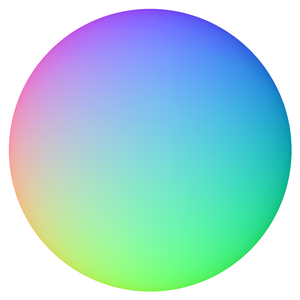}&
                \hspace{-2mm}\includegraphics[width=0.12\textwidth]{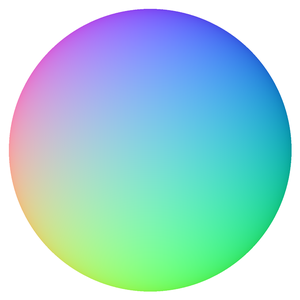}
                \\
                \hspace{-1mm}{\rotatebox{90}{\quad \hspace{4mm} {\makecell{Albedo}}}}&
                \hspace{-2mm}\includegraphics[width=0.12\textwidth]{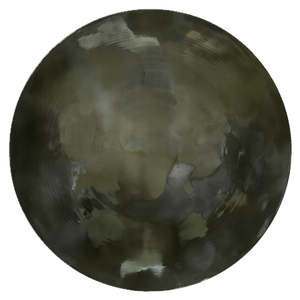}&
                \hspace{-2mm}\includegraphics[width=0.12\textwidth]{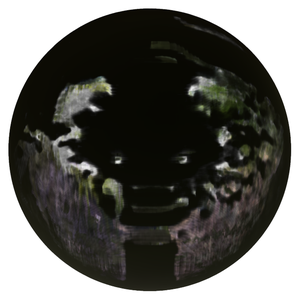}&
                \hspace{-2mm}\includegraphics[width=0.12\textwidth]{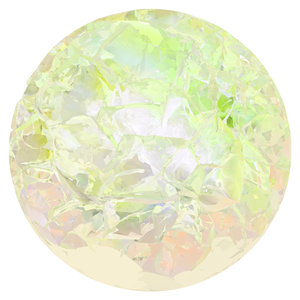}&
                \hspace{-2mm}\includegraphics[width=0.12\textwidth]{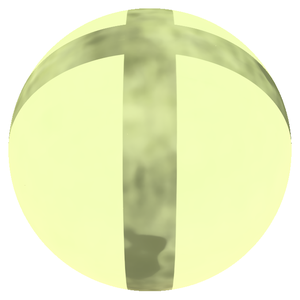}&
                \hspace{-2mm}\includegraphics[width=0.12\textwidth]{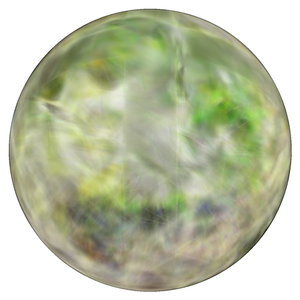}&
                \hspace{-2mm}\includegraphics[width=0.12\textwidth]{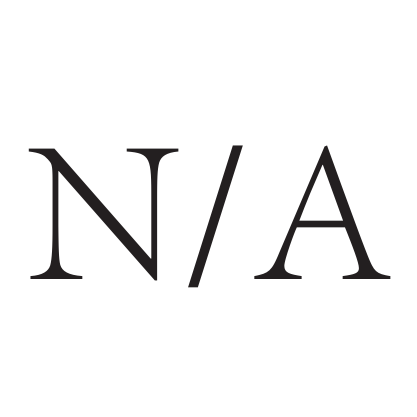}&
                \hspace{-2mm}\includegraphics[width=0.12\textwidth]{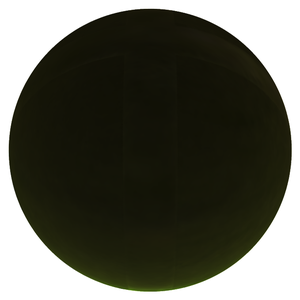}&
                \hspace{-2mm}\includegraphics[width=0.12\textwidth]{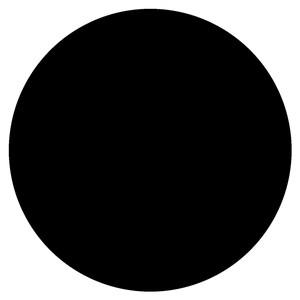}
                \vspace{-2mm}
                \\
                \vspace{-2mm}
                \hspace{-1mm}{\rotatebox{90}{\quad \hspace{4mm} {Normal}}}&
                \hspace{-2mm}\includegraphics[width=0.12\textwidth]{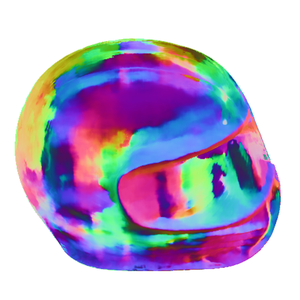}&
                \hspace{-2mm}\includegraphics[width=0.12\textwidth]{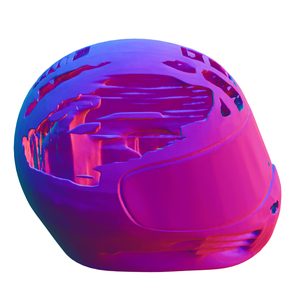}&
                \hspace{-2mm}\includegraphics[width=0.12\textwidth]{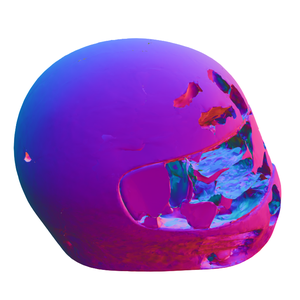}&
                \hspace{-2mm}\includegraphics[width=0.12\textwidth]{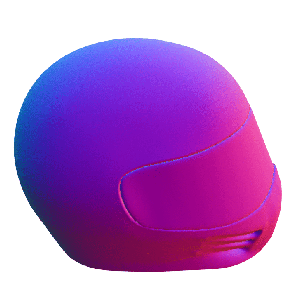}&
                \hspace{-2mm}\includegraphics[width=0.12\textwidth]{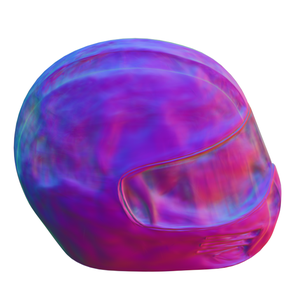}&
                \hspace{-2mm}\includegraphics[width=0.12\textwidth]{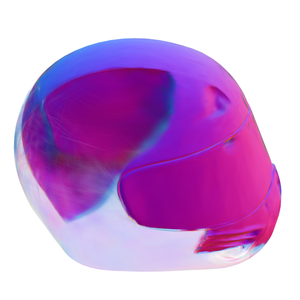}&
                \hspace{-2mm}\includegraphics[width=0.12\textwidth]{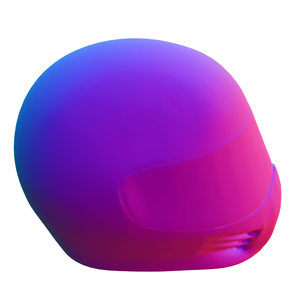}&
                \hspace{-2mm}\includegraphics[width=0.12\textwidth]{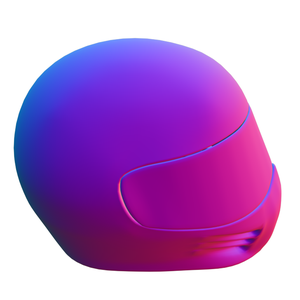}
                \\
                \hspace{-1mm}{\rotatebox{90}{\quad \hspace{4mm} {\makecell{Albedo}}}}&
                \hspace{-2mm}\includegraphics[width=0.12\textwidth]{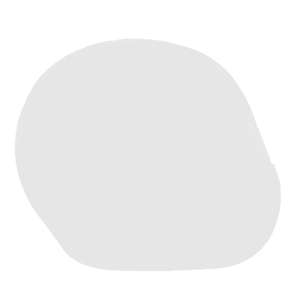}&
                \hspace{-2mm}\includegraphics[width=0.12\textwidth]{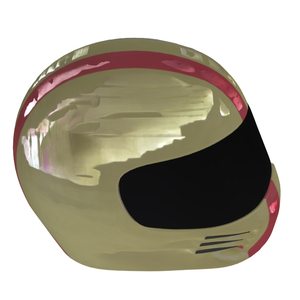}&
                \hspace{-2mm}\includegraphics[width=0.12\textwidth]{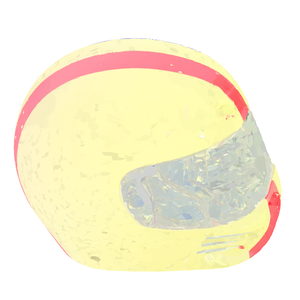}&
                \hspace{-2mm}\includegraphics[width=0.12\textwidth]{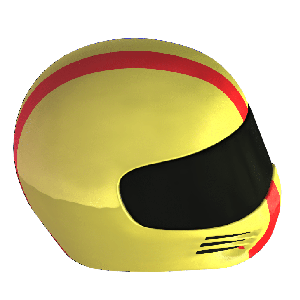}&
                \hspace{-2mm}\includegraphics[width=0.12\textwidth]{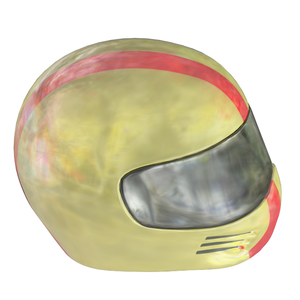}&
                \hspace{-2mm}\includegraphics[width=0.12\textwidth]{./img/na}&
                \hspace{-2mm}\includegraphics[width=0.12\textwidth]{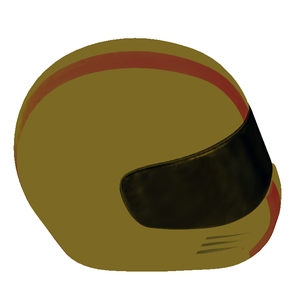}&
                \hspace{-2mm}\includegraphics[width=0.12\textwidth]{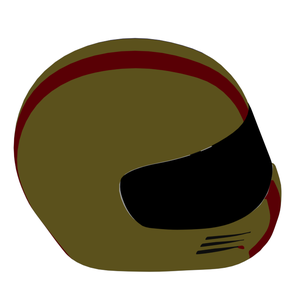}
                \\
                \hspace{-1mm}{\rotatebox{90}{\quad \hspace{0mm} {Normal}}}&
                \zoomin{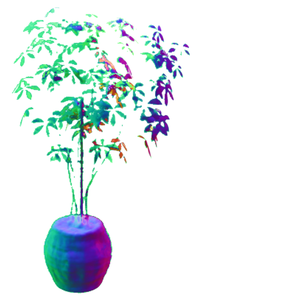}{0.5}{0.25}{1.4}{0.5}{1.15cm}{\mytmplen}{2}{south west}&
                \zoomin{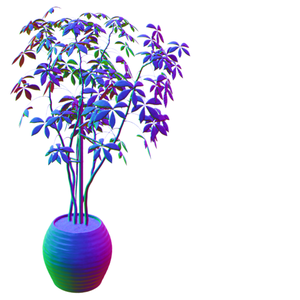}{0.5}{0.25}{1.4}{0.5}{1.15cm}{\mytmplen}{2}{south west}&
                \zoomin{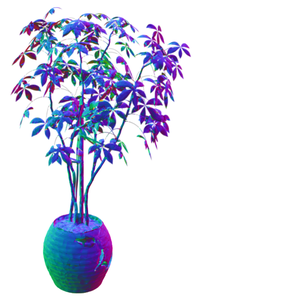}{0.5}{0.25}{1.4}{0.5}{1.15cm}{\mytmplen}{2}{south west}&
                \zoomin{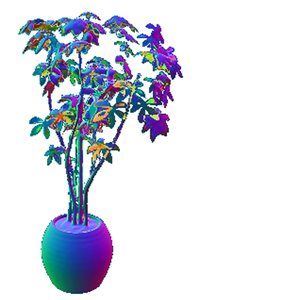}{0.5}{0.25}{1.4}{0.5}{1.15cm}{\mytmplen}{2}{south west}&
                \zoomin{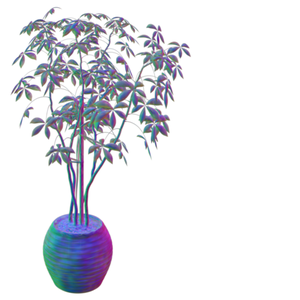}{0.5}{0.25}{1.4}{0.5}{1.15cm}{\mytmplen}{2}{south west}&
                \zoomin{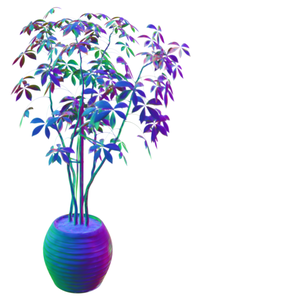}{0.5}{0.25}{1.4}{0.5}{1.15cm}{\mytmplen}{2}{south west}&
                \zoomin{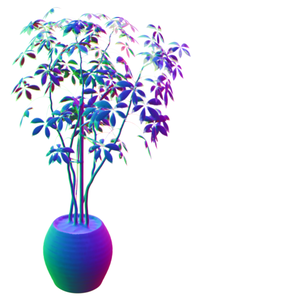}{0.5}{0.25}{1.4}{0.5}{1.15cm}{\mytmplen}{2}{south west}&
                \zoomin{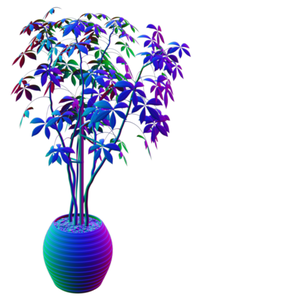}{0.5}{0.25}{1.4}{0.5}{1.15cm}{\mytmplen}{2}{south west}
                \\
                \hspace{-1mm}{\rotatebox{90}{\quad \hspace{0mm} {\makecell{Albedo}}}}&
                \zoomin{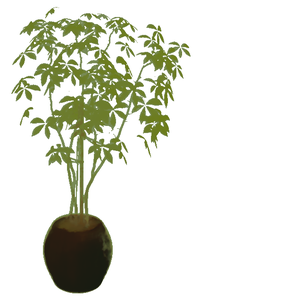}{0.5}{0.3}{1.4}{0.5}{1.15cm}{\mytmplen}{2}{south west}&
                \zoomin{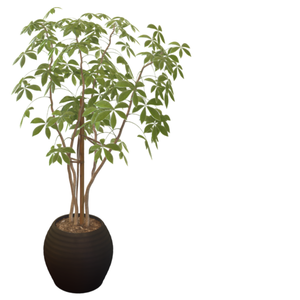}{0.5}{0.3}{1.4}{0.5}{1.15cm}{\mytmplen}{2}{south west}&
                \zoomin{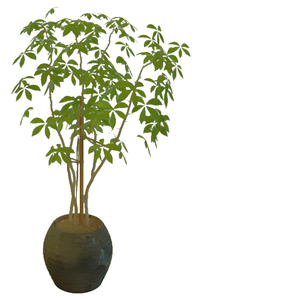}{0.5}{0.3}{1.4}{0.5}{1.15cm}{\mytmplen}{2}{south west}&
                \zoomin{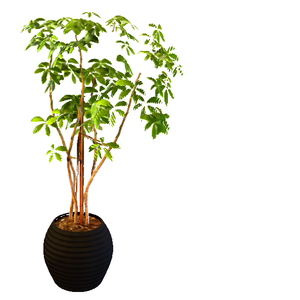}{0.5}{0.3}{1.4}{0.5}{1.15cm}{\mytmplen}{2}{south west}&
                \zoomin{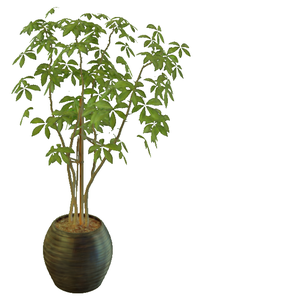}{0.5}{0.3}{1.4}{0.5}{1.15cm}{\mytmplen}{2}{south west}&
                \hspace{-2mm}\includegraphics[width=0.12\textwidth]{./img/na}&
                \zoomin{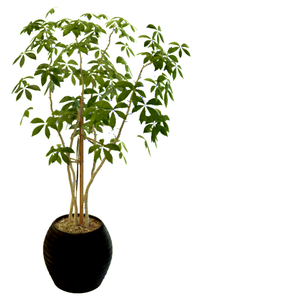}{0.5}{0.3}{1.4}{0.5}{1.15cm}{\mytmplen}{2}{south west}&
                \zoomin{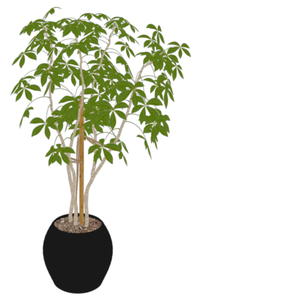}{0.5}{0.3}{1.4}{0.5}{1.15cm}{\mytmplen}{2}{south west}
                \\
                \hspace{-1mm}{\rotatebox{90}{\quad \hspace{4mm} {Normal}}}&
                \hspace{-2mm}\includegraphics[width=0.12\textwidth]{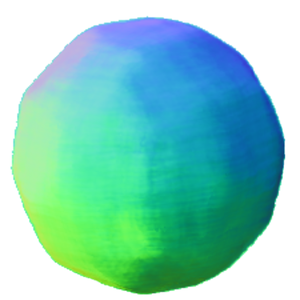}&
                \hspace{-2mm}\includegraphics[width=0.12\textwidth]{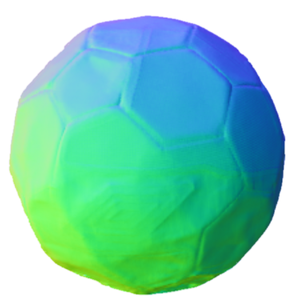}&
                \hspace{-2mm}\includegraphics[width=0.12\textwidth]{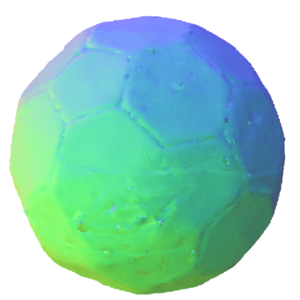}&
                \hspace{-2mm}\includegraphics[width=0.12\textwidth]{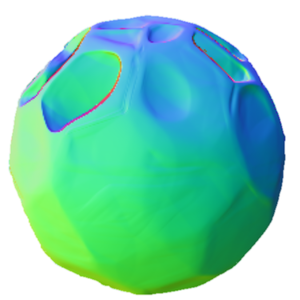}&
                \hspace{-2mm}\includegraphics[width=0.12\textwidth]{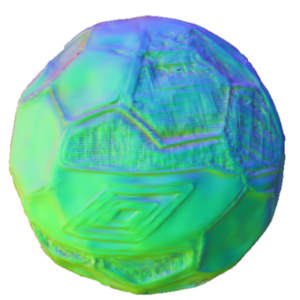}&
                \hspace{-2mm}\includegraphics[width=0.12\textwidth]{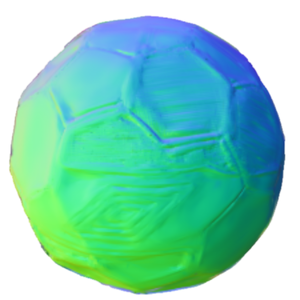}&
                \hspace{-2mm}\includegraphics[width=0.12\textwidth]{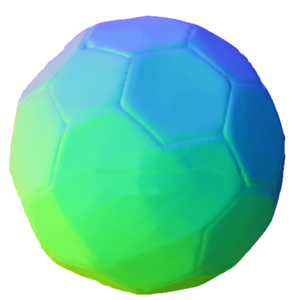}&
                \hspace{-2mm}\includegraphics[width=0.12\textwidth]{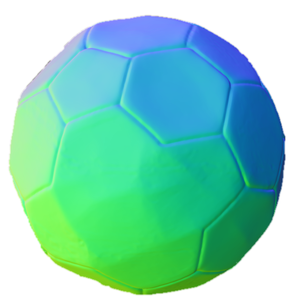}
                \\
                \hspace{-1mm}{\rotatebox{90}{\quad \hspace{4mm} {\makecell{Albedo}}}}&
                \hspace{-2mm}\includegraphics[width=0.12\textwidth]{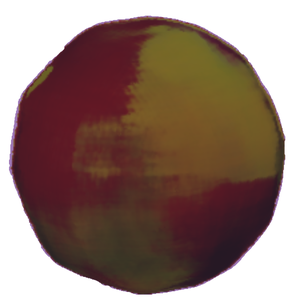}&
                \hspace{-2mm}\includegraphics[width=0.12\textwidth]{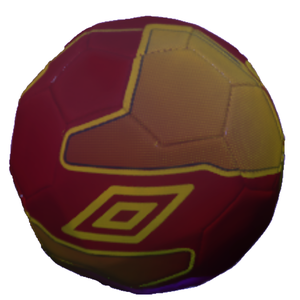}&
                \hspace{-2mm}\includegraphics[width=0.12\textwidth]{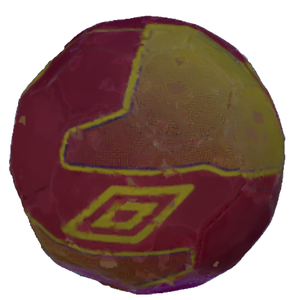}&
                \hspace{-2mm}\includegraphics[width=0.12\textwidth]{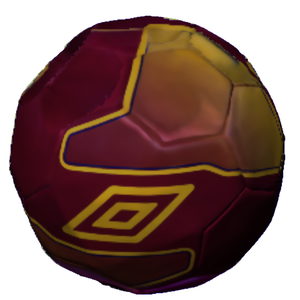}&
                \hspace{-2mm}\includegraphics[width=0.12\textwidth]{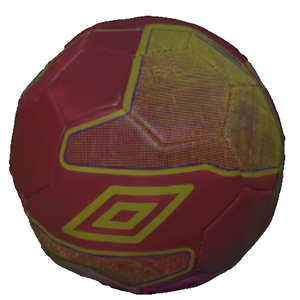}&
                \hspace{-2mm}\includegraphics[width=0.12\textwidth]{./img/na}&
                \hspace{-2mm}\includegraphics[width=0.12\textwidth]{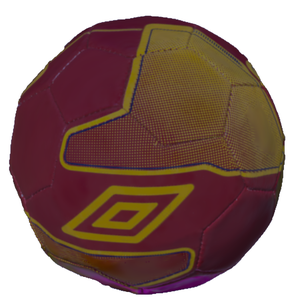}&
                \hspace{-2mm}\includegraphics[width=0.12\textwidth]{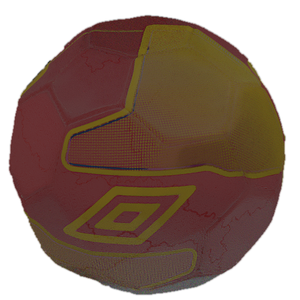}
                \\
                \hspace{-1mm}{\rotatebox{90}{\quad \hspace{4mm} {Normal}}}&
                \hspace{-2mm}\includegraphics[width=0.12\textwidth]{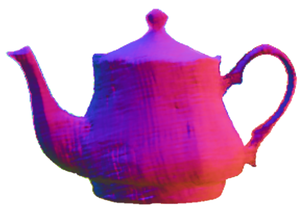}&
                \hspace{-2mm}\includegraphics[width=0.12\textwidth]{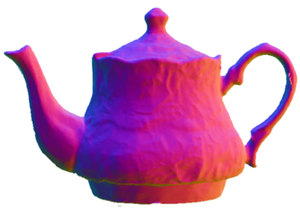}&
                \hspace{-2mm}\includegraphics[width=0.12\textwidth]{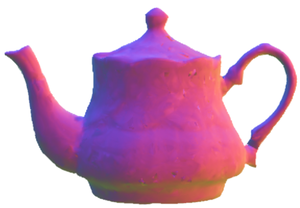}&
                \hspace{-2mm}\includegraphics[width=0.12\textwidth]{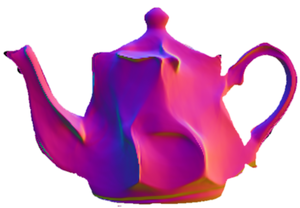}&
                \hspace{-2mm}\includegraphics[width=0.12\textwidth]{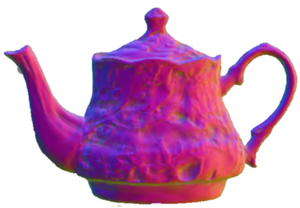}&
                \hspace{-2mm}\includegraphics[width=0.12\textwidth]{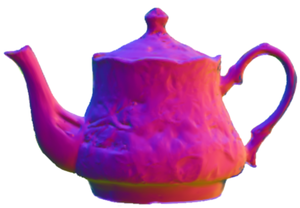}&
                \hspace{-2mm}\includegraphics[width=0.12\textwidth]{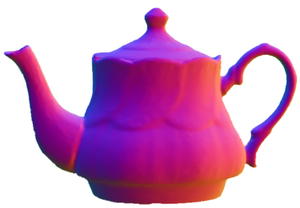}&
                \hspace{-2mm}\includegraphics[width=0.12\textwidth]{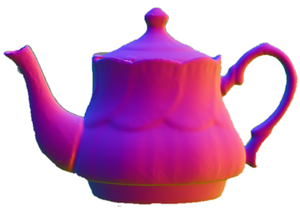}
                \vspace{-4mm}
                \\
                \hspace{-1mm}{\rotatebox{90}{\quad \hspace{4mm} {\makecell{Albedo}}}}&
                \hspace{-2mm}\includegraphics[width=0.12\textwidth]{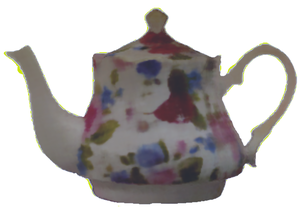}&
                \hspace{-2mm}\includegraphics[width=0.12\textwidth]{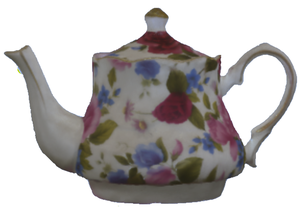}&
                \hspace{-2mm}\includegraphics[width=0.12\textwidth]{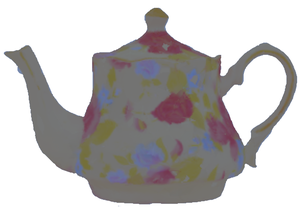}&
                \hspace{-2mm}\includegraphics[width=0.12\textwidth]{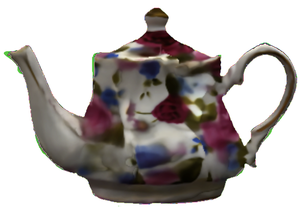}&
                \hspace{-2mm}\includegraphics[width=0.12\textwidth]{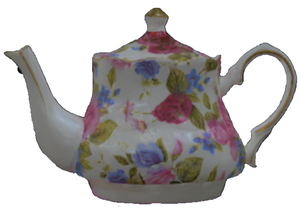}&
                \hspace{-2mm}\includegraphics[width=0.12\textwidth]{./img/na}&
                \hspace{-2mm}\includegraphics[width=0.12\textwidth]{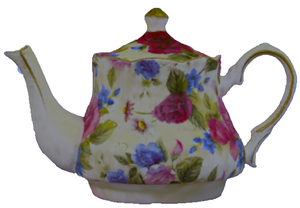}&
                \hspace{-2mm}\includegraphics[width=0.12\textwidth]{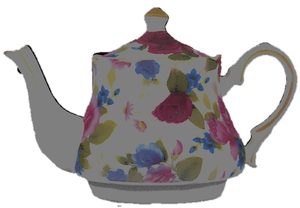}
                \\
                &
                \makecell{NeRFactor}&
                \makecell{\wttp{TensoIR}}&
                \makecell{NDR}&
                \makecell{NeRO}&
                \makecell{RGS}&
                \makecell{GShader}&
                \makecell{Ours}&
                \makecell{GT}
            \end{tabular*}
    }
    \caption{
        Decomposed \yl{result} comparisons with NeRFactor~\cite{NeRFactor}, \wttp{TensoIR}~\cite{TensoIR}, NDR~\cite{NvDiffRec}, NeRO~\cite{NeRO}, RelightableGaussian (RGS)~\cite{RelightableGaussian}, and GaussianShader (GShader)~\cite{GaussianShader}. 
        In every two rows, we show decomposed normal and diffuse albedo components by different methods and compare \yl{them} with the ground truth. 
        Note that GaussianShader only decomposes diffuse color instead of diffuse albedo so its diffuse albedo results are unavailable. 
    }
    \label{fig:decom}
\end{figure*}

\begin{table*}[t!]
    \caption{
        Quantitative comparison of decomposed diffuse albedos and normals by different methods. 
        Diffuse albedo is evaluated with PSNR, SSIM and LPIPS metrics while estimated normals are evaluated with MSE. 
        Note that GaussianShader (GShader)~\cite{GaussianShader} only decomposes diffuse color instead of diffuse albedo so its albedo results are unavailable.
    }
                \setlength\tabcolsep{0pt}
                \begin{tabular*}{\linewidth}{@{\extracolsep{\fill}} ccccccccccccc }
\hline
                    \multirow{2}[2]{*}{Methods} &                                                         \multicolumn{4}{c}{\tabincell{c}{NeRF Synthetic}}                                                         &                                                         \multicolumn{4}{c}{\tabincell{c}{Shiny Blender}}                                                          &                                                          \multicolumn{4}{c}{\tabincell{c}{Stanford ORB}}                                                          \\ \cmidrule{2-13}
                                                & \tabincell{c}{\small PSNR$^\uparrow$} & \tabincell{c}{\small SSIM$^\uparrow$} & \tabincell{c}{\small LPIPS$^\downarrow$} & \tabincell{c}{\small MSE$^\downarrow$} & \tabincell{c}{\small PSNR$^\uparrow$} & \tabincell{c}{\small SSIM$^\uparrow$} & \tabincell{c}{\small LPIPS$^\downarrow$} & \tabincell{c}{\small MSE$^\downarrow$} & \tabincell{c}{\small PSNR$^\uparrow$} & \tabincell{c}{\small SSIM$^\uparrow$} & \tabincell{c}{\small LPIPS$^\downarrow$} & \tabincell{c}{\small MSE$^\downarrow$} \\ \hline
                    NeRFactor                   & 19.01                                 & \sbc{0.871}                           & \bc{0.086}                               & 0.098                                  & 18.28                                 & 0.854                                 & \bc{0.106}                               & 0.132                                  & 28.62                                 & 0.958                                 & 0.046                                    & 0.067                                  \\
                    TensoIR                     & \bc{20.29}                            & \sbc{0.871}                           & \sbc{0.103}                              & \sbc{0.056}                            & \sbc{20.43}                           & \sbc{0.879}                           & 0.194                                    & 0.068                                  & 29.11                                 & 0.969                                 & 0.026                                    & \sbc{0.029}                            \\
                    NDR                         & 18.34                                 & 0.859                                 & 0.116                                    & \tbc{0.064}                            & \tbc{19.66}                           & 0.865                                 & 0.192                                    & \tbc{0.058}                            & \sbc{30.67}                           & \tbc{0.971}                           & \sbc{0.023}                              & 0.043                            \\
                    NeRO                        & \sbc{19.69}                           & 0.866                                 & 0.137                                    & 0.135                                  & 18.31                                 & \tbc{0.867}                           & \sbc{0.187}                              & \bc{0.037}                             & 24.70                                 & 0.948                                 & 0.045                                    & 0.144                                  \\
                    RGS                         & 18.13                                 & 0.862                                 & 0.115                                    & 0.072                                  & 18.10                                 & 0.847                                 & 0.198                                    & 0.065                                  & \tbc{29.33}                           & \bc{0.979}                            & \tbc{0.025}                              & 0.049                                  \\
                    GShader                     & -                                     & -                                     & -                                        & 0.115                                  & -                                     & -                                     & -                                        & 0.093                                  & -                                     & -                                     & -                                        & \tbc{0.037}                            \\ \hline
                    w/o NFD                     & 16.25                                 & 0.821                                 & 0.138                                    & 0.143                                  & 16.58                                 & 0.839                                 & 0.189                                    & 0.102                                  & 29.04                                 & 0.964                                 & 0.039                                    & 0.081                                  \\
                    Ours                        & \tbc{19.14}                           & \bc{0.878}                            & \tbc{0.109}                              & \bc{0.053}                             & \bc{20.65}                            & \bc{0.883}                            & \tbc{0.188}                              & \bc{0.037}                             & \bc{31.99}                            & \sbc{0.977}                           & \bc{0.021}                               & \bc{0.024}                             \\ \hline
                \end{tabular*}
    \label{tab:decom}
\end{table*}

\begin{figure*}[!t]
    \centering
    {   
        \newlength\mytmplenrelight
        \setlength\mytmplenrelight{.11\linewidth}
        \setlength\tabcolsep{0pt}
        \begin{tabular}{ccccccccc}
            {\includegraphics[width=0.11\linewidth]{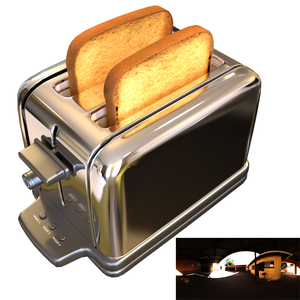}}&
            {\includegraphics[width=0.11\linewidth]{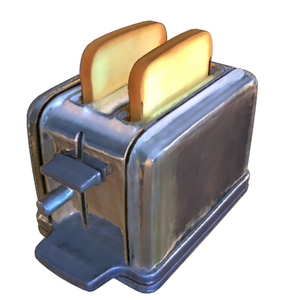}}&
            {\includegraphics[width=0.11\linewidth]{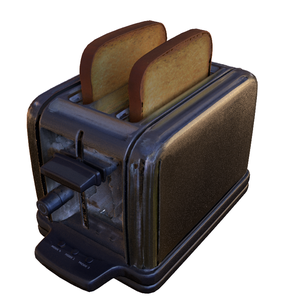}}&
            {\includegraphics[width=0.11\linewidth]{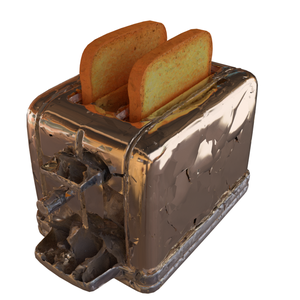}}&
            {\includegraphics[width=0.11\linewidth]{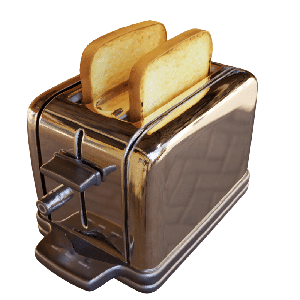}}&
            {\includegraphics[width=0.11\linewidth]{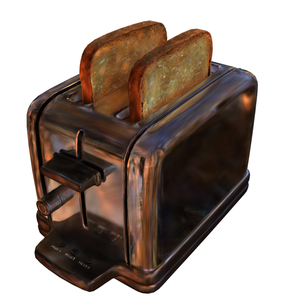}}&
            {\includegraphics[width=0.11\linewidth]{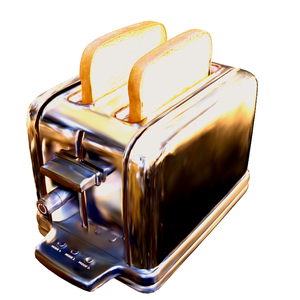}}&
            {\includegraphics[width=0.11\linewidth]{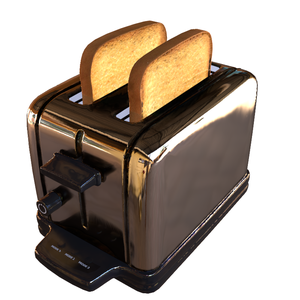}}&
            {\includegraphics[width=0.11\linewidth]{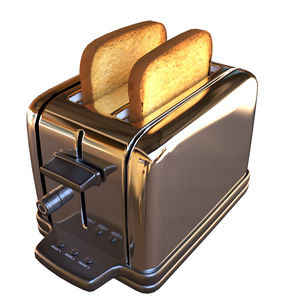}}
            \\
            {\includegraphics[width=0.11\linewidth]{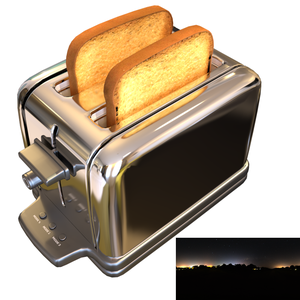}}&
            {\includegraphics[width=0.11\linewidth]{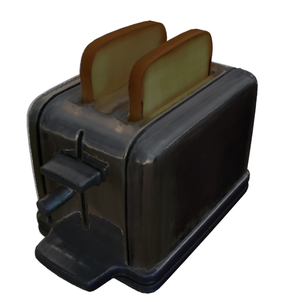}}&
            {\includegraphics[width=0.11\linewidth]{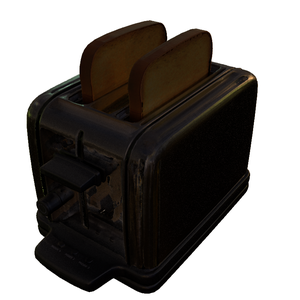}}&
            {\includegraphics[width=0.11\linewidth]{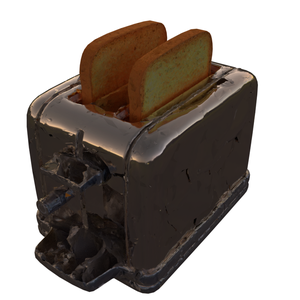}}&
            {\includegraphics[width=0.11\linewidth]{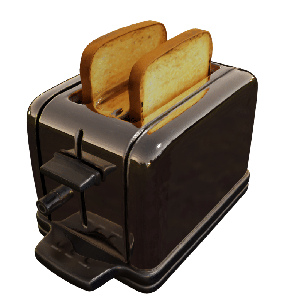}}&
            {\includegraphics[width=0.11\linewidth]{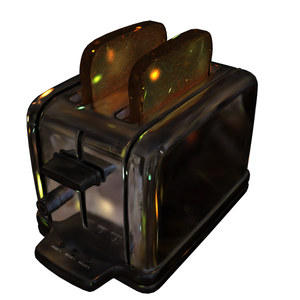}}&
            {\includegraphics[width=0.11\linewidth]{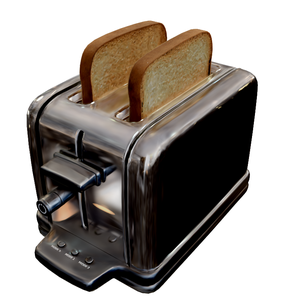}}&
            {\includegraphics[width=0.11\linewidth]{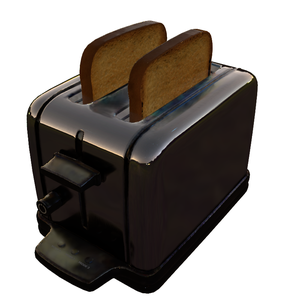}}&
            {\includegraphics[width=0.11\linewidth]{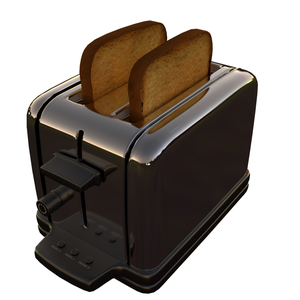}}
            \\
            {\includegraphics[width=0.11\linewidth]{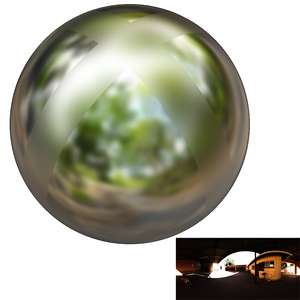}}&
            {\includegraphics[width=0.11\linewidth]{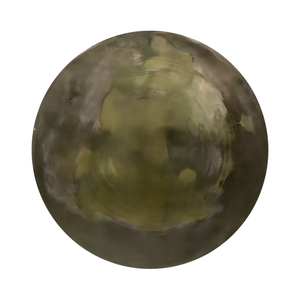}}&
            {\includegraphics[width=0.11\linewidth]{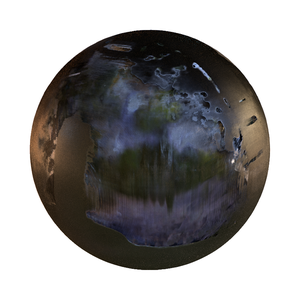}}&
            {\includegraphics[width=0.11\linewidth]{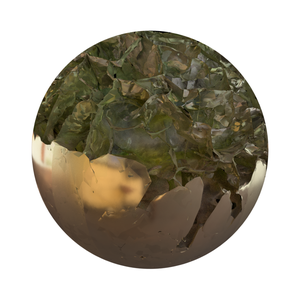}}&
            {\includegraphics[width=0.11\linewidth]{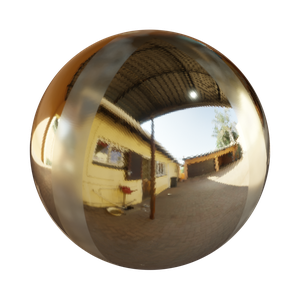}}&
            {\includegraphics[width=0.11\linewidth]{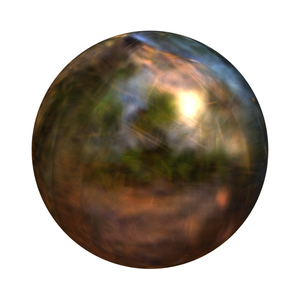}}&
            {\includegraphics[width=0.11\linewidth]{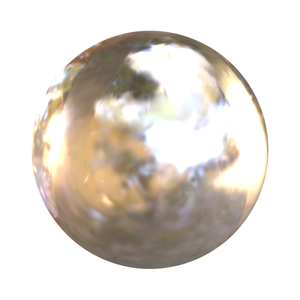}}&
            {\includegraphics[width=0.11\linewidth]{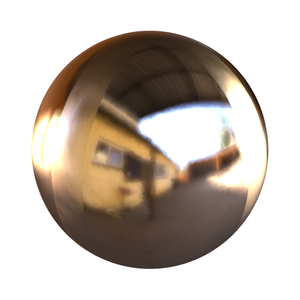}}&
            {\includegraphics[width=0.11\linewidth]{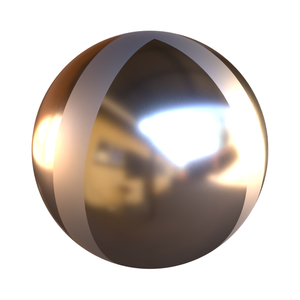}}
            \\
            {\includegraphics[width=0.11\linewidth]{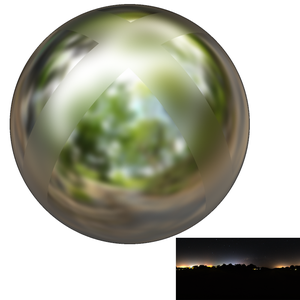}}&
            {\includegraphics[width=0.11\linewidth]{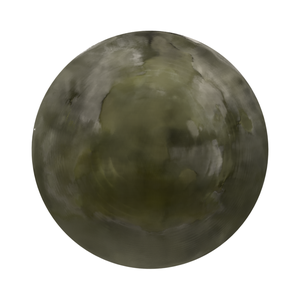}}&
            {\includegraphics[width=0.11\linewidth]{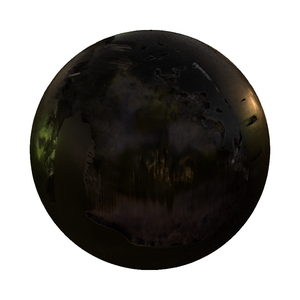}}&
            {\includegraphics[width=0.11\linewidth]{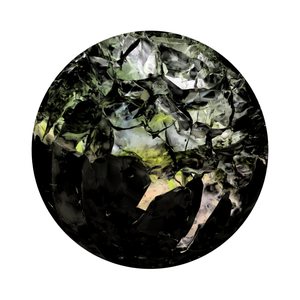}}&
            {\includegraphics[width=0.11\linewidth]{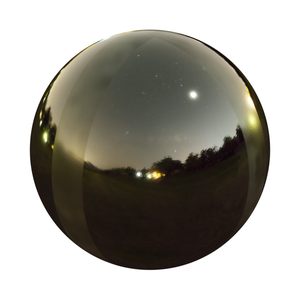}}&
            {\includegraphics[width=0.11\linewidth]{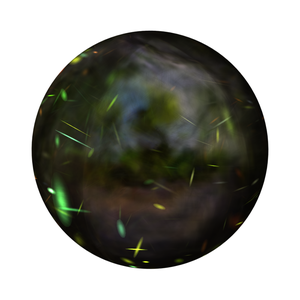}}&
            {\includegraphics[width=0.11\linewidth]{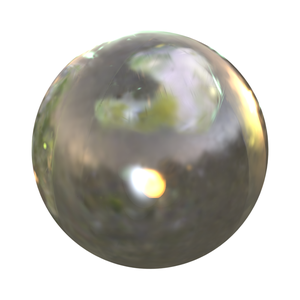}}&
            {\includegraphics[width=0.11\linewidth]{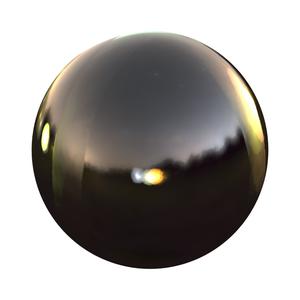}}&
            {\includegraphics[width=0.11\linewidth]{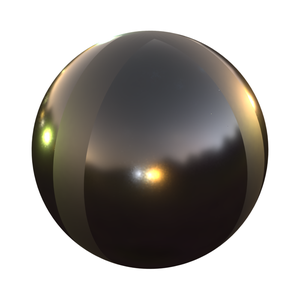}}
            \vspace{-2mm}
            \\
            \vspace{-2mm}
            {\includegraphics[width=0.11\linewidth]{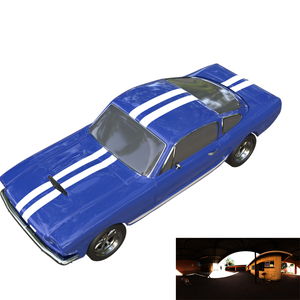}}&
            \zoomin{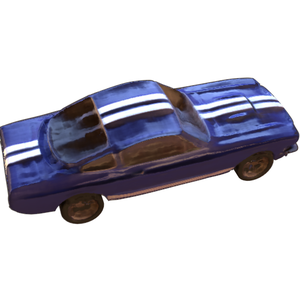}{1.15}{1.2}{0.5}{0.4}{1cm}{\mytmplenrelight}{1.75}{south west}&
            \zoomin{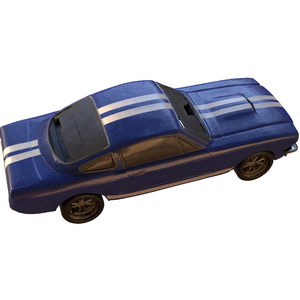}{1.15}{1.2}{0.5}{0.4}{1cm}{\mytmplenrelight}{1.75}{south west}&
            \zoomin{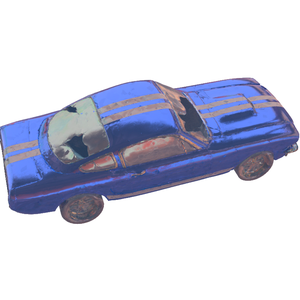}{1.15}{1.2}{0.5}{0.4}{1cm}{\mytmplenrelight}{1.75}{south west}&
            \zoomin{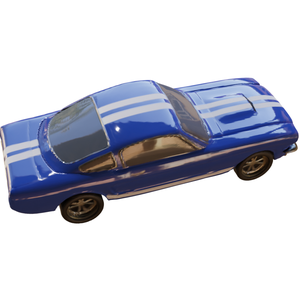}{1.15}{1.2}{0.5}{0.4}{1cm}{\mytmplenrelight}{1.75}{south west}&
            \zoomin{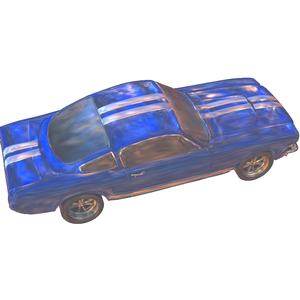}{1.15}{1.2}{0.5}{0.4}{1cm}{\mytmplenrelight}{1.75}{south west}&
            \zoomin{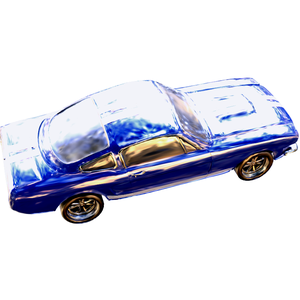}{1.15}{1.2}{0.5}{0.4}{1cm}{\mytmplenrelight}{1.75}{south west}&
            \zoomin{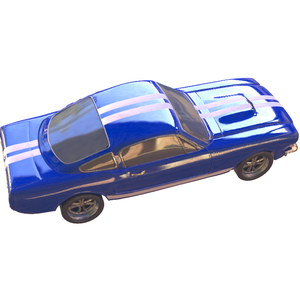}{1.15}{1.2}{0.5}{0.4}{1cm}{\mytmplenrelight}{1.75}{south west}&
            \zoomin{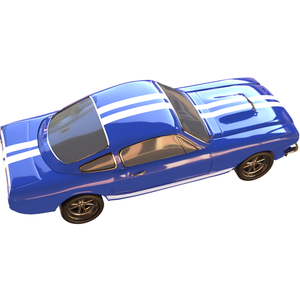}{1.15}{1.2}{0.5}{0.4}{1cm}{\mytmplenrelight}{1.75}{south west}
            \\
            {\includegraphics[width=0.11\linewidth]{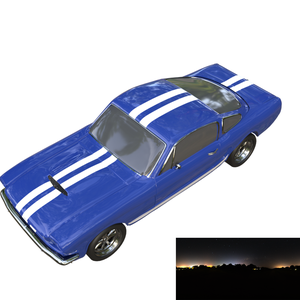}}&
            \zoomin{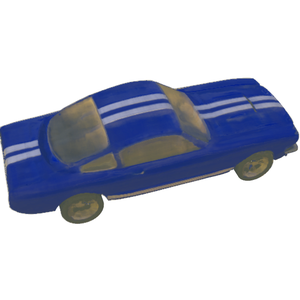}{1.15}{1.2}{0.5}{0.4}{1cm}{\mytmplenrelight}{1.75}{south west}&
            \zoomin{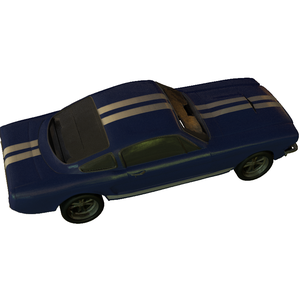}{1.15}{1.2}{0.5}{0.4}{1cm}{\mytmplenrelight}{1.75}{south west}&
            \zoomin{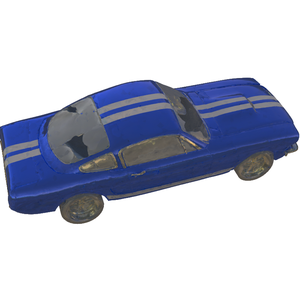}{1.15}{1.2}{0.5}{0.4}{1cm}{\mytmplenrelight}{1.75}{south west}&
            \zoomin{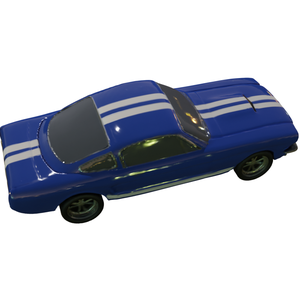}{1.15}{1.2}{0.5}{0.4}{1cm}{\mytmplenrelight}{1.75}{south west}&
            \zoomin{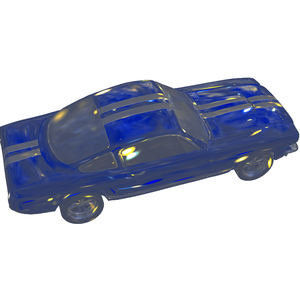}{1.15}{1.2}{0.5}{0.4}{1cm}{\mytmplenrelight}{1.75}{south west}&
            \zoomin{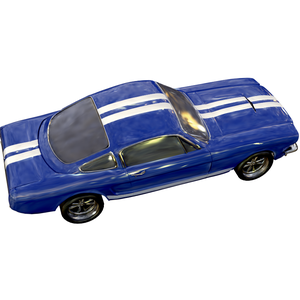}{1.15}{1.2}{0.5}{0.4}{1cm}{\mytmplenrelight}{1.75}{south west}&
            \zoomin{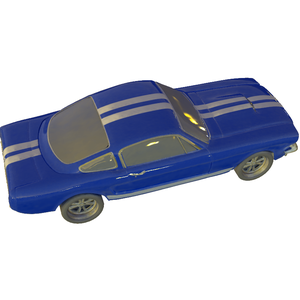}{1.15}{1.2}{0.5}{0.4}{1cm}{\mytmplenrelight}{1.75}{south west}&
            \zoomin{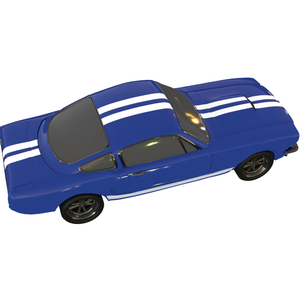}{1.15}{1.2}{0.5}{0.4}{1cm}{\mytmplenrelight}{1.75}{south west}
                \\
                {\includegraphics[width=0.11\linewidth]{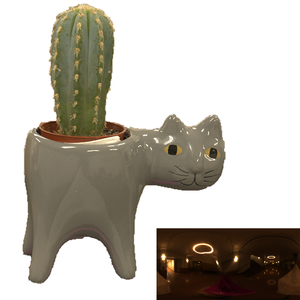}}&
                \zoomin{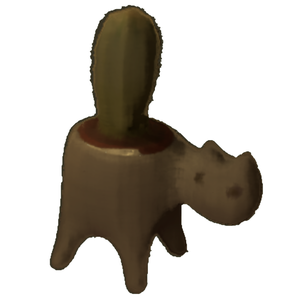}{1.4}{0.8}{0.5}{0.4}{1cm}{\mytmplenrelight}{1.75}{south west}&
                \zoomin{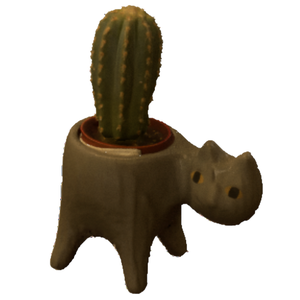}{1.4}{0.8}{0.5}{0.4}{1cm}{\mytmplenrelight}{1.75}{south west}&
                \zoomin{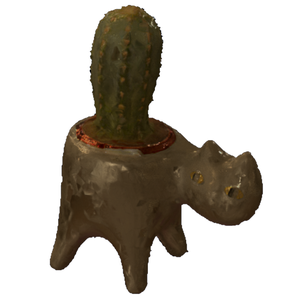}{1.4}{0.8}{0.5}{0.4}{1cm}{\mytmplenrelight}{1.75}{south west}&
                \zoomin{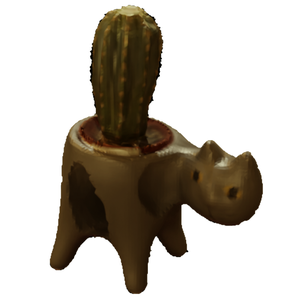}{1.4}{0.8}{0.5}{0.4}{1cm}{\mytmplenrelight}{1.75}{south west}&
                \zoomin{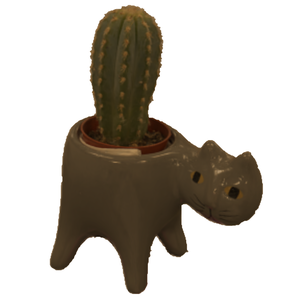}{1.4}{0.8}{0.5}{0.4}{1cm}{\mytmplenrelight}{1.75}{south west}&
                \zoomin{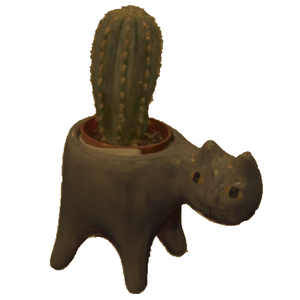}{1.4}{0.8}{0.5}{0.4}{1cm}{\mytmplenrelight}{1.75}{south west}&
                \zoomin{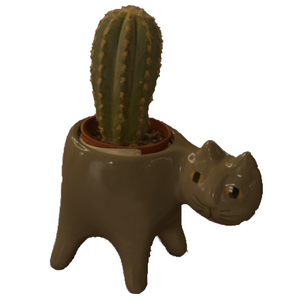}{1.4}{0.8}{0.5}{0.4}{1cm}{\mytmplenrelight}{1.75}{south west}&
                \zoomin{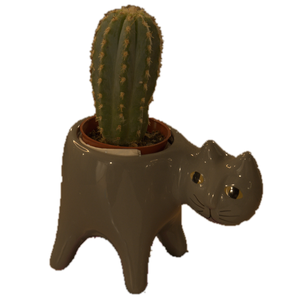}{1.4}{0.8}{0.5}{0.4}{1cm}{\mytmplenrelight}{1.75}{south west}
                \\
                {\includegraphics[width=0.11\linewidth]{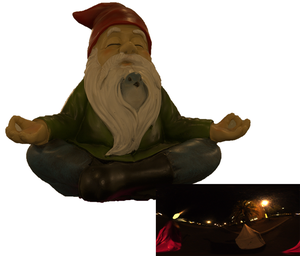}}&
                \zoomin{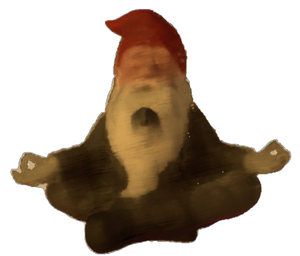}{1}{1.4}{0.5}{0.4}{1cm}{\mytmplenrelight}{1.75}{south west}&
                \zoomin{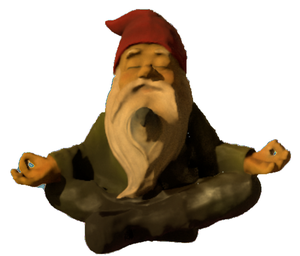}{1}{1.4}{0.5}{0.4}{1cm}{\mytmplenrelight}{1.75}{south west}&
                \zoomin{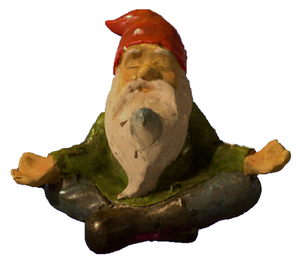}{1}{1.4}{0.5}{0.4}{1cm}{\mytmplenrelight}{1.75}{south west}&
                \zoomin{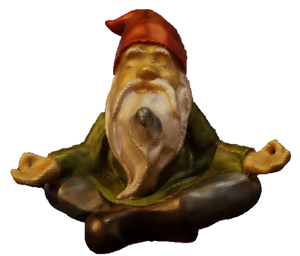}{1}{1.4}{0.5}{0.4}{1cm}{\mytmplenrelight}{1.75}{south west}&
                \zoomin{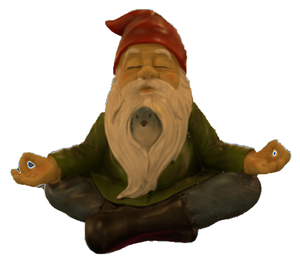}{1}{1.4}{0.5}{0.4}{1cm}{\mytmplenrelight}{1.75}{south west}&
                \zoomin{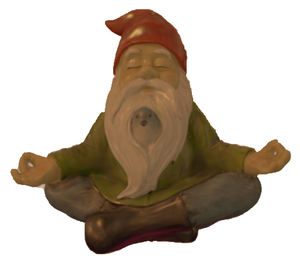}{1}{1.4}{0.5}{0.4}{1cm}{\mytmplenrelight}{1.75}{south west}&
                \zoomin{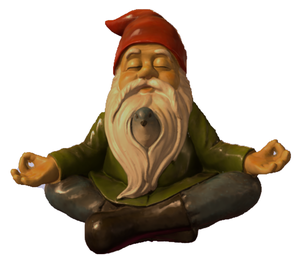}{1}{1.4}{0.5}{0.4}{1cm}{\mytmplenrelight}{1.75}{south west}&
                \zoomin{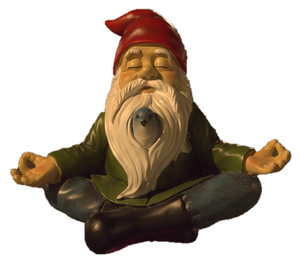}{1}{1.4}{0.5}{0.4}{1cm}{\mytmplenrelight}{1.75}{south west}
                \\
                Input&
                NeRFactor&
                \wttp{TensoIR}&
                NDR&
                NeRO&
                RGS&
                GShader&
                Ours&
                GT
            \end{tabular}
        }
        \caption{
            Relighting comparisons with NeRFactor~\cite{NeRFactor}, \wttp{TensoIR}~\cite{TensoIR}, NDR~\cite{NvDiffRec}, NeRO~\cite{NeRO}, RelightableGaussian (RGS)~\cite{RelightableGaussian}, and GaussianShader (GShader)~\cite{GaussianShader}. 
            In the first column, we show the input scene and target environment map. 
            Relighting results by different methods and the ground truth are in other columns. 
        }
        \label{fig:relight}
    \end{figure*}

\begin{table*}[t!]
	\caption{
        \wttp{
		Quantitative comparison of relighting results using SSIM, PSNR, and LPIPS metrics. 
		Results are averaged over ten different viewpoints with eight different environment maps on NeRF Synthetic and Shiny Blender datasets. 
        For the Stanford ORB dataset, relighting results are evaluated on the provided 20 image-envmap pairs. 
        }
	}
		\centering
			\setlength\tabcolsep{0pt}
			\begin{tabular*}{\linewidth}{@{\extracolsep{\fill}} cccccccccc }

\hline
			    \multirow{2}[2]{*}{Methods} &                           \multicolumn{3}{c}{\tabincell{c}{NeRF Synthetic}}                            &                            \multicolumn{3}{c}{\tabincell{c}{Shiny Blender}}                   &                 \multicolumn{3}{c}{\tabincell{c}{Stanford ORB}}         \\ \cmidrule{2-10}
			                                & \tabincell{c}{PSNR $^\uparrow$} & \tabincell{c}{SSIM $^\uparrow$} & \tabincell{c}{LPIPS $^\downarrow$} & \tabincell{c}{PSNR $^\uparrow$} & \tabincell{c}{SSIM $^\uparrow$} & \tabincell{c}{LPIPS $^\downarrow$} & \tabincell{c}{PSNR $^\uparrow$} & \tabincell{c}{SSIM $^\uparrow$} & \tabincell{c}{LPIPS $^\downarrow$}\\ \hline
			    NeRFactor                   & 18.40                           & 0.868                           & 0.094                              & 19.79                           & 0.901                           & 0.086                             &30.03&0.970&0.028 \\
                TensoIR                     & \bc{20.25}                      & \bc{0.916}                      & 0.071                              & 19.17                           & 0.887                           & 0.150                             &29.60&0.969&0.026 \\
			    NDR                         & \tbc{19.89}                     & 0.899                           & \sbc{0.066}                        & \tbc{20.70}                     & \tbc{0.921}                     & 0.063                             &\tbc{30.41}&0.968&0.023 \\
			    NeRO                        & \sbc{20.09}                     & \sbc{0.915}                     & \tbc{0.067}                        & \sbc{21.86}                     & \sbc{0.929}                     & 0.059                             &29.67&0.966&0.028 \\
			    RGS                         & 18.27                           & 0.859                           & 0.073                              & 18.03                           & 0.832                           & 0.155                             &29.86&\tbc{0.975}&\sbc{0.018} \\
			    GShader                     & 18.12                           & 0.838                           & 0.145                              & 20.35                           & 0.917                           & \bc{0.042}                        &\sbc{30.72}&\sbc{0.976}&\tbc{0.019} \\ \hline
			    Forward Shading             & 18.21                           & 0.833                           & 0.070                              & 20.05                           & 0.914                           & \tbc{0.053}                       &29.94&0.970&0.024 \\
			    Ours                        & 19.24                           & \tbc{0.904}                     & \bc{0.057}                         & \bc{22.24}                      & \bc{0.939}                      & \sbc{0.049}                       &\bc{31.11}&\bc{0.978}&\bc{0.017} \\ \hline
			\end{tabular*}
	\label{tab:relight}
\end{table*}

\subsection{Decomposition}
To evaluate the decoupling ability of our method, we render optimized diffuse albedo and normal images and compare them with the corresponding ground truth images. 
Baseline methods include NeRFactor~\cite{NeRFactor}, \wttp{TensoIR}~\cite{TensoIR}, NDR~\cite{NvDiffRec}, NeRO~\cite{NeRO}, RelightableGaussian~\cite{RelightableGaussian}, and GuassianShader~\cite{GaussianShader} that model texture and lighting separately. 
As shown in Fig.~\ref{fig:decom}, the smooth or low-resolution lighting representation in NeRFactor and \wttp{TensoIR} may cause incorrect geometry or texture estimation. 
NDR still suffers from its tetrahedral representation and produces an irregular normal estimation and wrong texture estimation. 
\wt{NeRO produces faithful geometry but less correct texture estimation (first to fourth \yl{rows}).}
For RelightableGaussian and GaussianShader, the \yl{normals are} directly learned from the \yl{Gaussian} splatting representation and can be rough on shiny scenes, leading to wrong textures. 
Note that GaussianShader decomposes diffuse color instead of diffuse albedo. 
Overall our method produces smooth normal estimation without losing geometric details and the extracted textures have less lighting baked \yl{in}. 
We also report the PSNR, SSIM \yl{and} LPIPS metrics for the extracted diffuse albedo components and the MSE value for the normal image in Table~\ref{tab:decom} and our method comes on top. 
\wttp{Note that before evaluating the albedo component, we compute the scales between estimated normals and the ground truth normals following NeRFactor~\cite{NeRFactor} except for glossy scenes (`materials' and `mic' in the NeRF Synthetic dataset and `ball' and `toaster' in the Shiny Blender dataset) since the scales would be all 0 and result in infinite PSNRs for all methods.}

\subsection{Editing}
\begin{figure}[!t]
    \centering{
        \begin{tabular}{cccccc}
            \hspace{-4mm}
            \rotatebox{90}{\quad \hspace{-2mm} {\small Geometry}}
            &
            \hspace{-3mm}{\includegraphics[width=0.19\linewidth]{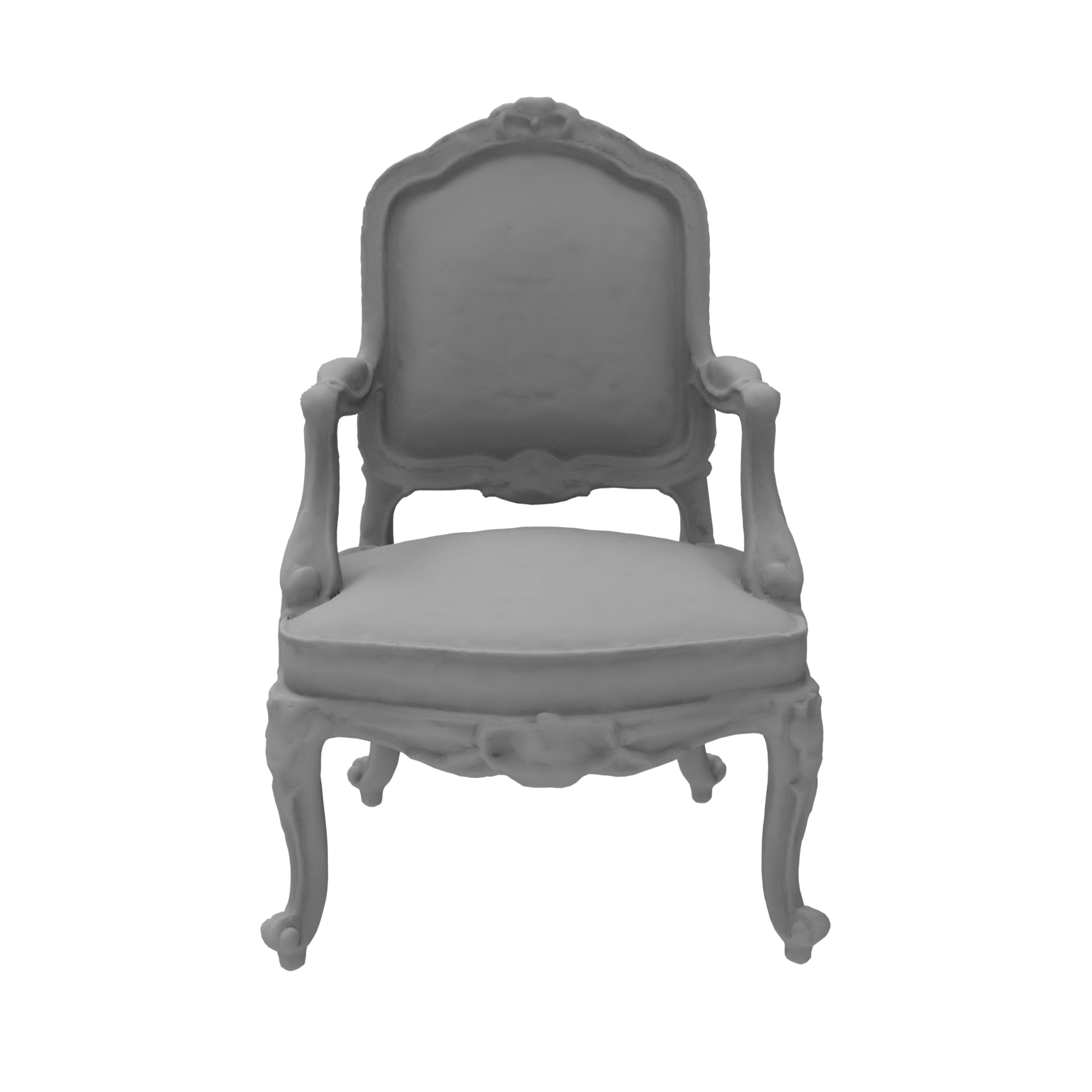}}&
            \hspace{-3mm}{\includegraphics[width=0.19\linewidth]{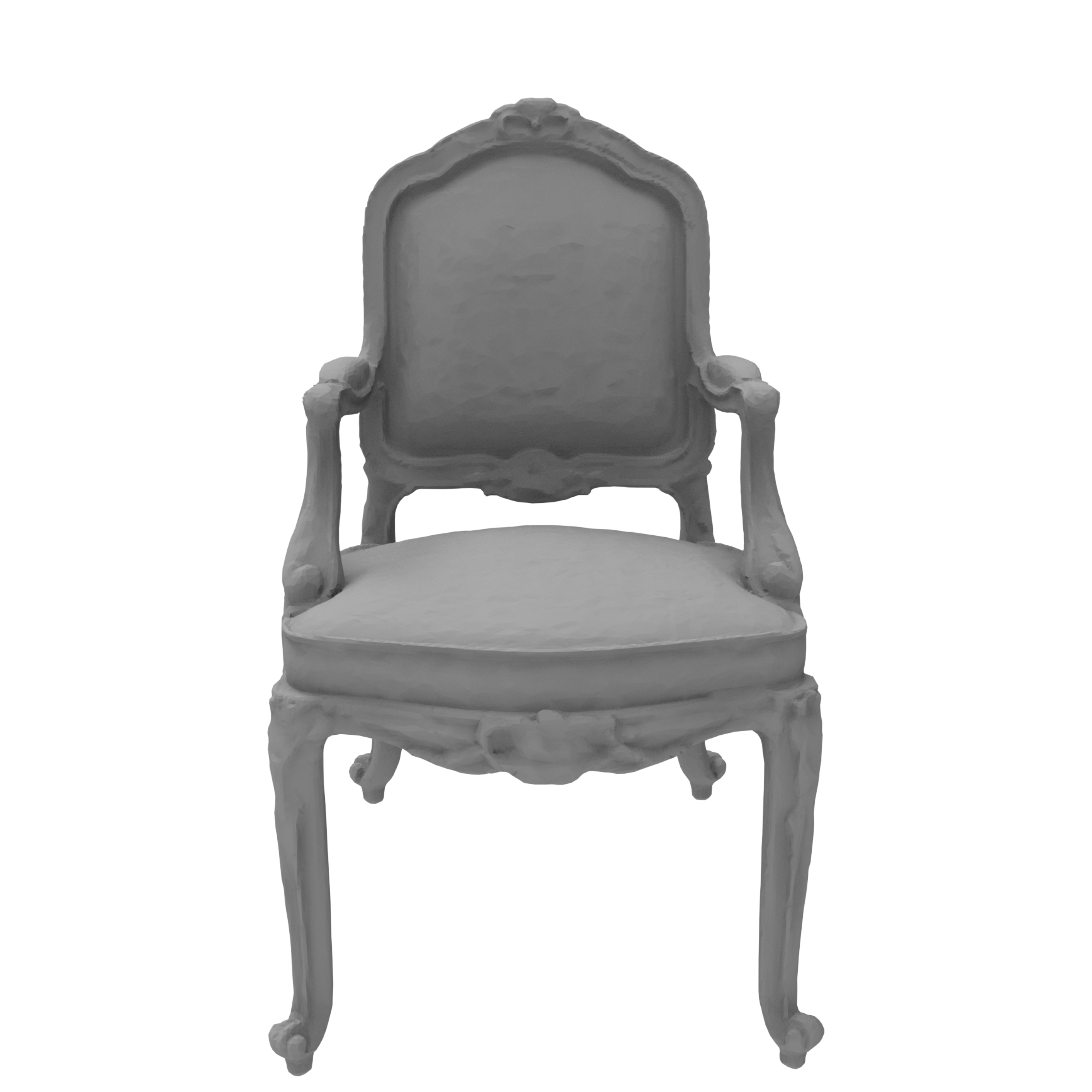}}&
            \hspace{-3mm}{\includegraphics[width=0.19\linewidth]{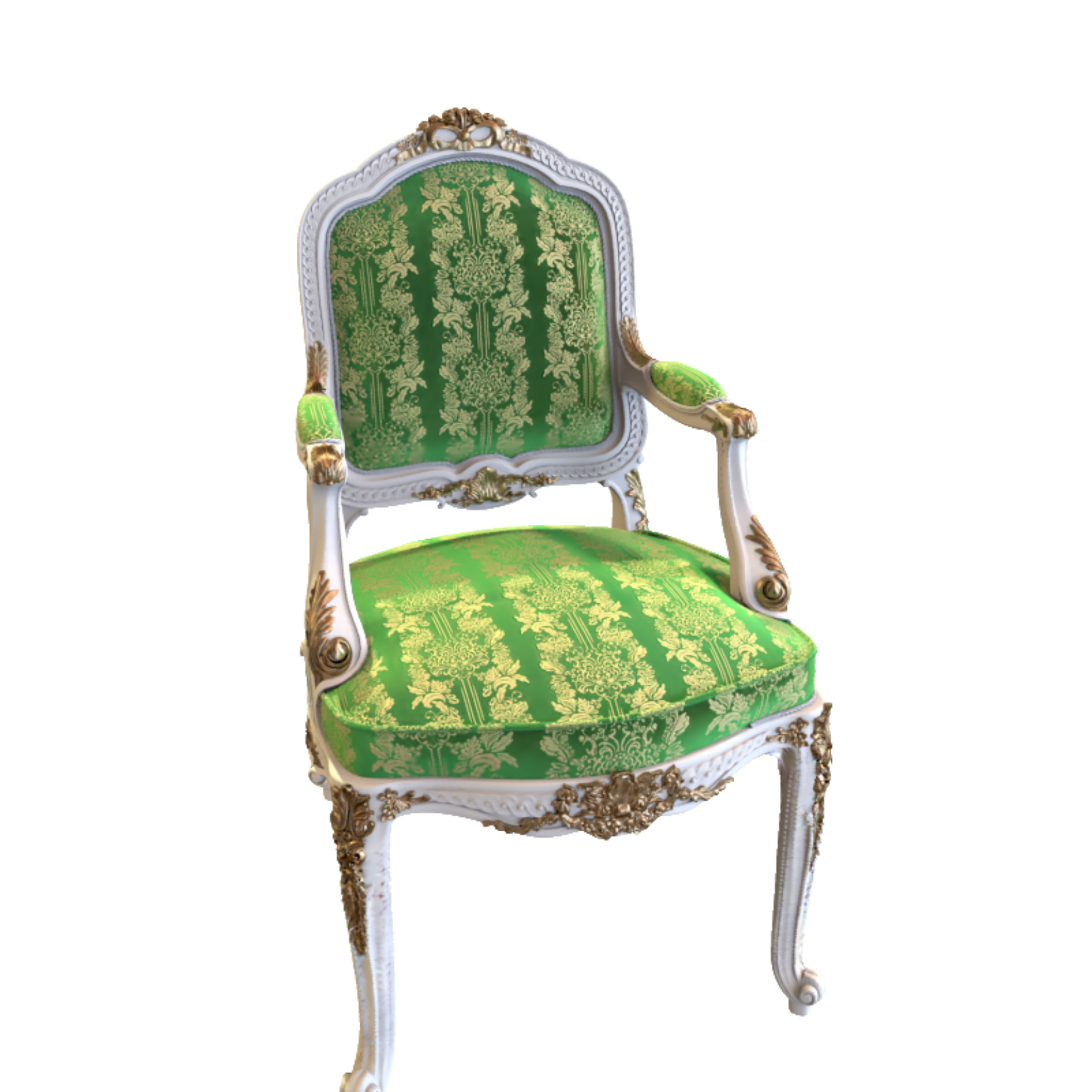}}&
            \hspace{-3mm}{\includegraphics[width=0.19\linewidth]{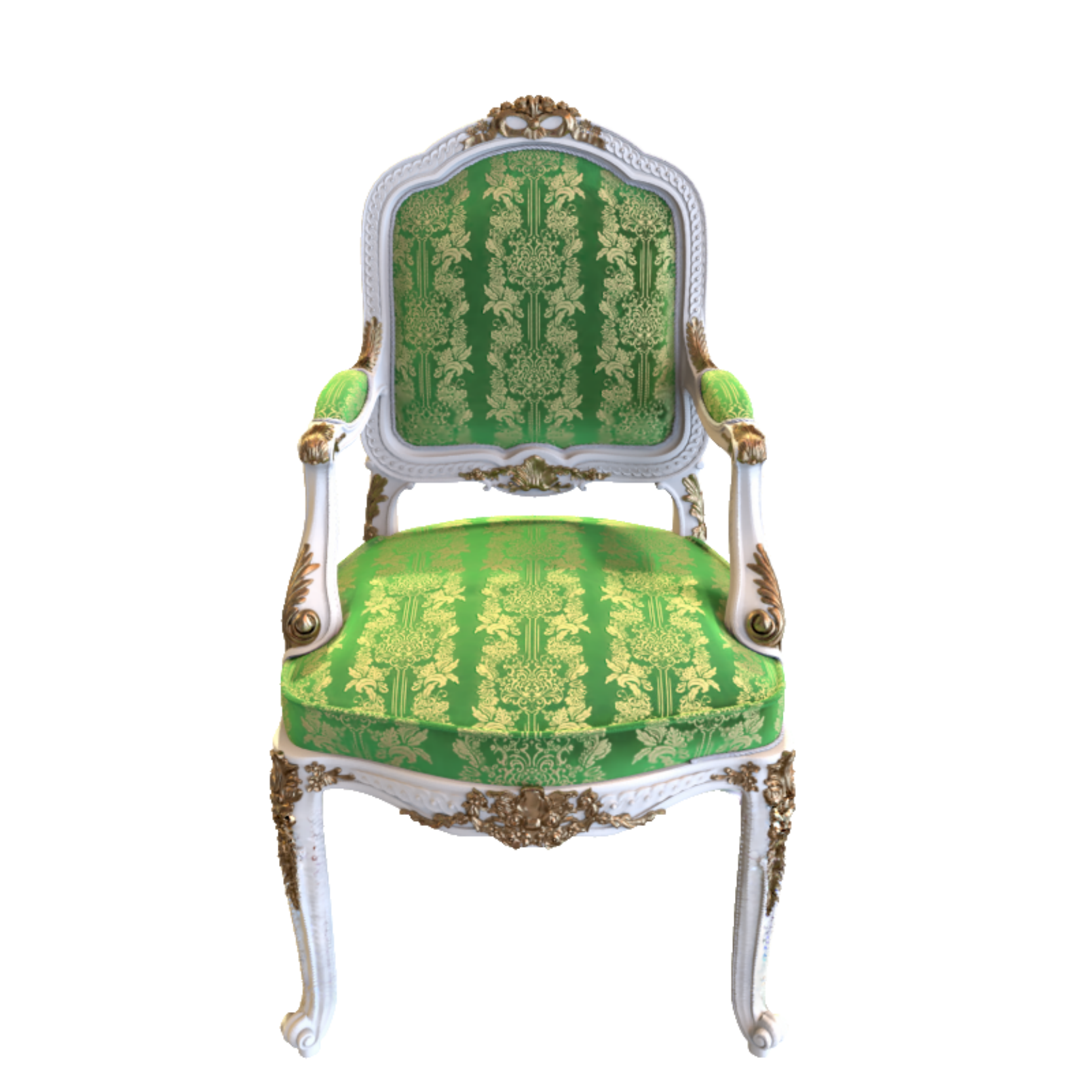}}&
            \hspace{-3mm}{\includegraphics[width=0.19\linewidth]{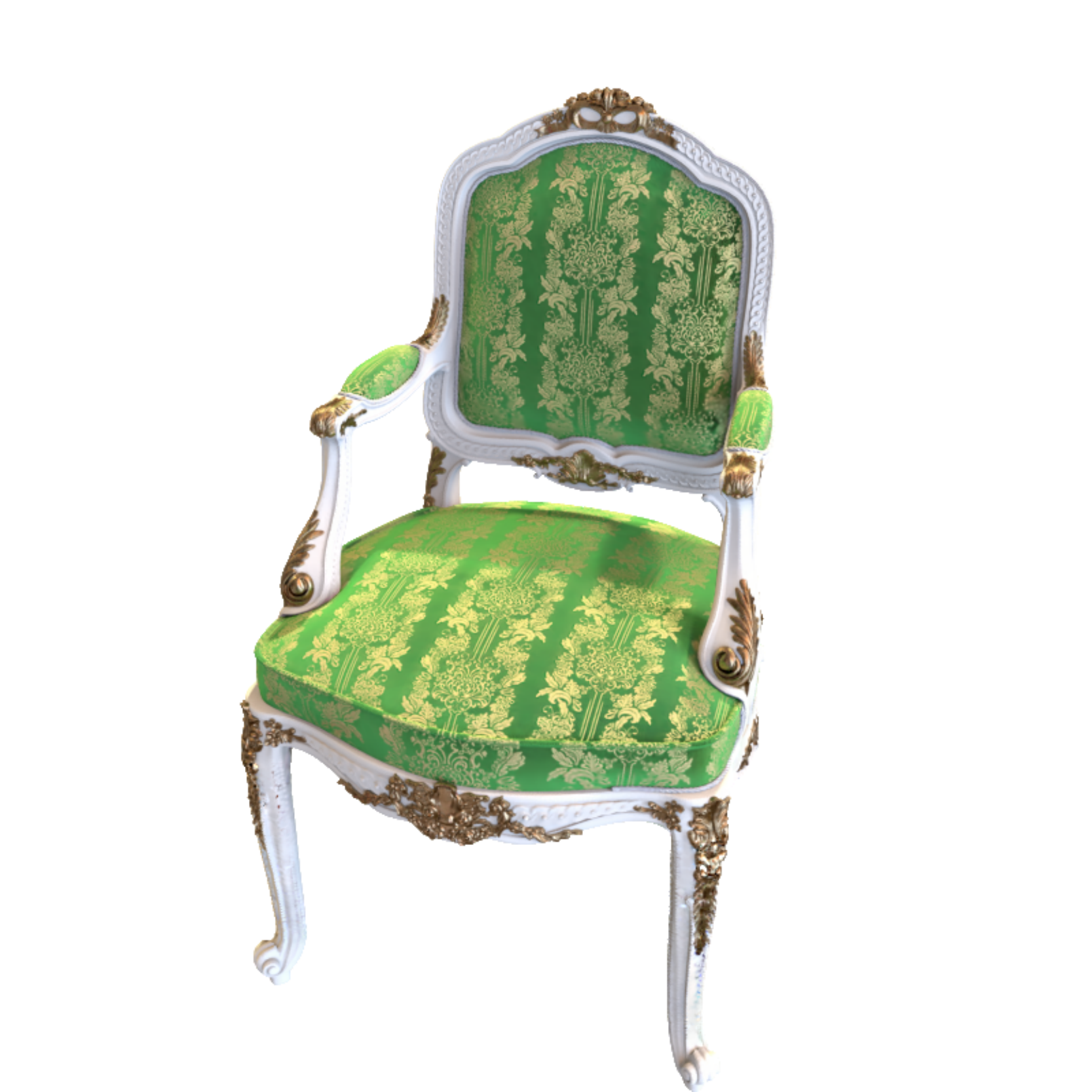}}
            \\
            \hspace{-4mm}
            \rotatebox{90}{\quad \hspace{-3mm} {\small Geometry}}
            &
            \hspace{-3mm}{\includegraphics[width=0.19\linewidth]{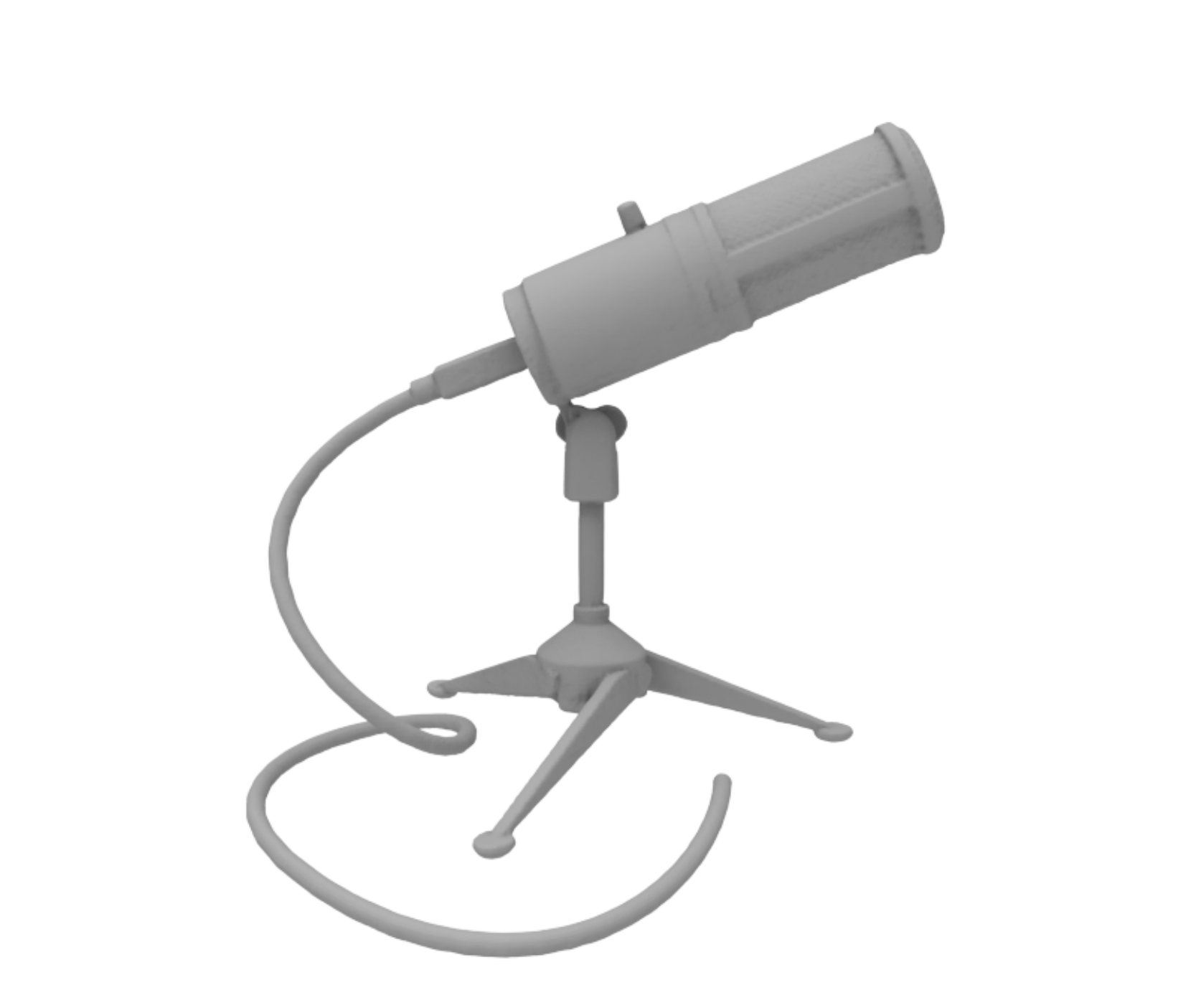}}&
            \hspace{-3mm}{\includegraphics[width=0.19\linewidth]{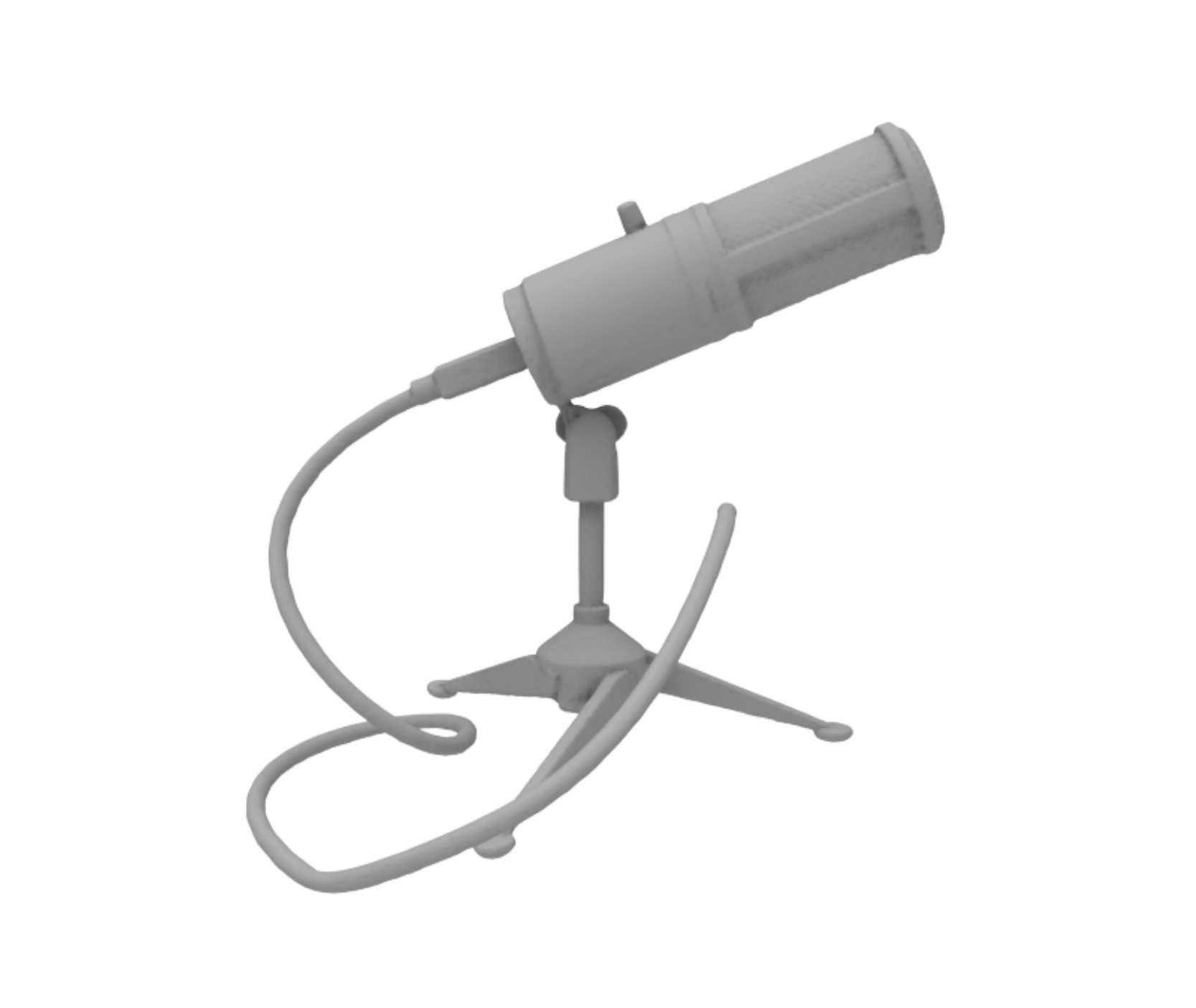}}&
            \hspace{-3mm}{\includegraphics[width=0.19\linewidth]{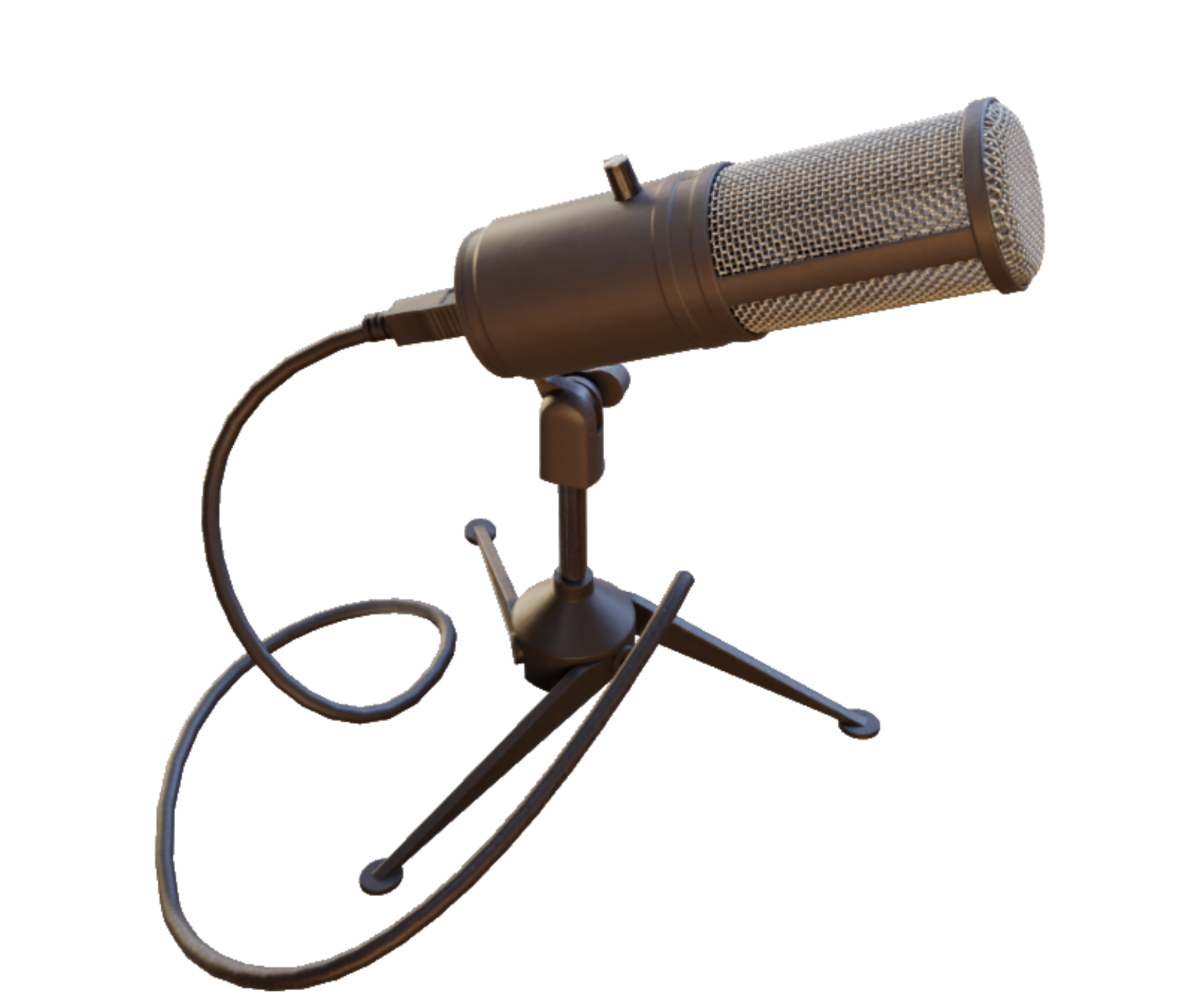}}&
            \hspace{-3mm}{\includegraphics[width=0.19\linewidth]{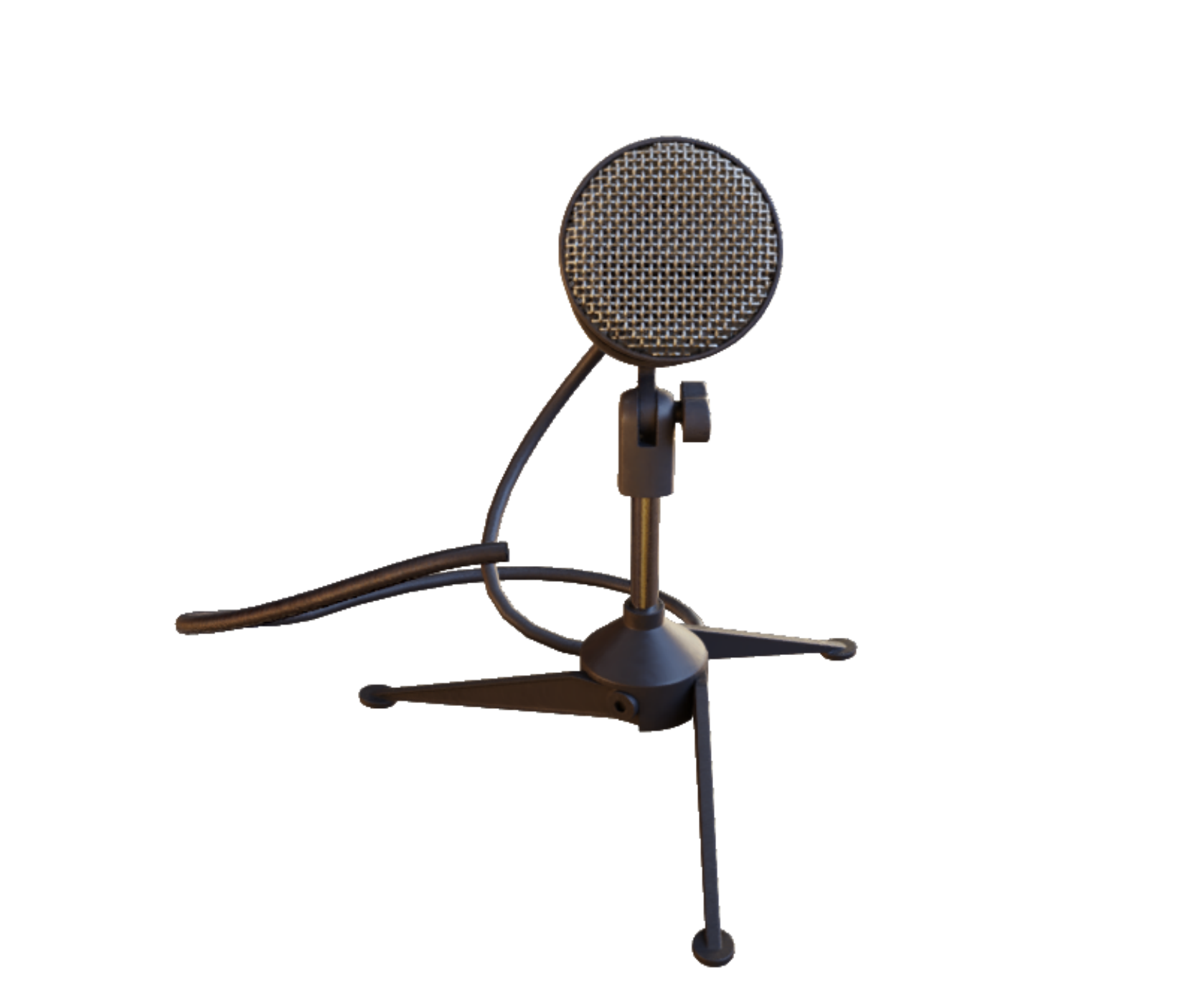}}&
            \hspace{-3mm}{\includegraphics[width=0.19\linewidth]{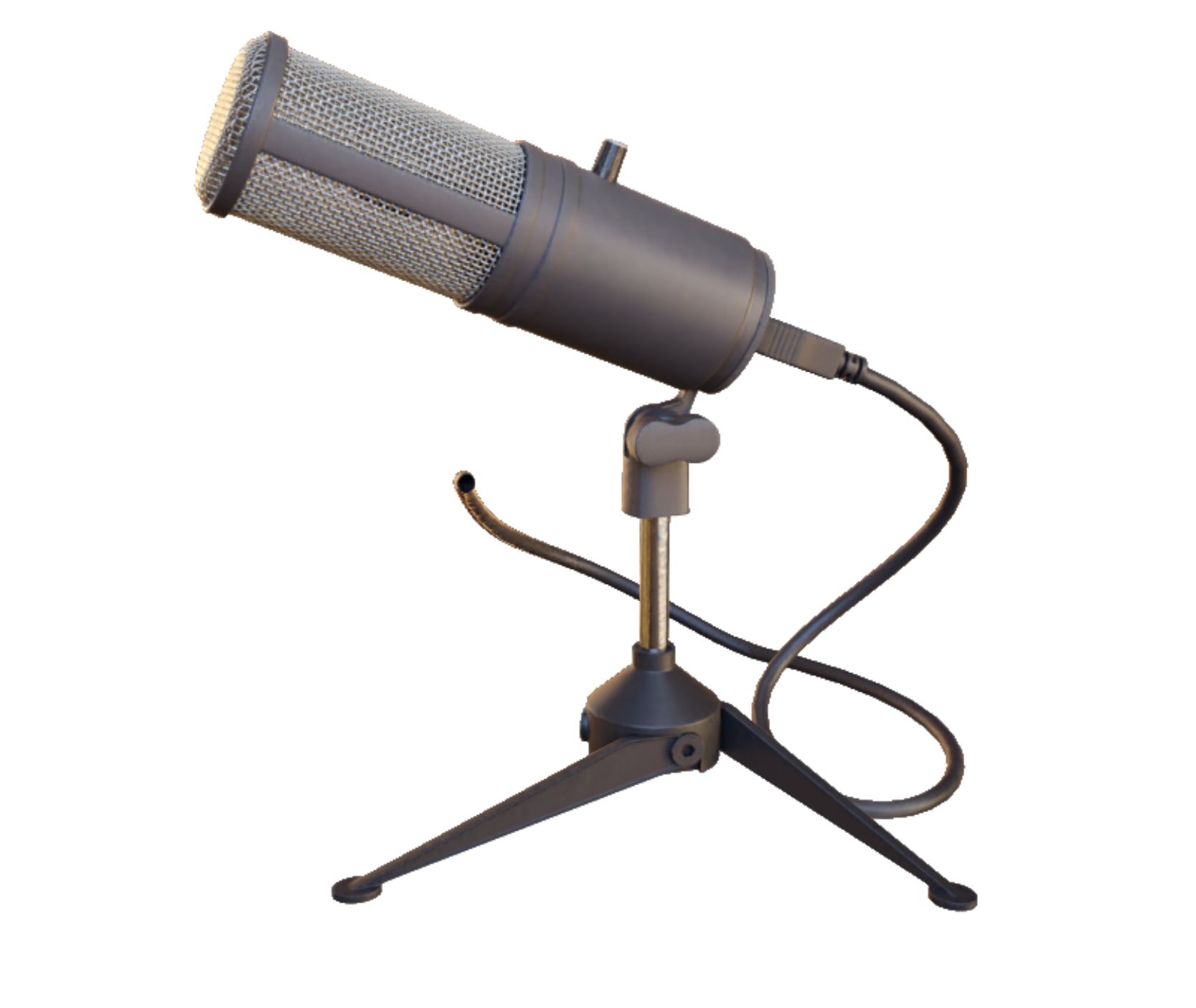}}
            \\
            \hspace{-4mm}
            \rotatebox{90}{\quad \hspace{-1mm} {\small Texture}}
            &
            \hspace{-3mm}{\includegraphics[width=0.19\linewidth]{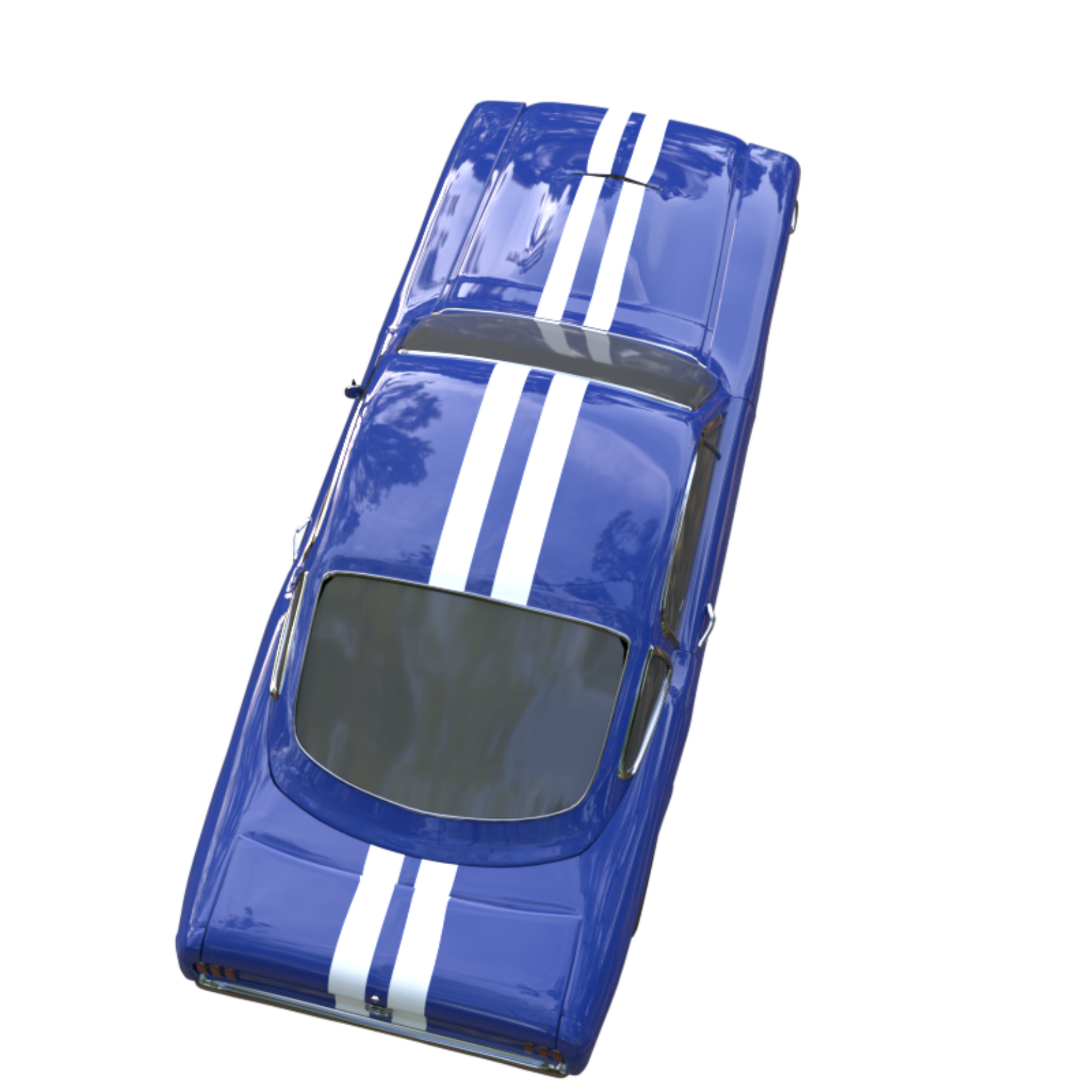}}&
            \hspace{-3mm}{\includegraphics[width=0.19\linewidth]{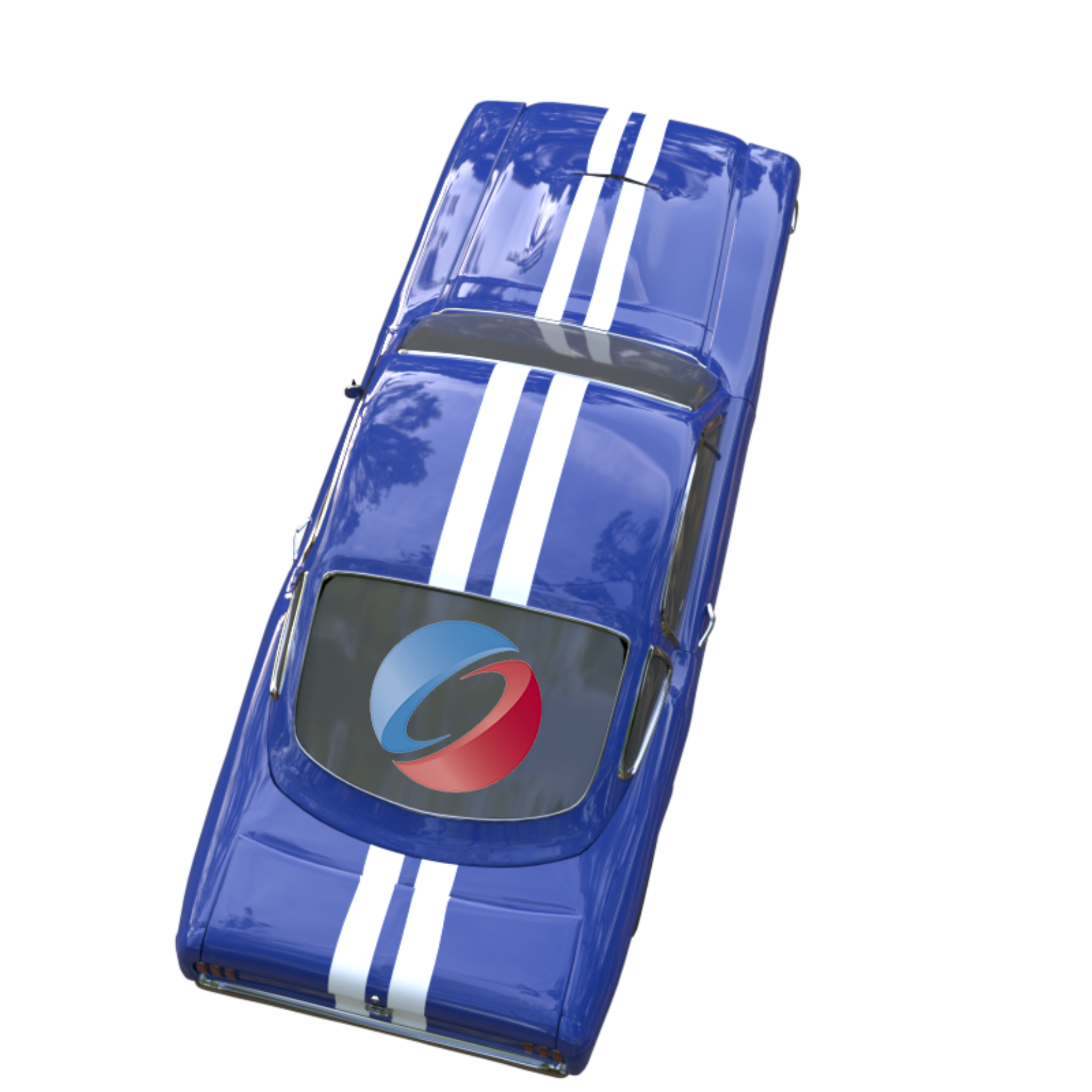}}&
            \hspace{-3mm}{\includegraphics[width=0.19\linewidth]{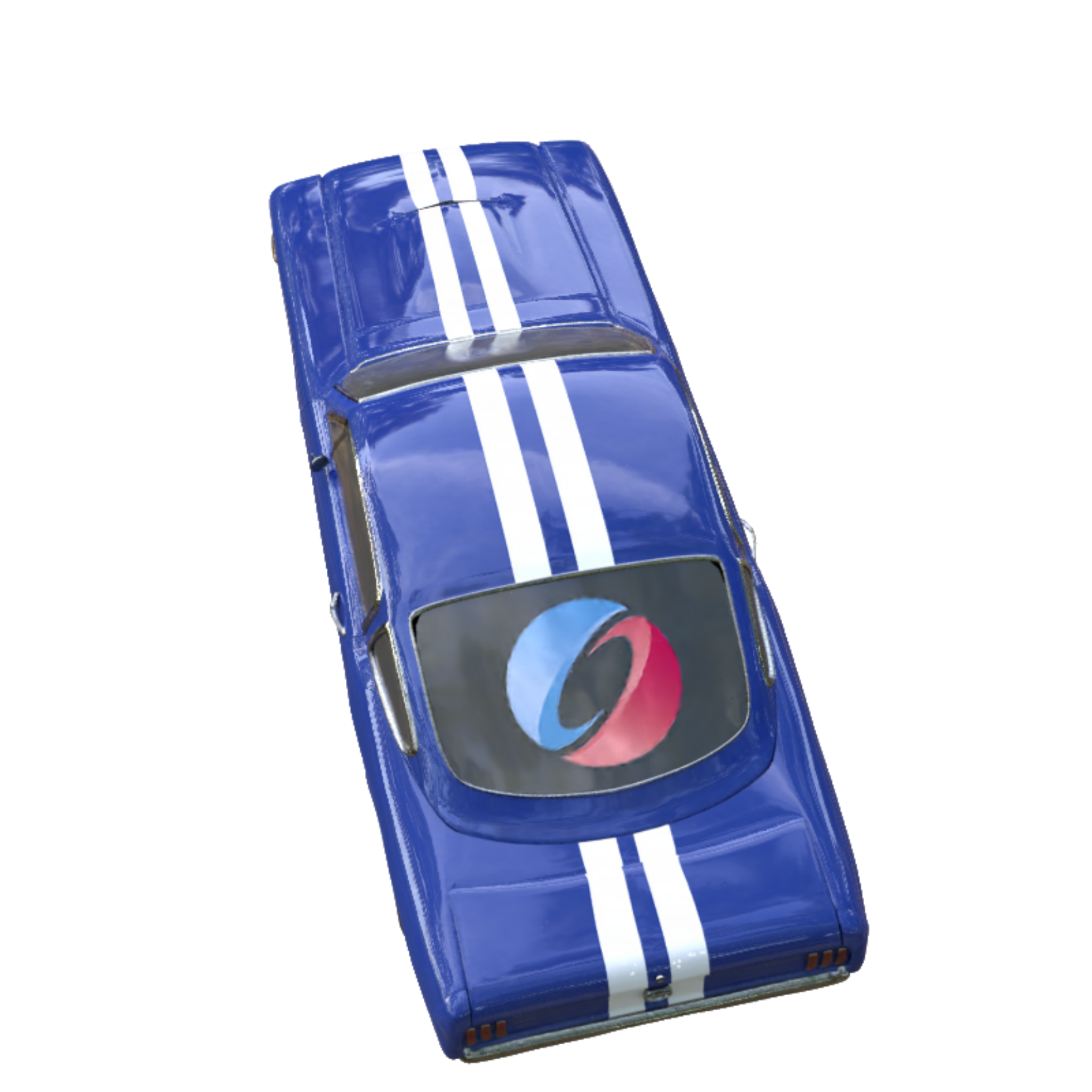}}&
            \hspace{-3mm}{\includegraphics[width=0.19\linewidth]{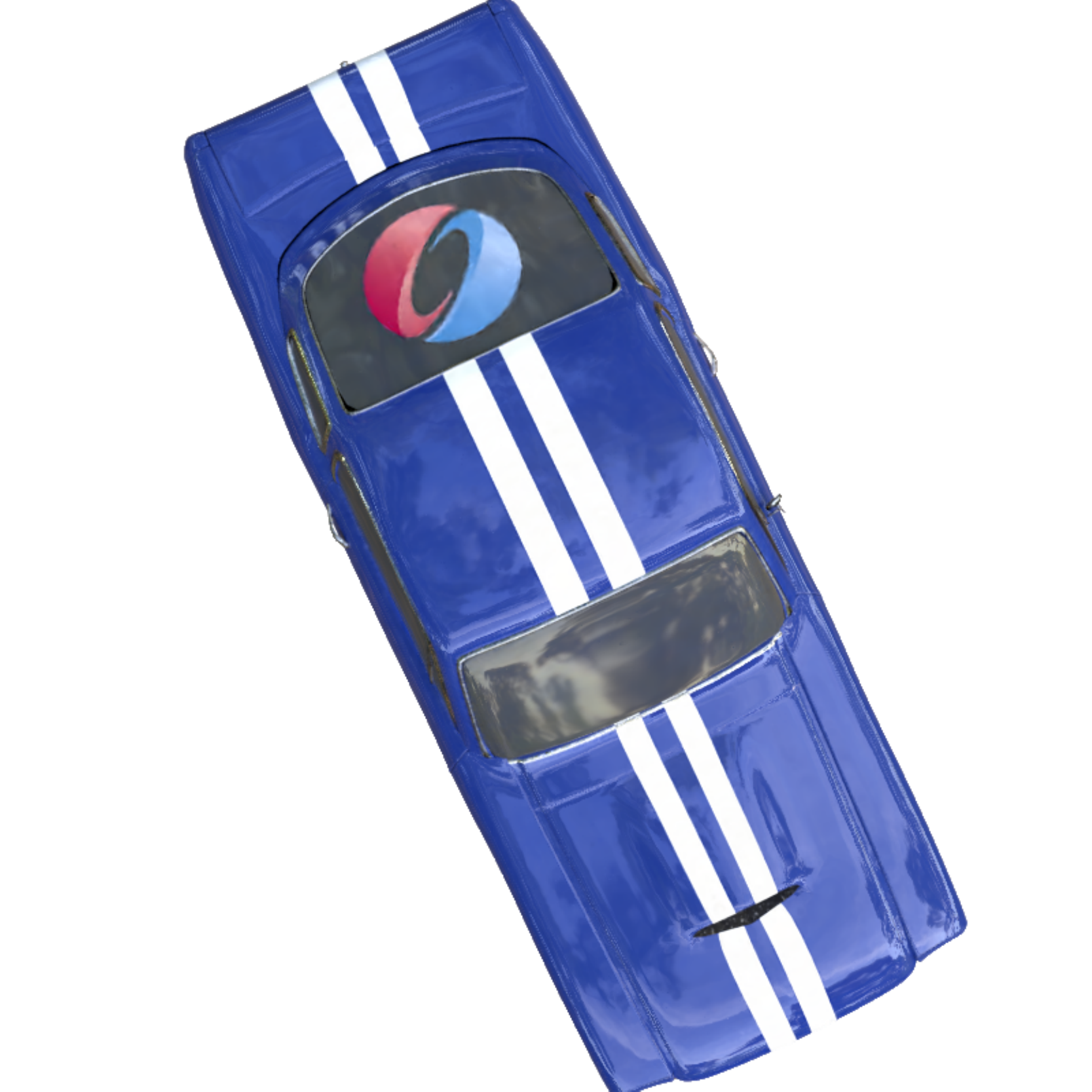}}&
            \hspace{-3mm}{\includegraphics[width=0.19\linewidth]{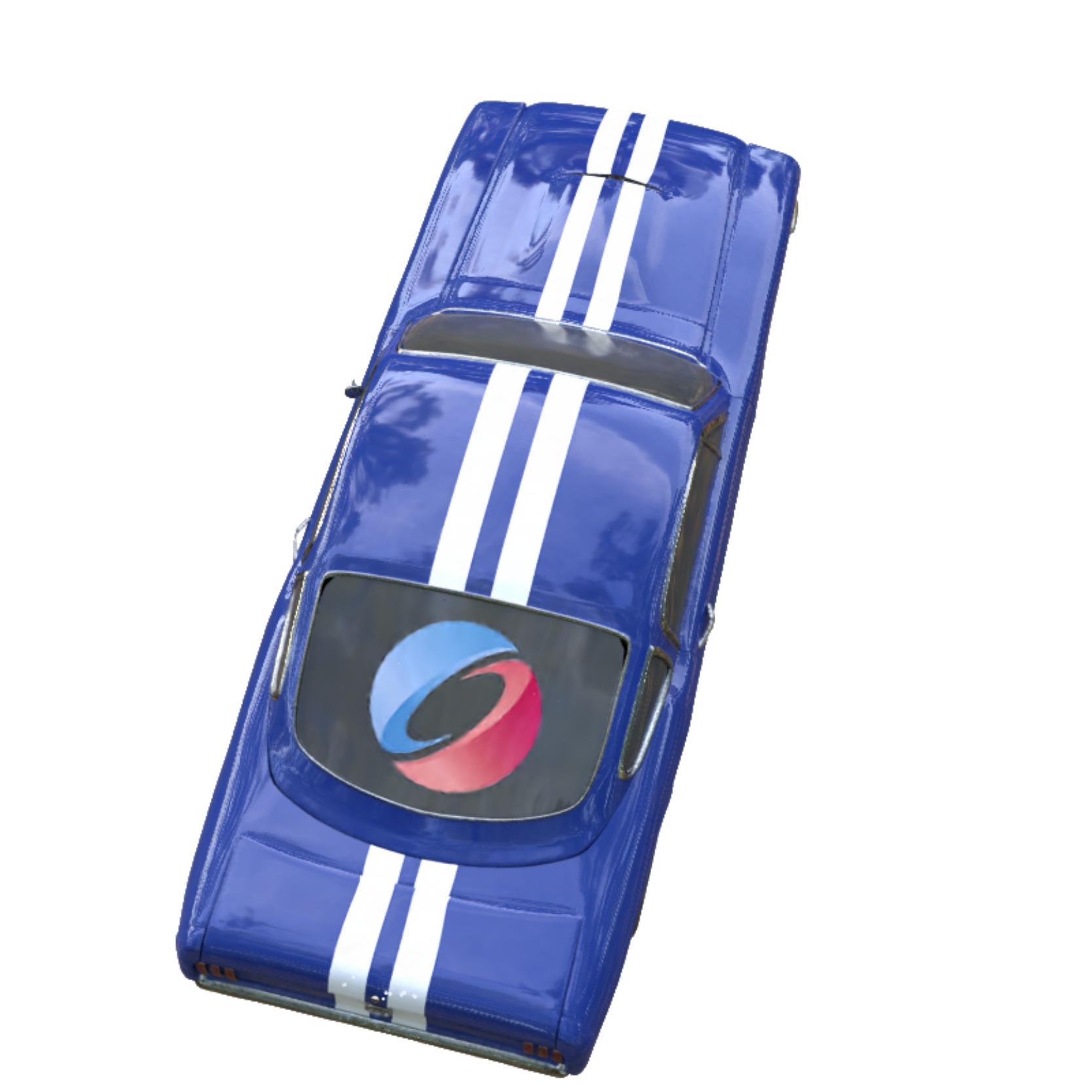}}
            \\
            \hspace{-4mm}
            \rotatebox{90}{\quad \hspace{-2mm} {\small Texture}}
            &
            \hspace{-3mm}{\includegraphics[width=0.19\linewidth]{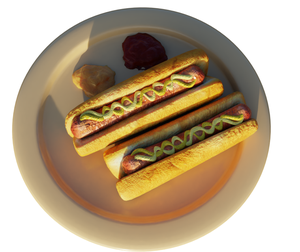}}&
            \hspace{-3mm}{\includegraphics[width=0.19\linewidth]{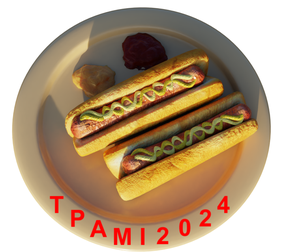}}&
            \hspace{-3mm}{\includegraphics[width=0.19\linewidth]{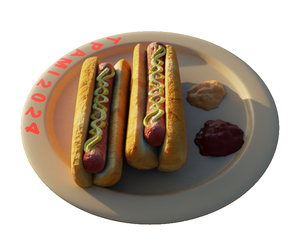}}&
            \hspace{-3mm}{\includegraphics[width=0.19\linewidth]{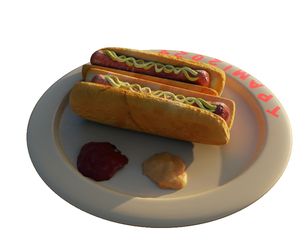}}&
            \hspace{-3mm}{\includegraphics[width=0.19\linewidth]{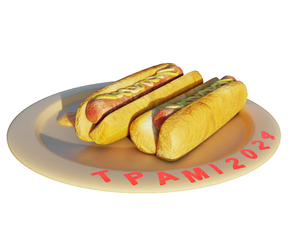}}
            \\
            \hspace{-4mm}
            \rotatebox{90}{\quad \hspace{-2mm} {\small Texture}}
            &
            \hspace{-3mm}{\includegraphics[width=0.19\linewidth]{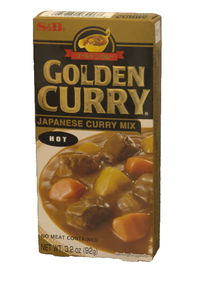}}&
            \hspace{-3mm}{\includegraphics[width=0.19\linewidth]{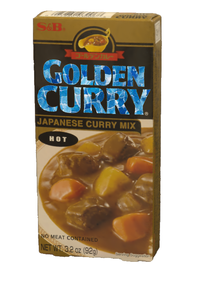}}&
            \hspace{-3mm}{\includegraphics[width=0.19\linewidth]{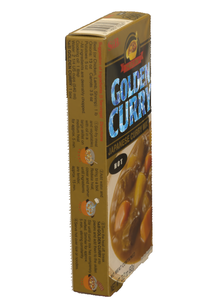}}&
            \hspace{-3mm}{\includegraphics[width=0.19\linewidth]{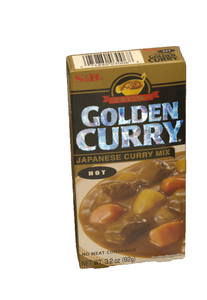}}&
            \hspace{-3mm}{\includegraphics[width=0.19\linewidth]{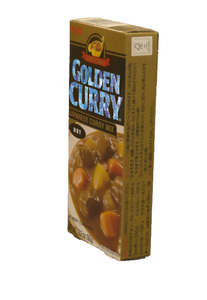}}
            \\
            &
            \hspace{-4mm}\makecell{Input}&
            \hspace{-4mm}\makecell{Edit}&
            \hspace{-4mm}\makecell{View \#1}&
            \hspace{-4mm}\makecell{View \#2}&
            \hspace{-4mm}\makecell{View \#3}
        \end{tabular}
    }
    \caption{
        Geometry editing and texture editing results. 
        For each row, we show input and edited geometry/image in the first two columns, and in the \yl{remaining} three columns we show edited \yl{Gaussians} rendered from three different viewpoints. 
    }
    \label{fig:geo_texture_edit}
\end{figure}

We show relighting results and compare with NeRFactor~\cite{NeRFactor}, \wttp{TensoIR}~\cite{TensoIR}, NDR~\cite{NvDiffRec}, NeRO~\cite{NeRO}, RelightableGaussian~\cite{RelightableGaussian}, and GuassianShader ~\cite{GaussianShader} in Fig.~\ref{fig:relight}. 
Limited by the lighting representations, NeRFactor and \wttp{TensoIR} are unable to render high-frequency reflections. 
The irregular geometry of NDR makes the scene shaded in a wrong way. 
\wt{
NeRO can produce realistic reflection but its material estimation is less accurate, leading to less faithful relighting (third to fourth \yl{rows}). 
In addition, it directly uses the mesh extracted by an SDF network for relighting, which may produce level-set stripe-like artifacts (better viewed in video) or miss geometry details like the metal frame on the side window in the car example.
}
GaussianShader and RelightableGaussian produce less faithful normal and texture estimation in the decomposition process and the \textit{forward rendering} technique they apply causes blending artifacts even if the normal estimation is plausible (\wt{zoom-in views of the car example in Fig.~\ref{fig:relight}}). 
In contrast, the relighting results produced by our method are more realistic. 
This is because the geometry of \name distilled from a signed distance function, which works as a good basis for the geometry, texture and lighting decomposition. 
In addition, \name uses the \textit{deferred shading}, which avoids blending artifacts when rendering \yl{Gaussians} in a new illumination. 
Quantitative results on \yl{the} relighting task are reported in Table~\ref{tab:relight} and our method achieves comparable results with NeRO~\cite{NeRO}. 
We show geometry and texture editing results in Fig.~\ref{fig:geo_texture_edit}. 
\yl{For texture} editing results in \yl{the} last two rows, the editing is performed on the diffuse component. 

\subsection{Ablation}
In this subsection, we conduct several ablation studies to prove the effectiveness of our design choices. 

\subsection{Normal Field Distillation}
\begin{figure}[t]
        \centering
        {
            \begin{tabular}{ccc}
                \hspace{-2mm}
                {\includegraphics[width=0.19\linewidth]{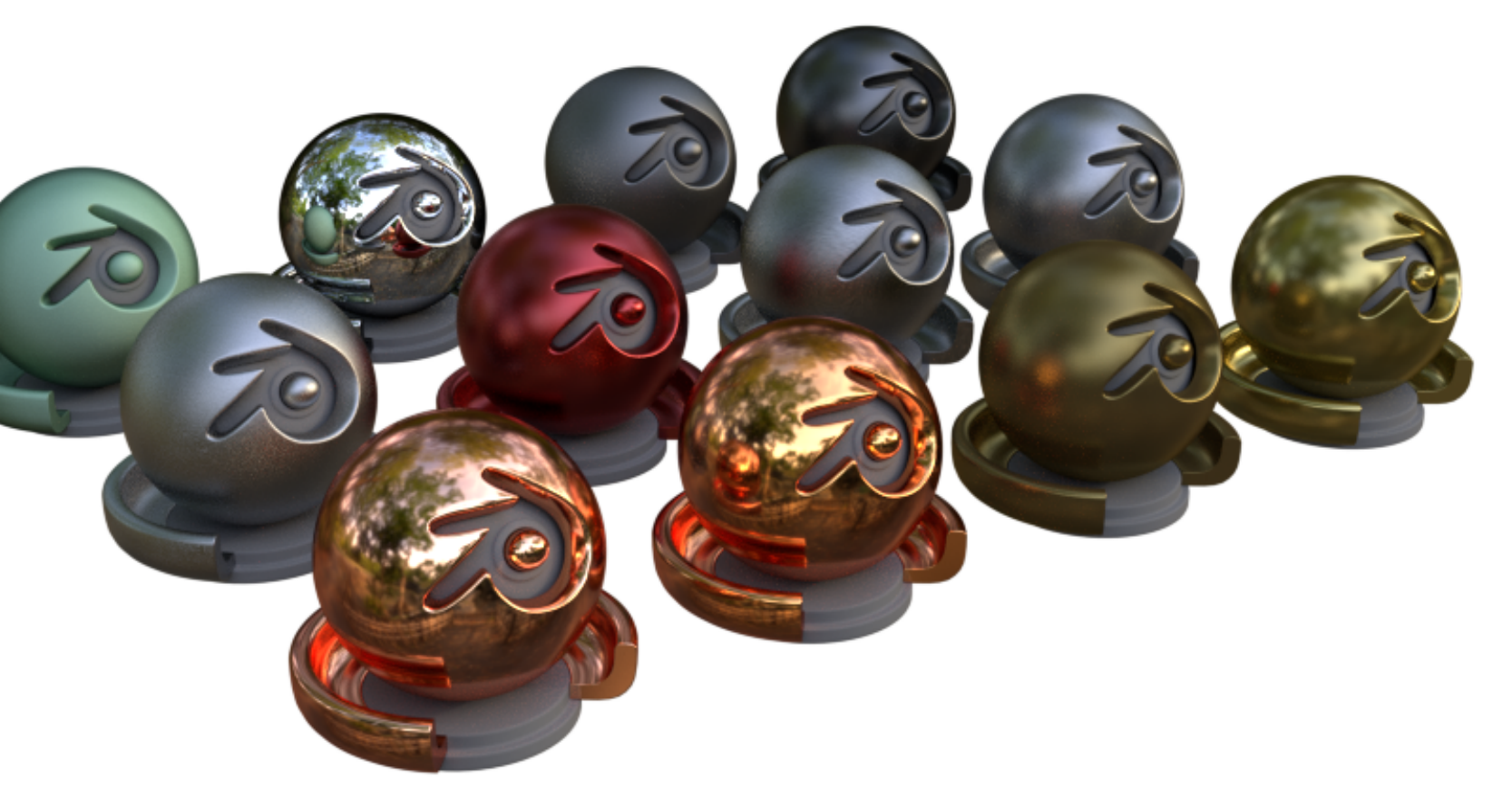}}
                {\includegraphics[width=0.19\linewidth]{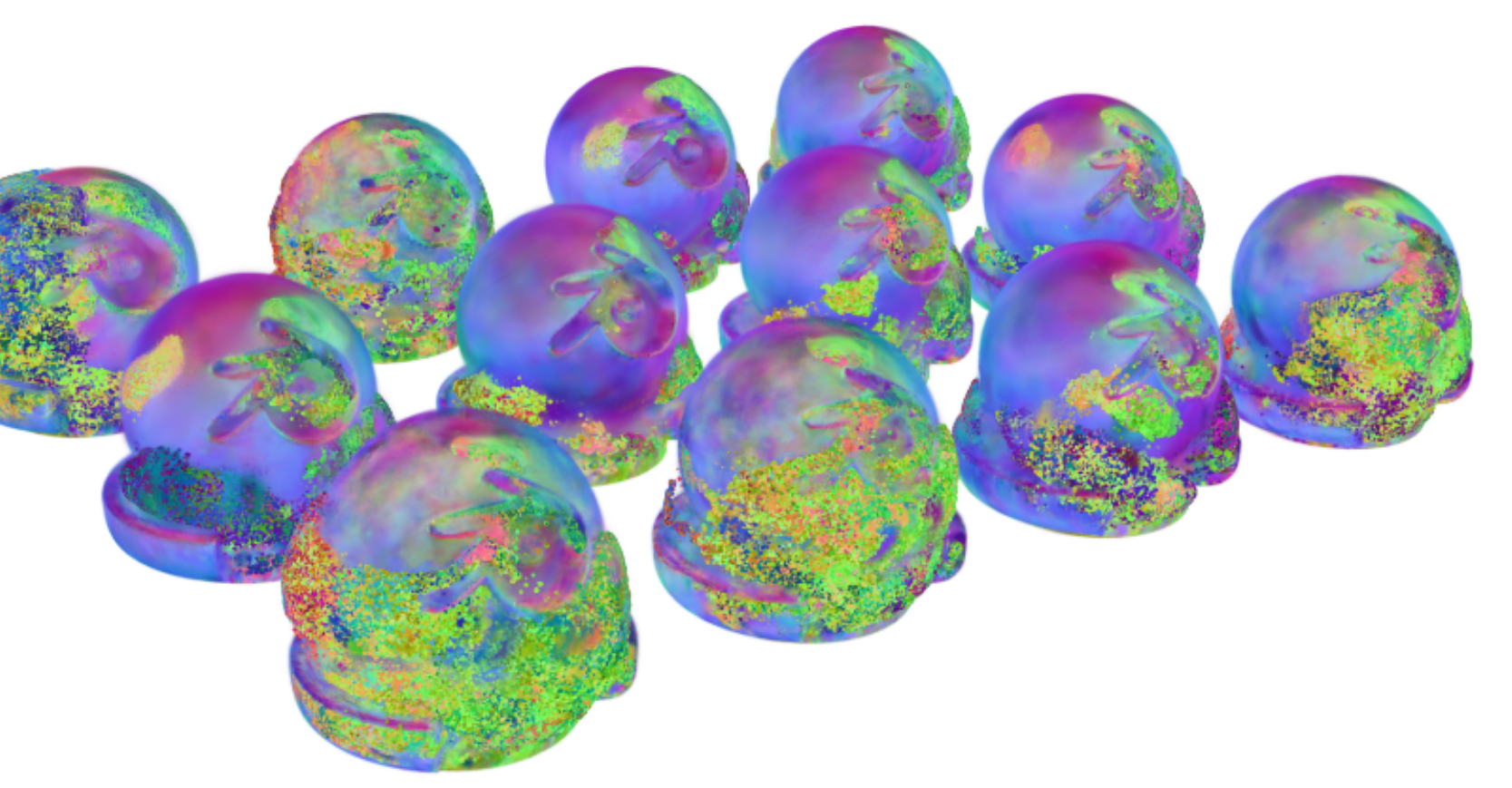}}
                {\includegraphics[width=0.19\linewidth]{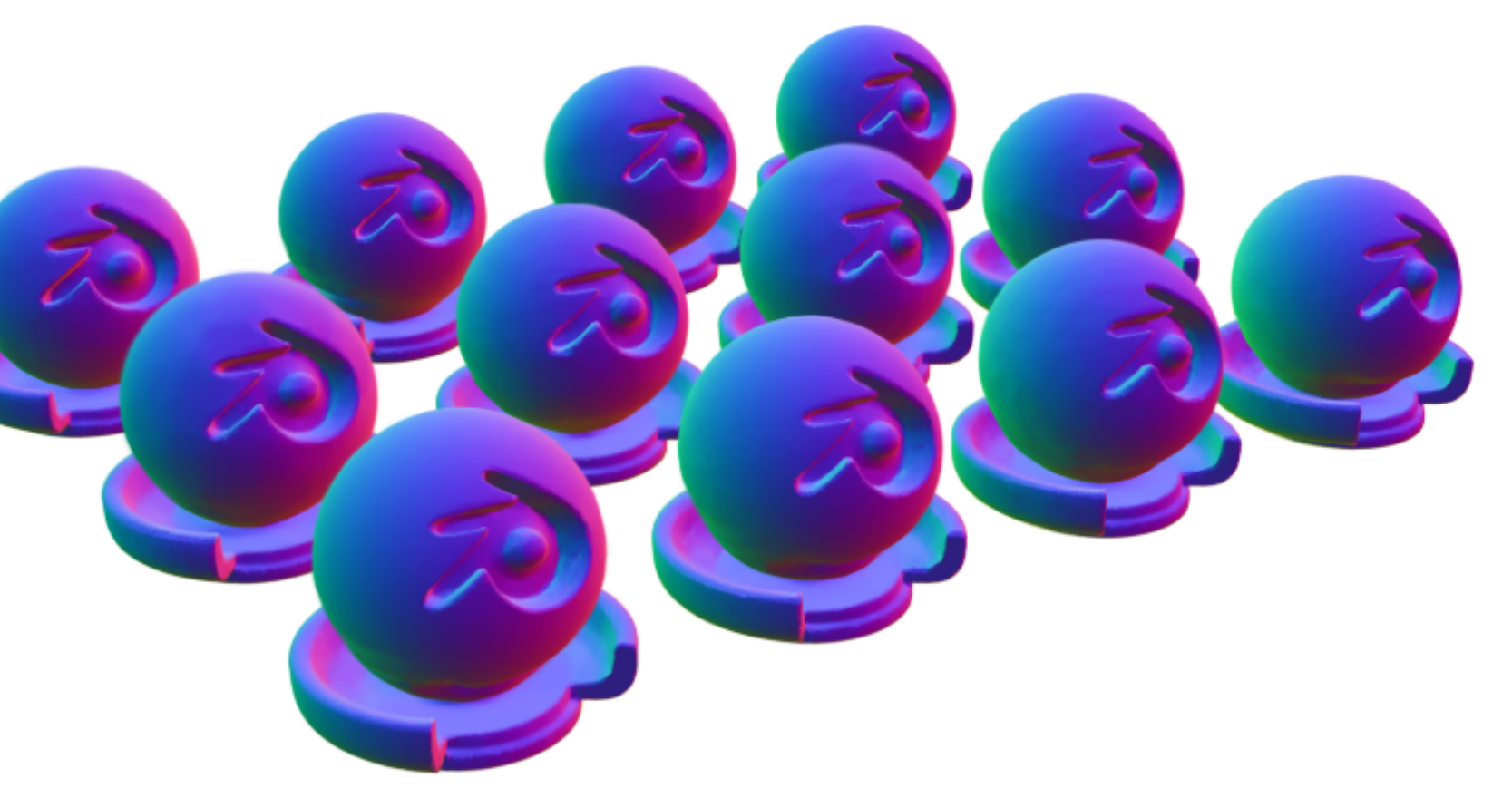}}
                {\includegraphics[width=0.19\linewidth]{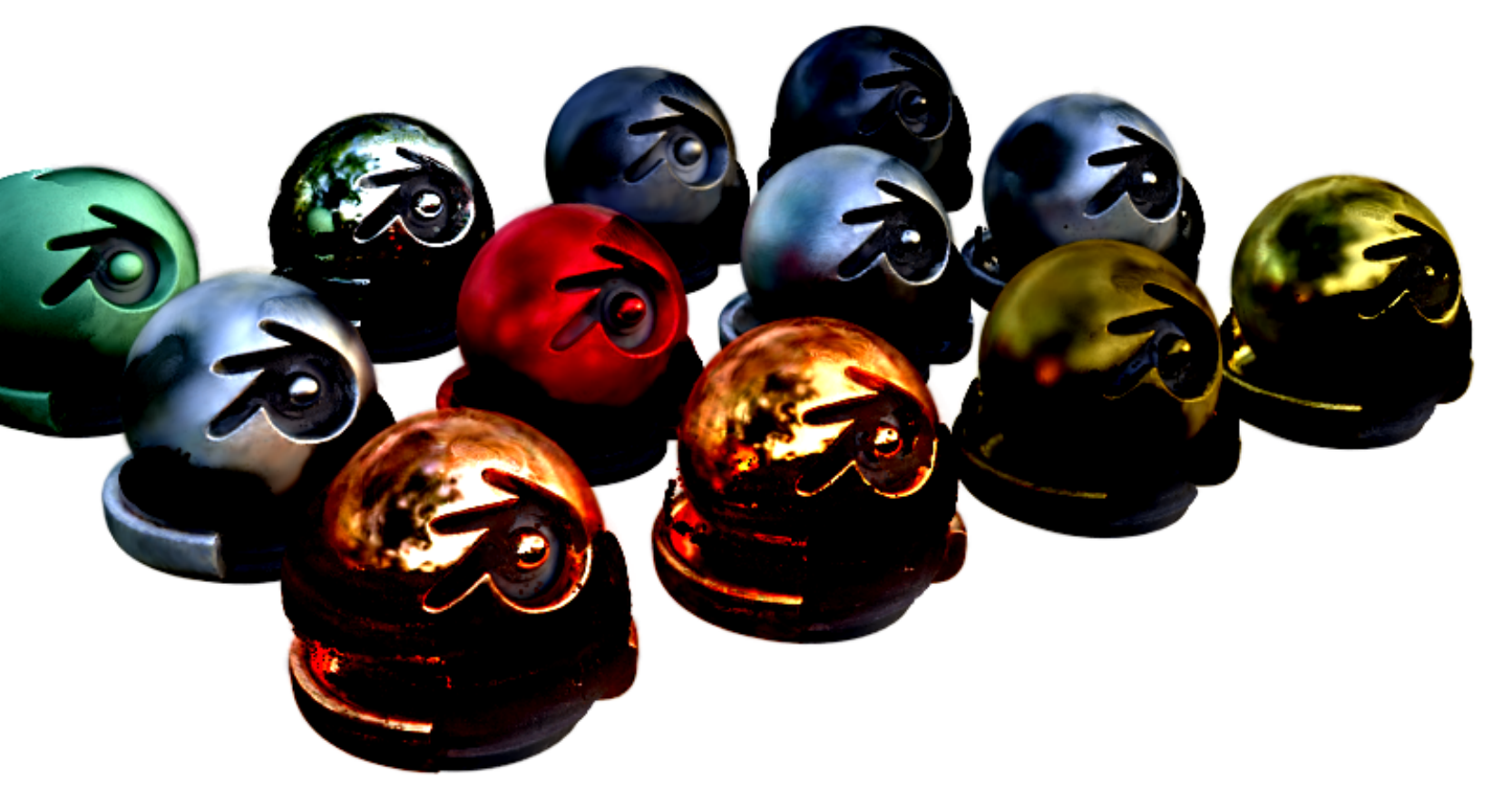}}
                {\includegraphics[width=0.19\linewidth]{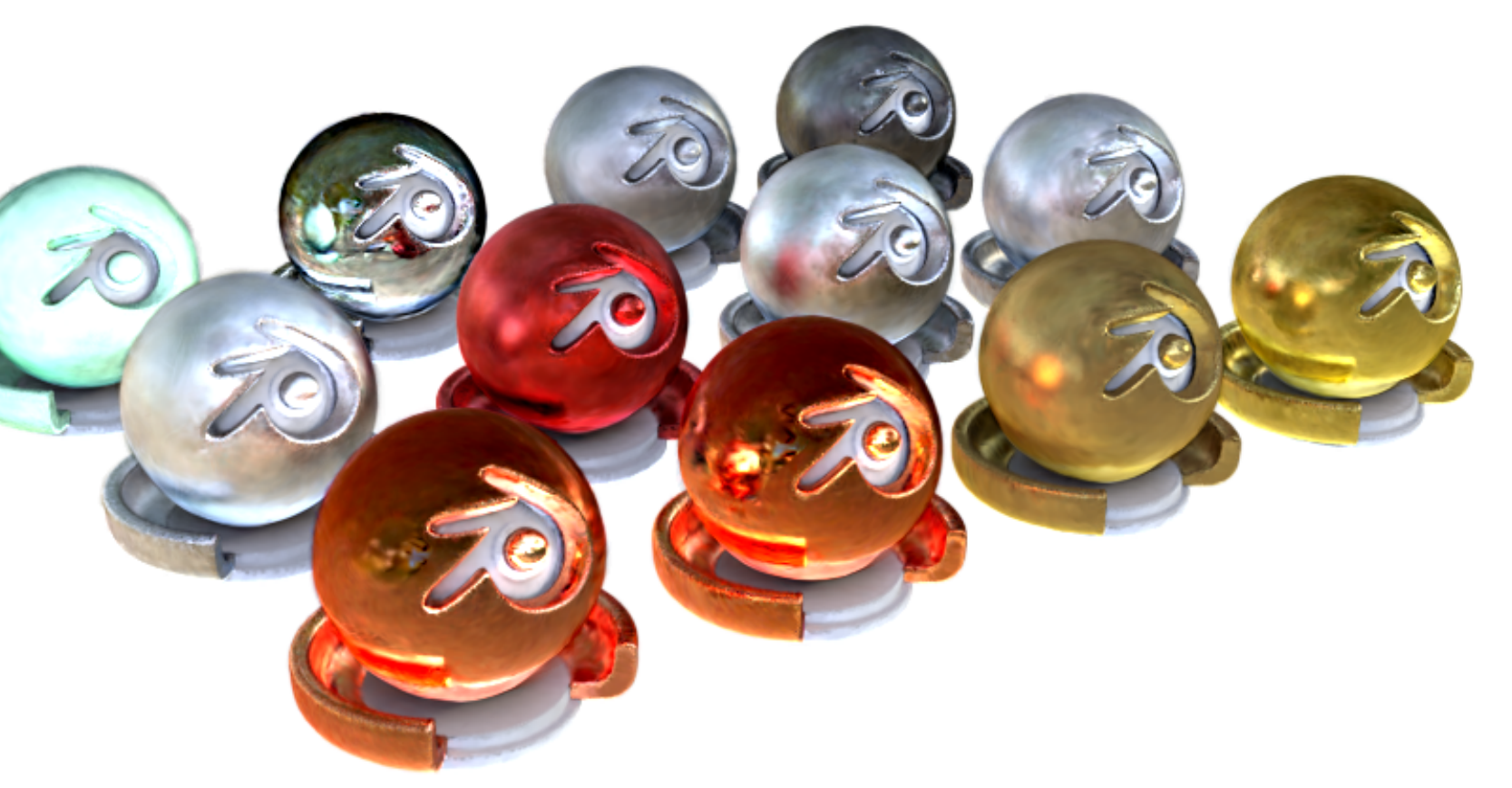}}
                \\
                \hspace{-2mm}
                \subcaptionbox{\scriptsize Input}{\includegraphics[width=0.19\linewidth]{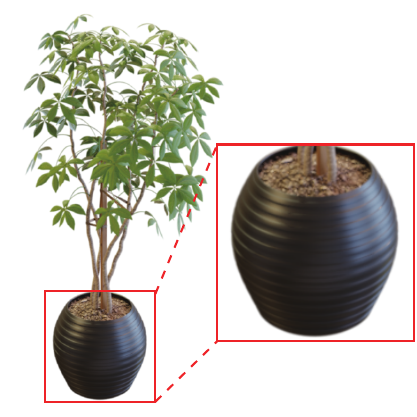}}
                \subcaptionbox{\scriptsize w/o NFD}{\includegraphics[width=0.19\linewidth]{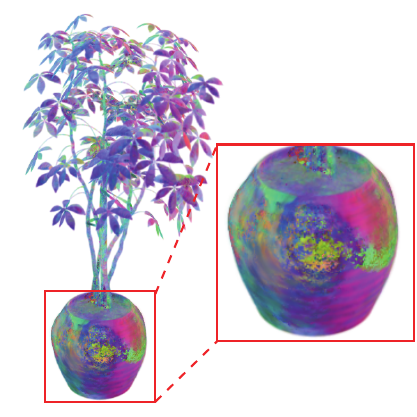}}
                \subcaptionbox{\scriptsize w/ NFD}{\includegraphics[width=0.19\linewidth]{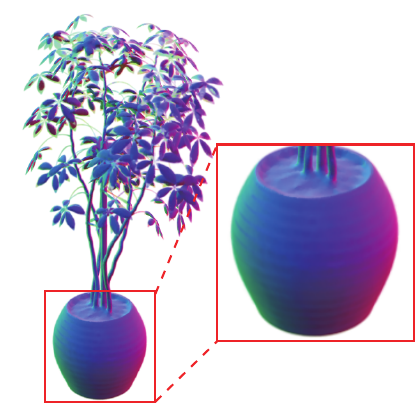}}
                \subcaptionbox{\scriptsize w/o NFD}{\includegraphics[width=0.19\linewidth]{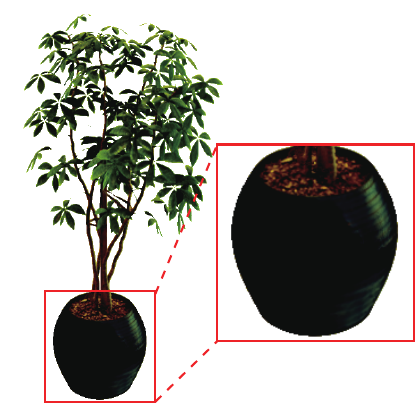}}
                \subcaptionbox{\scriptsize w/ NFD}{\includegraphics[width=0.19\linewidth]{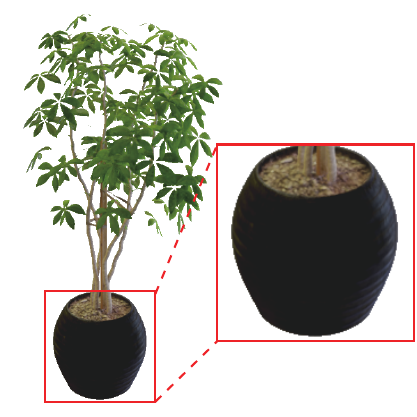}}
            \end{tabular}
        }
        \caption{
            Qualitative comparisons between decomposed normals and diffuse albedos with normal field distillation (w/ NFD) and without normal field distillation (w/o NFD). 
        }
        \label{fig:ablate_NFD}
\end{figure}

To produce a plausible geometry for the decomposition problem, we distill the normal direction from a signed distance function by encouraging the rendered normal of \yl{Gaussians} to be consistent with the normal of a signed distance function. 
Here, we generate the decomposed results without normal field distillation (w/o NFD) and compare them with the full pipeline qualitatively in Fig.~\ref{fig:ablate_NFD}. 
We also report numerical results of decomposed results in Table~\ref{tab:decom} and the decomposition quality sees a drop without normal field distillation. 

\subsection{Deferred Shading}
\begin{figure}[t]
    \centering
    {
        \begin{tabular}{c}
            \hspace{-4mm}
            {\includegraphics[width=0.16\linewidth]{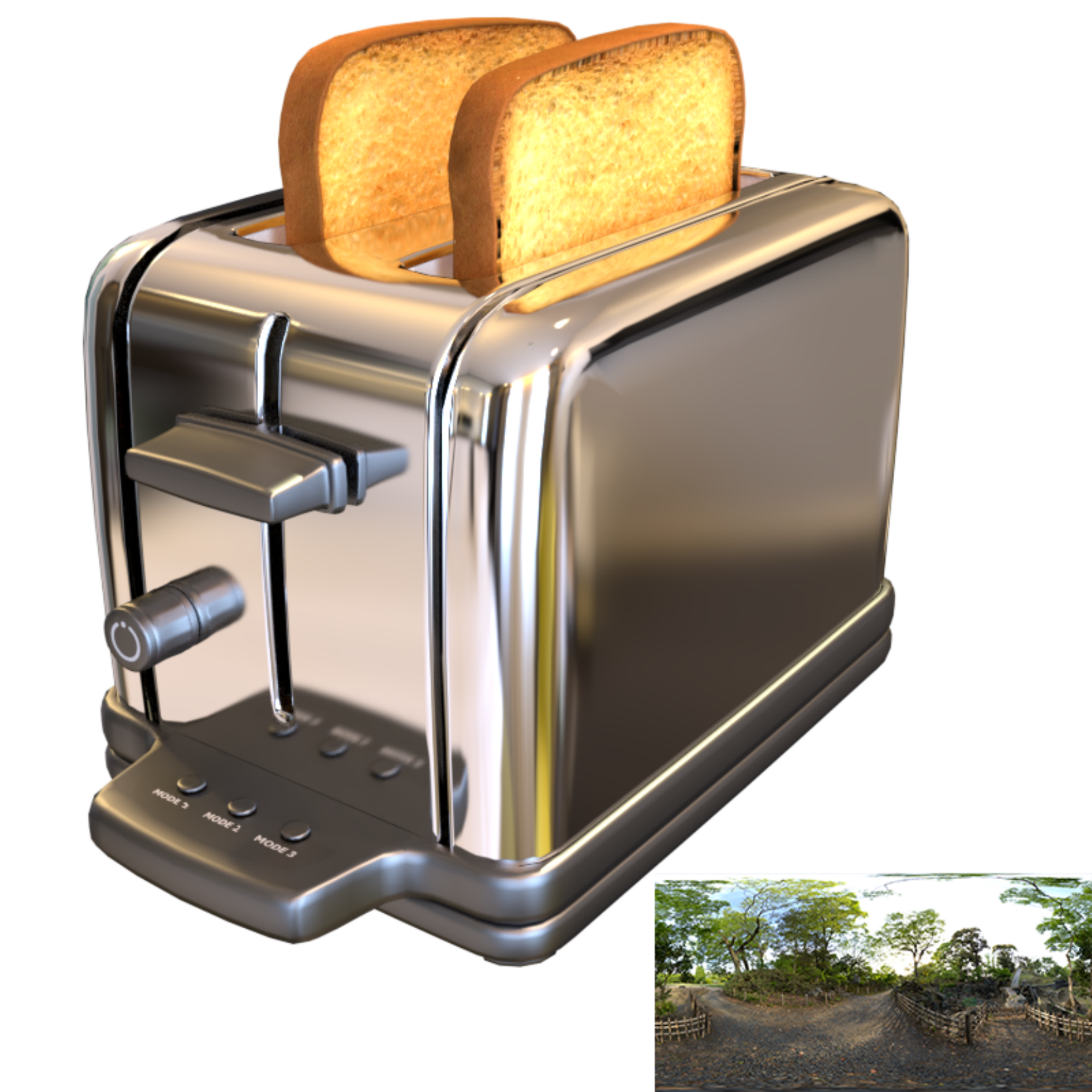}}
            {\includegraphics[width=0.16\linewidth]{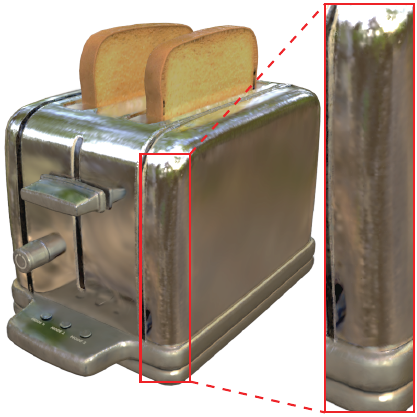}}
            {\includegraphics[width=0.16\linewidth]{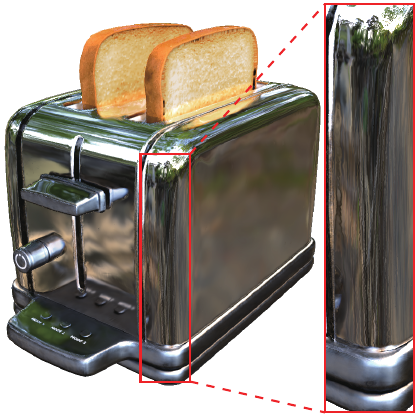}}
            {\includegraphics[width=0.16\linewidth]{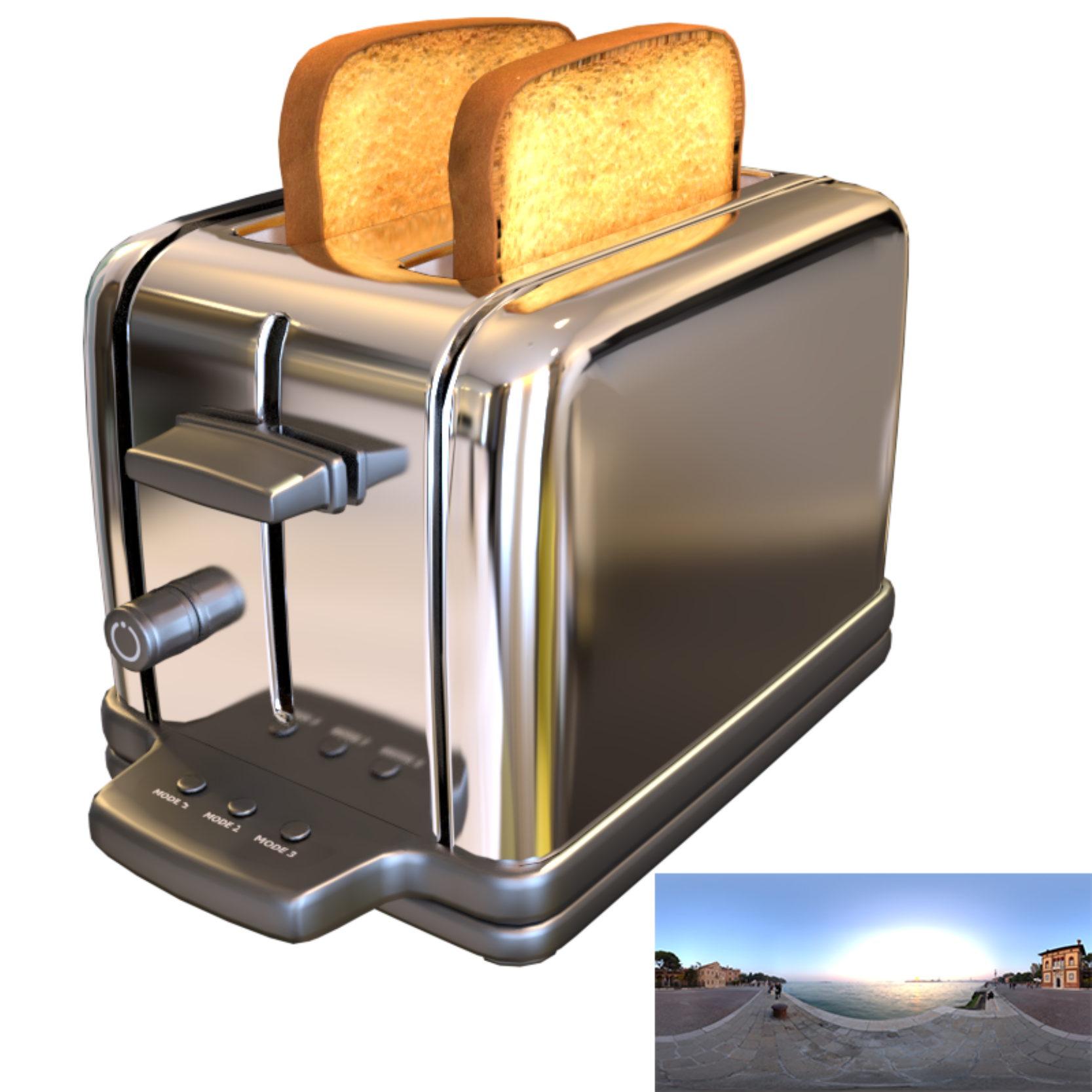}}
            {\includegraphics[width=0.16\linewidth]{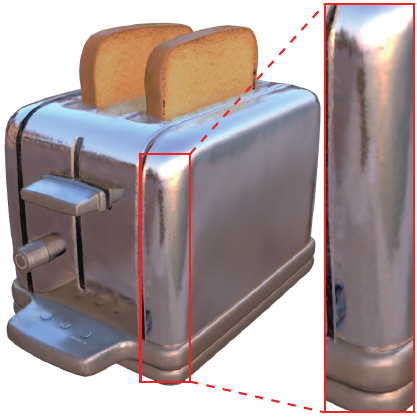}}
            {\includegraphics[width=0.16\linewidth]{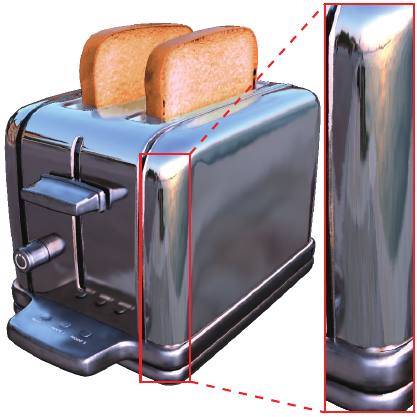}}
            \\
            \hspace{-4mm}
            {\includegraphics[width=0.16\linewidth]{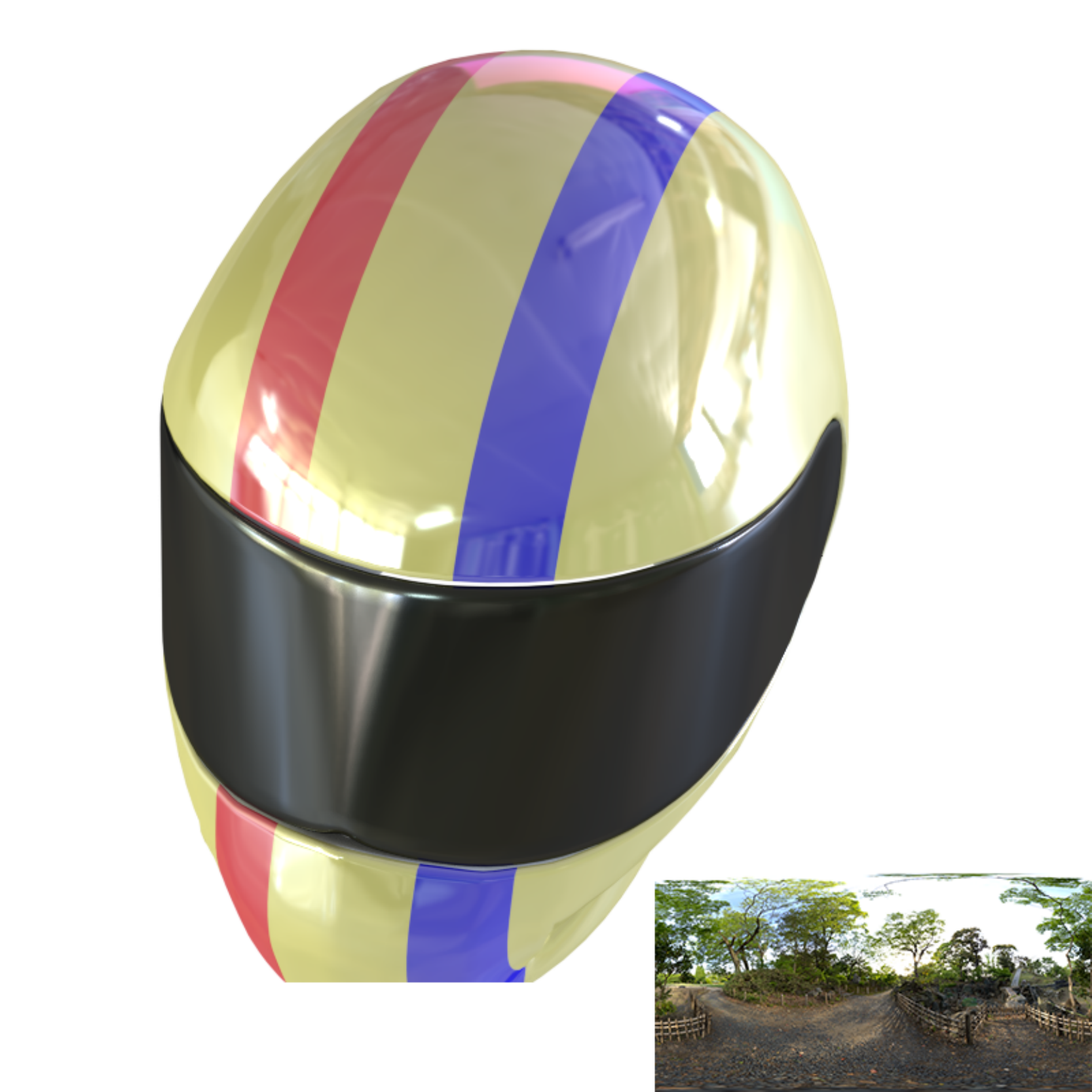}}
            {\includegraphics[width=0.16\linewidth]{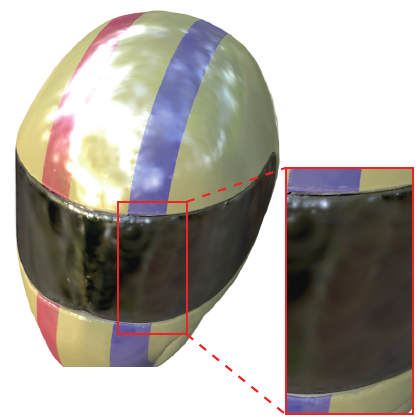}}
            {\includegraphics[width=0.16\linewidth]{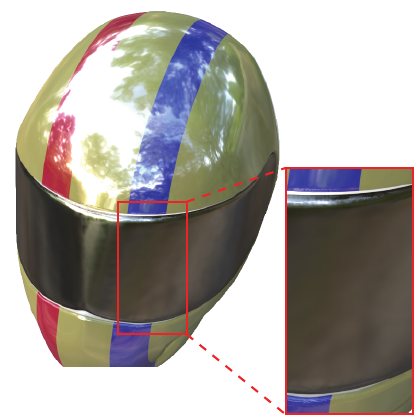}}
            {\includegraphics[width=0.16\linewidth]{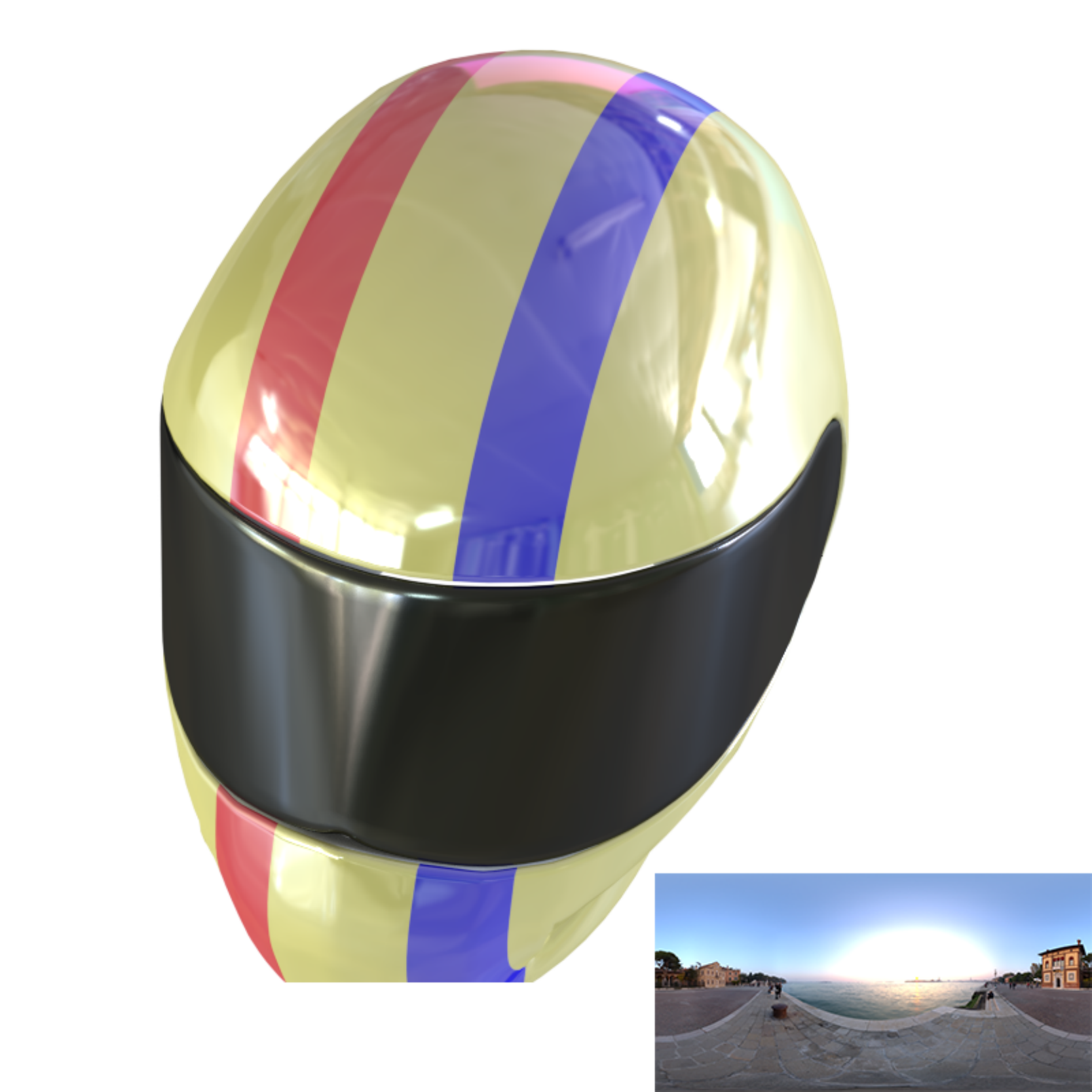}}
            {\includegraphics[width=0.16\linewidth]{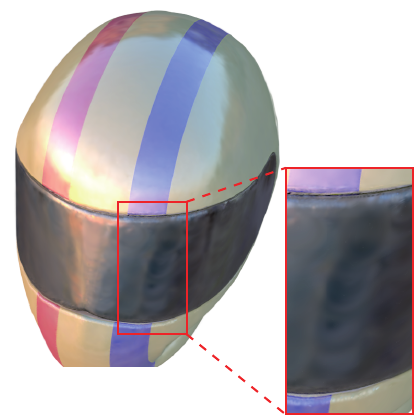}}
            {\includegraphics[width=0.16\linewidth]{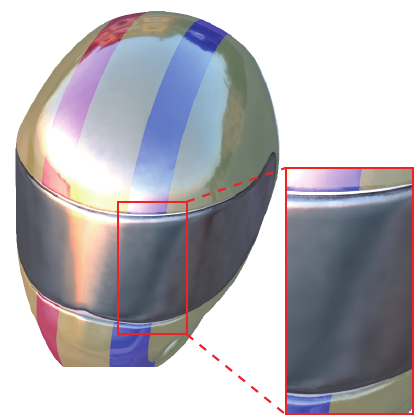}}
            \\
            \hspace{-4mm}
            {\includegraphics[width=0.16\linewidth]{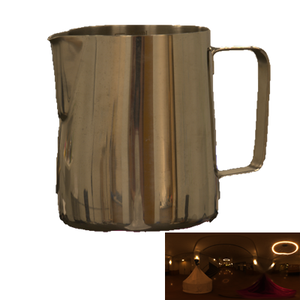}}
            {\includegraphics[width=0.16\linewidth]{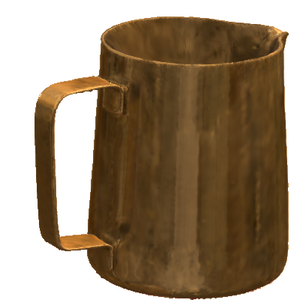}}
            {\includegraphics[width=0.16\linewidth]{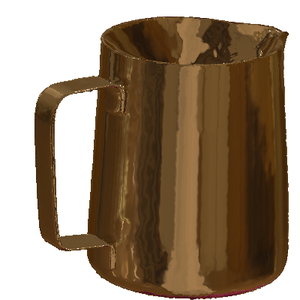}}
            {\includegraphics[width=0.16\linewidth]{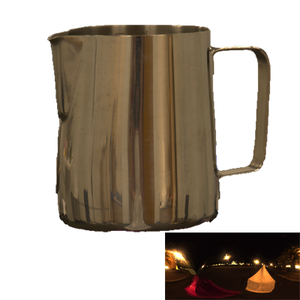}}
            {\includegraphics[width=0.16\linewidth]{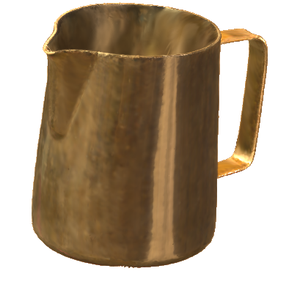}}
            {\includegraphics[width=0.16\linewidth]{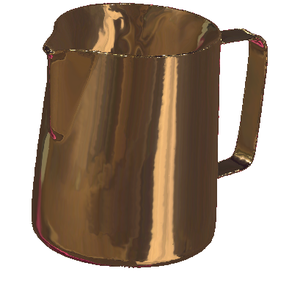}}
            \\
            \hspace{-4mm}
            \subcaptionbox{Input}{\includegraphics[width=0.16\linewidth]{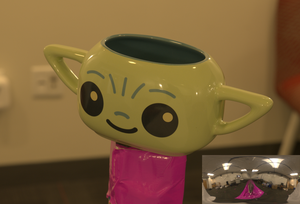}}
            \subcaptionbox{F.S.}{\includegraphics[width=0.16\linewidth]{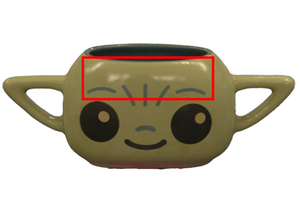}}
            \subcaptionbox{D.S.}{\includegraphics[width=0.16\linewidth]{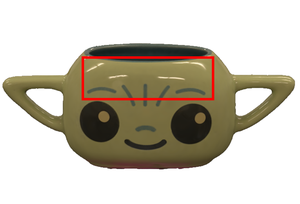}}
            \subcaptionbox{Input}{\includegraphics[width=0.16\linewidth]{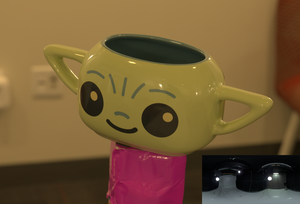}}
            \subcaptionbox{F.S.}{\includegraphics[width=0.16\linewidth]{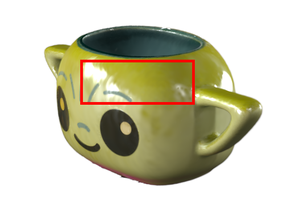}}
            \subcaptionbox{D.S.}{\includegraphics[width=0.16\linewidth]{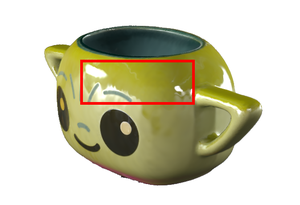}}
        \end{tabular}
    }
    \caption{
        Qualitative comparisons between relighting results with \textit{forward shading} (F.S.) and with \textit{deferred shading} (D.S.). 
        The relit scene contains blending artifacts when we use \textit{forward shading} while \textit{deferred shading} can avoid them. 
        The bottom two rows are from the real Stanford ORB dataset~\cite{Stanford-ORB}.
    }
     \vspace{-3mm}
    \label{fig:ablate_DS}
\end{figure}

In the decomposition and relighting process, we utilize the \textit{deferred shading} technique instead \yl{of} \textit{forward shading} commonly used in previous methods. 
Here we ablate its influence on the relighting results in Fig.~\ref{fig:ablate_DS}. 
\wttp{Even \yl{though} the geometry is enhanced by the normal field distillation, we can \yl{still} observe notable Gaussian shape---like blending artifacts when we apply the \textit{forward shading} technique.} 
By contrast, the \textit{deferred shading} technique we use can successfully resolve this issue and produce smooth and realistic relighting results. 
We also evaluate it quantitatively in Table~\ref{tab:relight} and the relighting quality drops when it is not applied. 

\begin{figure}[!h]
    \centering
    {
        \begin{tabular}{c}
            \hspace{-4mm}
            {\includegraphics[width=0.16\linewidth]{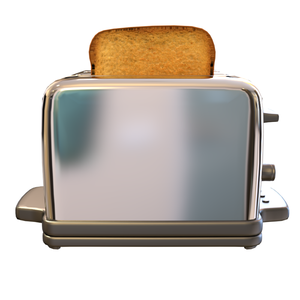}}
            {\includegraphics[width=0.16\linewidth]{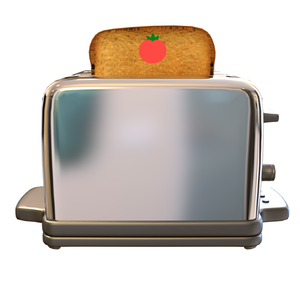}}
            {\includegraphics[width=0.16\linewidth]{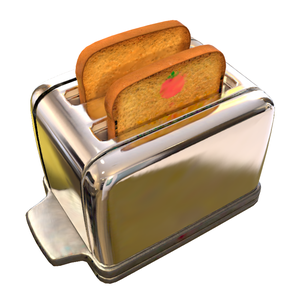}}
            {\includegraphics[width=0.16\linewidth]{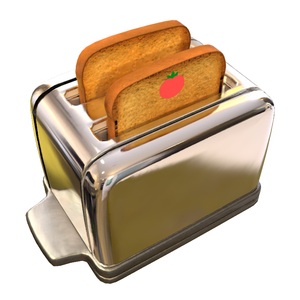}}
            {\includegraphics[width=0.16\linewidth]{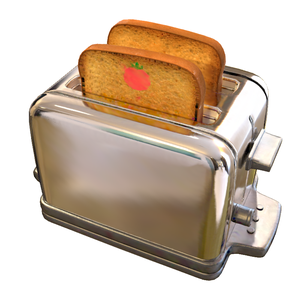}}
            {\includegraphics[width=0.16\linewidth]{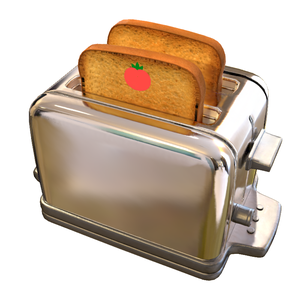}}
            \\
            \hspace{-4mm}
            \subcaptionbox{\scriptsize Input}{\includegraphics[width=0.16\linewidth]{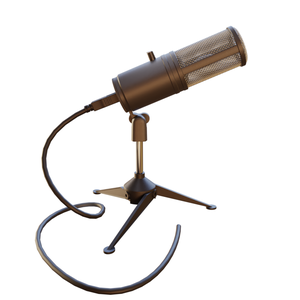}}
            \subcaptionbox{\scriptsize Edit}{\includegraphics[width=0.16\linewidth]{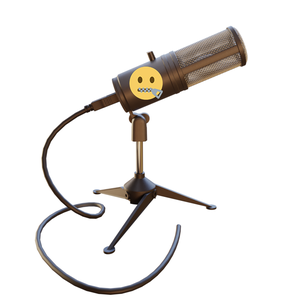}}
            \subcaptionbox{\scriptsize w/o R.V.}{\includegraphics[width=0.16\linewidth]{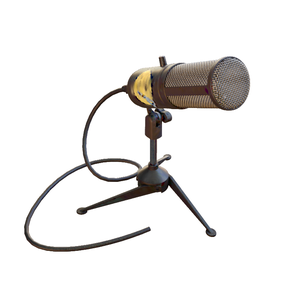}}
            \subcaptionbox{\scriptsize w/ R.V.}{\includegraphics[width=0.16\linewidth]{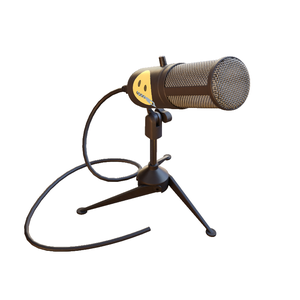}}
            \subcaptionbox{\scriptsize w/o R.V.}{\includegraphics[width=0.16\linewidth]{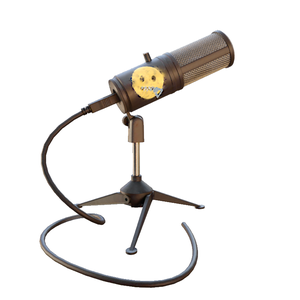}}
            \subcaptionbox{\scriptsize w/ R.V.}{\includegraphics[width=0.16\linewidth]{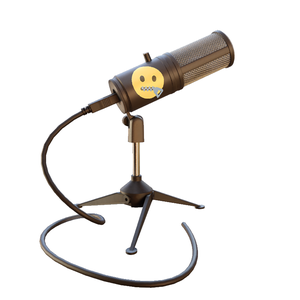}}
        \end{tabular}
    }
    \caption{
        \wttp{
        Qualitative comparisons between texture editing results with rendered results from random viewpoints (w/ R.V.) and without (w/o R.V.). 
        The rendered results of edited gaussians splatting from novel viewpoints can exhibit inconsistency without inputting rendered results from random viewpoints. 
        }
    }
    \label{fig:ablate_texture_edit}
\end{figure}

\wttp{
\subsection{Random View Inputs in Texture Editing}
As mentioned in Sec.~\ref{sec:texture_edit}, we perform texture editing on the Gaussian splatting representation by optimizing Gaussians' attributes to fit the rendered results to both the input edited image and randomly rendered images from different viewpoints. 
Here we ablate how randomly generated edited results influence the final rendered results of edited Gaussians in Fig.~\ref{fig:ablate_texture_edit}. 
It shows that optimizing Gaussians' attributes only from the input edited viewpoint can result in overfitting and exhibit inconsistent novel views. 
}

\begin{figure}[!h]
    \centering
    {
        \begin{tabular}{cc|cc}
            \hspace{-2mm}\subcaptionbox{\small Input}{\includegraphics[width=0.24\linewidth]{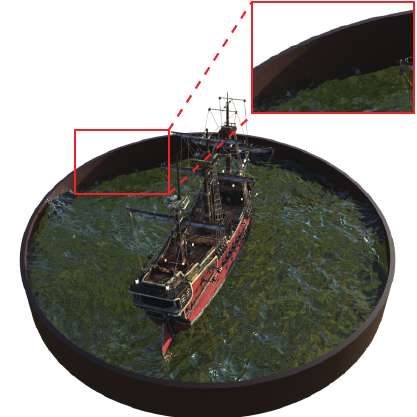}}&
            \hspace{-2mm}\subcaptionbox{\small Diffuse Albedo}{\includegraphics[width=0.24\linewidth]{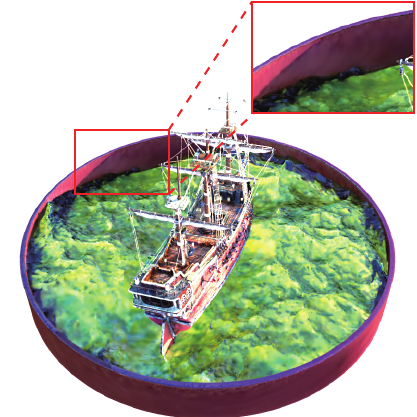}}&
            \hspace{-2mm}\subcaptionbox{\small Edit}{\includegraphics[width=0.24\linewidth]{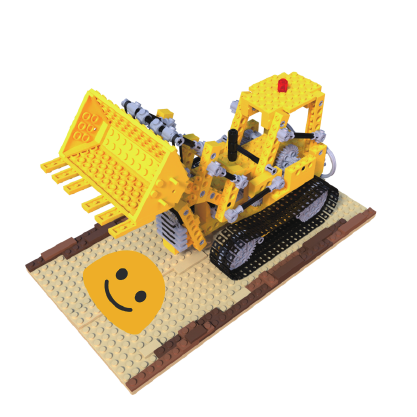}}&
            \hspace{-2mm}\subcaptionbox{\small Novel View}{\includegraphics[width=0.24\linewidth]{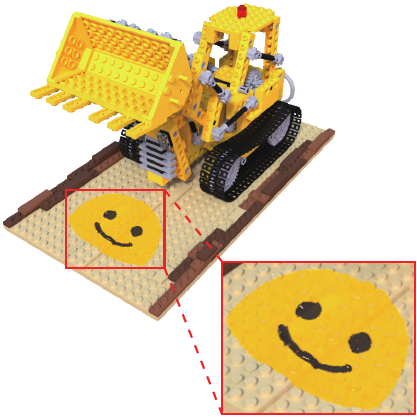}}
        \end{tabular}
    }
    \caption{
        Limitations. \name bakes shadow into texture on scenes containing shadow ((a)-(b)) and may result in noises after texture editing ((c)-(d)).
    }
    \label{fig:limit}
\end{figure}

\section{Conclusion}
In this paper, we present \name, a \yl{Gaussian} splatting representation with editable geometry, texture and lighting. 
To make this happen, we define optimizable texture attributes on \yl{Gaussians} and an environment map for the physically based rendering process. 
To ensure proper geometry reconstruction, \name distills the normal field from previous multi-view reconstruction methods onto \yl{Gaussians'} learnable normal attributes. 
Notably, we propose to use the \textit{deferred rendering} technique to enable realistic relighting. 
However, as shown in Fig.~\ref{fig:limit} our method may produce wrong decoupled results on scenes with shadow. 
For future directions, it is promising to resolve the ambiguity by capturing scenes under multiple lighting conditions and introducing visibility into the rendering process like~\cite{FEGR, SOL-NeRF}. 
On the editing side, the optimized \yl{Gaussians} after texture editing may contain \yl{noise} since \yl{Gaussian} splatting is a global representation where a pixel is influenced by multiple \yl{Gaussians}. 
Adaptive cloning and splitting in the edited region might produce better editing results.

\bibliographystyle{IEEEtran}
\bibliography{main}

\begin{IEEEbiography}[{\includegraphics[width=1in,height=1.25in,clip,keepaspectratio]{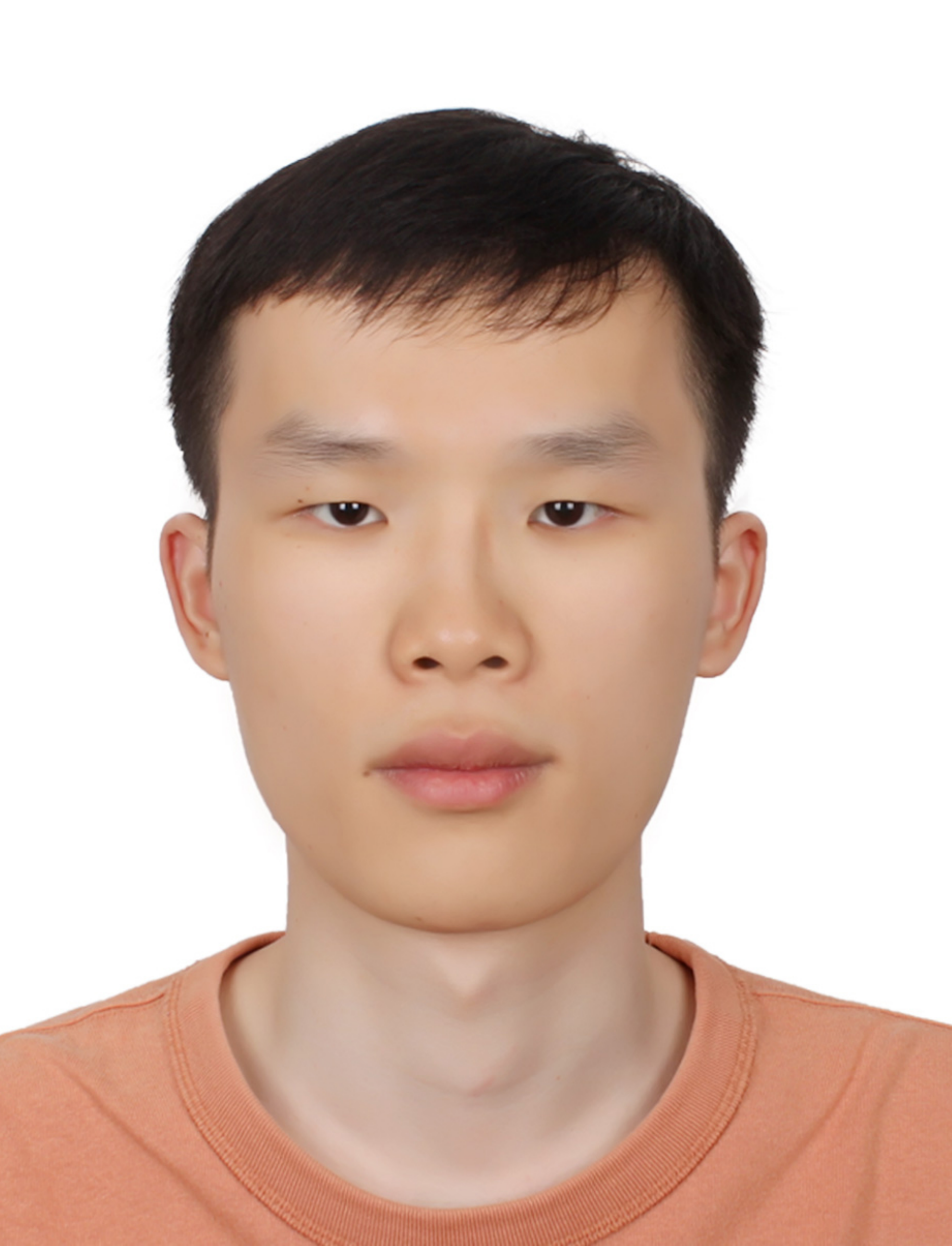}}]
    {Tong Wu} received his bachelor’s degree in computer science from Huazhong University of Science and Technology in 2019. He is currently a PhD candidate at the Institute of Computing Technology, Chinese Academy of Sciences. His research interests include computer graphics and computer vision. 
\end{IEEEbiography}

\begin{IEEEbiography}[{\includegraphics[width=1in,height=1.25in,clip,keepaspectratio]{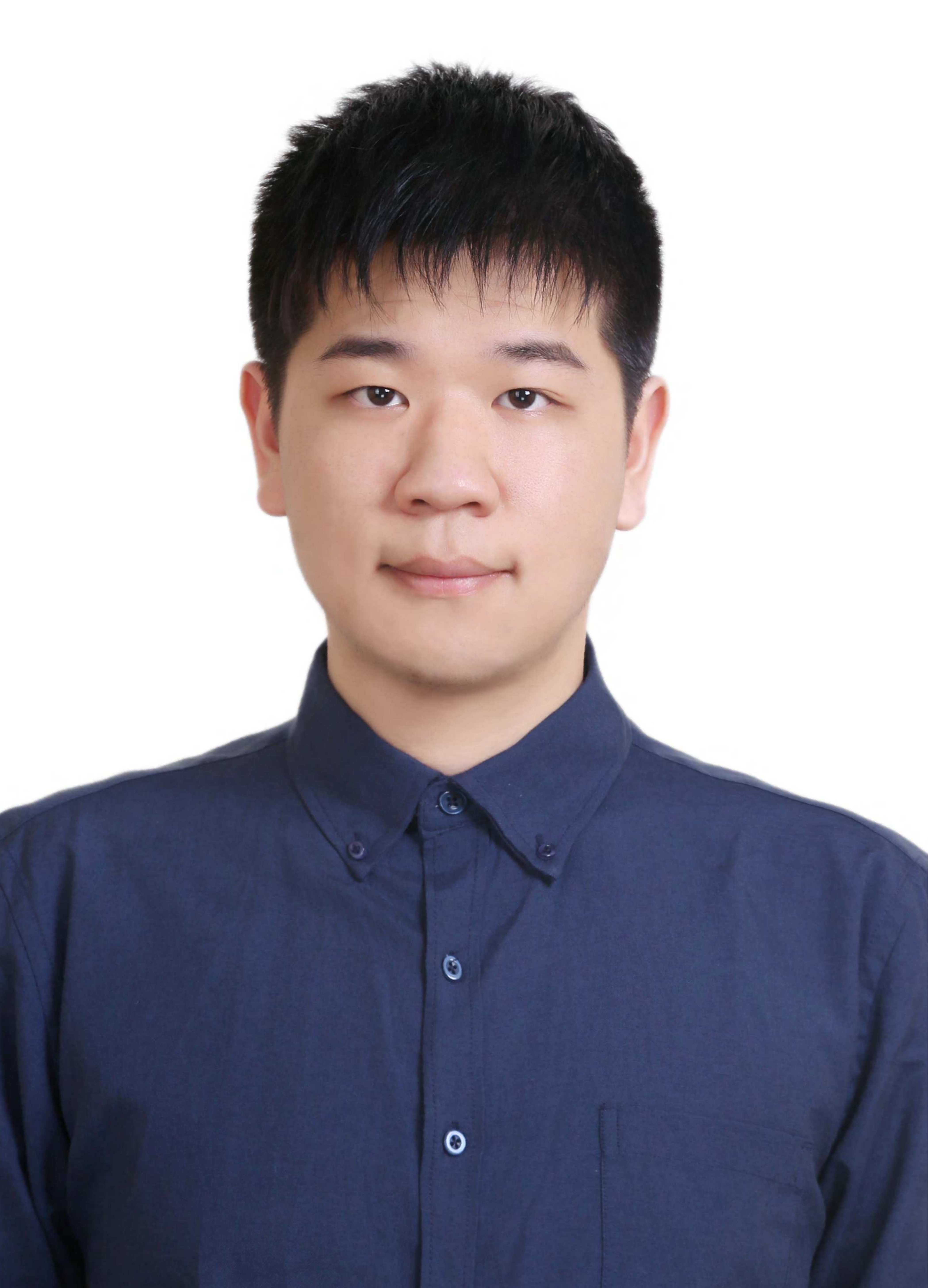}}]
    {Jia-Mu Sun} is a Master student of  Computer Science at Institute of Computing Technology, Chinese Academy of Sciences. He received his bachelor's degree from Huazhong University of Science and Technology. His research interests include geometry learning and rendering.
\end{IEEEbiography}

\begin{IEEEbiography}[{\includegraphics[width=1in,height=1.25in,clip,keepaspectratio]{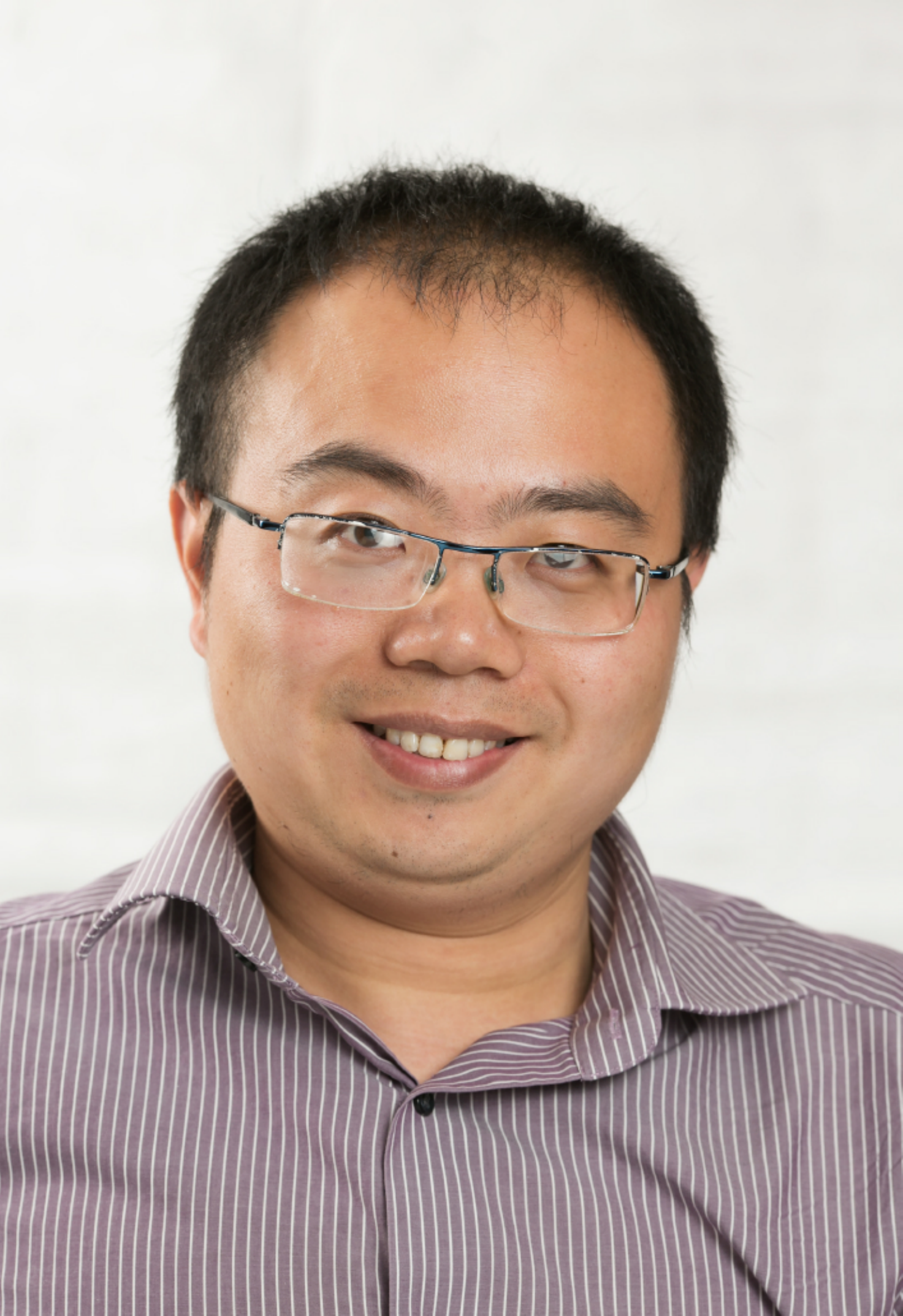}}]
    {Yu-Kun Lai} received his bachelor’s degree and PhD degree in computer science from Tsinghua University in 2003 and 2008, respectively. He is currently a Professor in the School of Computer Science \& Informatics, Cardiff University. His research interests include computer graphics, geometry processing, image processing and computer vision. He is on the editorial boards of {\em IEEE Transactions on Visualization and Computer Graphics} and {\em The Visual Computer}.
\end{IEEEbiography}

\begin{IEEEbiography}[{\includegraphics[width=1in,height=1.25in,clip,keepaspectratio]{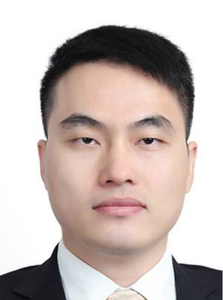}}]
    {Yuewen Ma} received his PhD degree from Nanyang Technological University, Singapore, in 2013. He has been engaged in the research and product of computer graphics and 3D vision for a long time. He is currently the leader of 3D reconstruction at ByteDance Pico.
\end{IEEEbiography}

\begin{IEEEbiography}[{\includegraphics[width=1in,height=1.25in,clip,keepaspectratio]{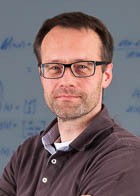}}]
    {Leif Kobbelt} received his master’s and PhD degrees from the University of Karlsruhe, Germany, in 1992 and 1994, respectively. He is a full professor and the head of the Computer Graphics Group with the RWTH Aachen University, Germany. His research interests include all areas of computer graphics and geometry processing with a focus on multi-resolution and free-form modeling as well as the efficient handling of polygonal mesh data.
\end{IEEEbiography}

\begin{IEEEbiography}[{\includegraphics[width=1in,height=1.25in,clip,keepaspectratio]{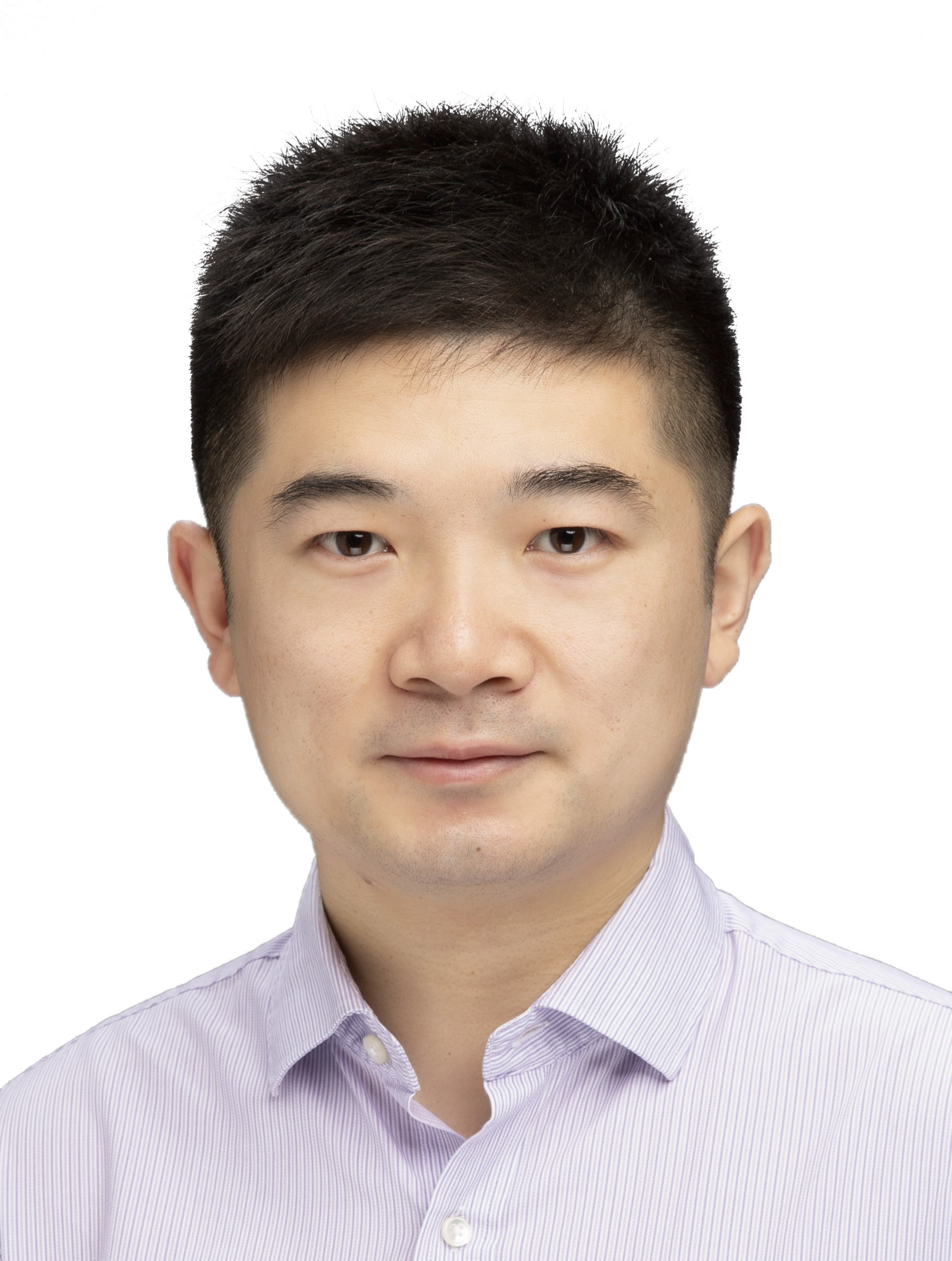}}]
    {Lin Gao} received his bachelor’s degree in mathematics from Sichuan University and his PhD degree in computer science from Tsinghua University. He is currently an Associate Professor at the Institute of Computing Technology, Chinese Academy of Sciences. He has been awarded the Newton Advanced Fellowship from the Royal Society and the Asia Graphics Association Young Researcher Award. His research interests include computer graphics and geometric processing.
\end{IEEEbiography}

\end{document}